%% file: main.tex
\documentclass[10pt,twocolumn,letterpaper]{article}

\usepackage{iccv}      


\definecolor{iccvblue}{rgb}{0.21,0.49,0.74}
\usepackage[pagebackref,breaklinks,colorlinks,allcolors=iccvblue]{hyperref}
\usepackage{booktabs}    
\usepackage{multirow}    
\usepackage{array}       
\usepackage{float}
\usepackage{makecell}
\usepackage{multirow}
\usepackage{array}
\usepackage{colortbl}
\usepackage{placeins}

\usepackage{pifont}
\newcommand{\blackcheck}{\ding{51}}
\usepackage{xcolor}  
\newcommand{\redcross}{\textcolor{red}{$\times$}}  

\def\paperID{9344} 
\def\confName{ICCV}
\def\confYear{2025}

\title{MME-Unify: A Comprehensive Benchmark for Unified Multimodal Understanding and Generation Models}

\author{
Wulin Xie$^{1,*}$, Yi-Fan Zhang$^{1,5,*,\dagger}$, Chaoyou Fu$^{2,5}$, Yang Shi$^{3}$ \\
Bingyan Nie$^{1}$, Hongkai Chen$^{4}$, Zhang Zhang$^{1}$, Liang Wang$^{1}$, Tieniu Tan$^{2}$ \\
\\
$^{1}$CASIA, $^{2}$NJU, $^{3}$PKU,  $^{4}$Vivo, $^{5}$M-M-E \\
$^{*}$Equal Contribution \quad $^{\dagger}$Project leader \\
\url{https://mme-unify.github.io/}
}

\begin{document}

\maketitle

\input{sec/0_abstract}

\input{sec/1_intro}

\input{sec/2_formatting}
\input{sec/3_finalcopy}
{
\small
\bibliographystyle{ieeenat_fullname}
\bibliography{main}
}
\appendix
\input{sec/X_suppl}
\end{document}

%% file: sec/0_abstract.tex
\begin{abstract}

Unified Multimodal Large Language Models (U-MLLMs) have garnered considerable interest for their ability to seamlessly integrate generation and comprehension tasks. However, existing research lacks a unified evaluation standard, often relying on isolated benchmarks to assess these capabilities. Moreover, current work highlights the potential of ``mixed-modality generation capabilities'' through case studies—such as generating auxiliary lines in images to solve geometric problems, or reasoning through a problem before generating a corresponding image. Despite this, there is no standardized benchmark to assess models on such unified tasks. To address this gap, we introduce MME-Unify, also termed as MME-U, the first benchmark designed to evaluate multimodal comprehension, generation, and mixed-modality generation capabilities. For comprehension and generation tasks, we curate a diverse set of tasks from 12 datasets, aligning their formats and metrics to develop a standardized evaluation framework. For unified tasks, we design five subtasks to rigorously assess how models’ understanding and generation capabilities can mutually enhance each other. Evaluation of 12 U-MLLMs, including Janus-Pro, EMU3, and Gemini2-Flash, reveals significant room for improvement, particularly in areas such as instruction following and image generation quality.

\end{abstract}

%% file: sec/1_intro.tex
\begin{figure}[t]
  \centering
  \begin{subfigure}[b]{0.4\textwidth}
    \centering
    \includegraphics[width=1.0\linewidth]{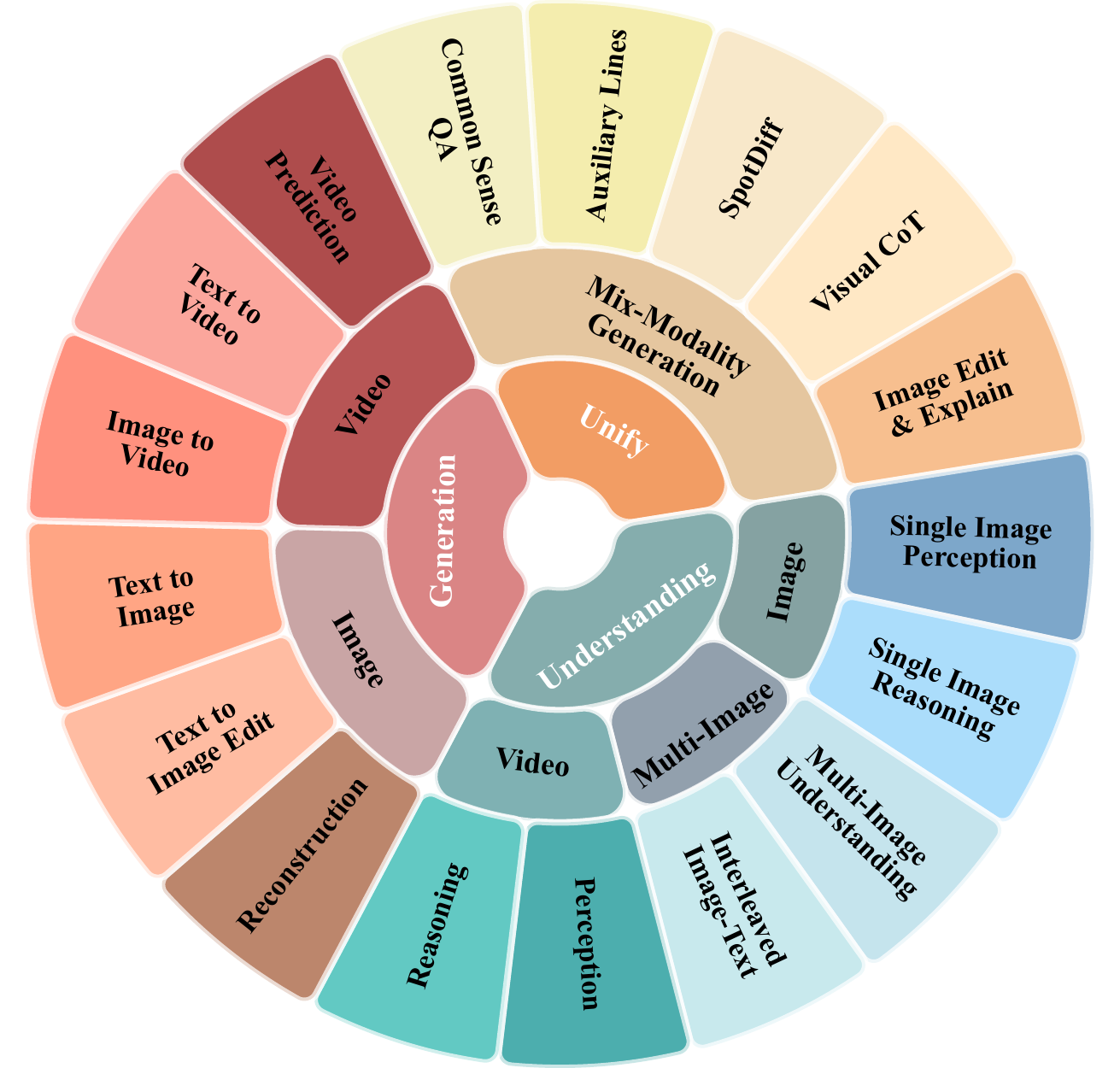}
    \caption{MME-U tasks.}
    \label{fig:sub-a}
  \end{subfigure}
  \begin{subfigure}[b]{0.48\textwidth}
    \centering
    \includegraphics[width=\linewidth]{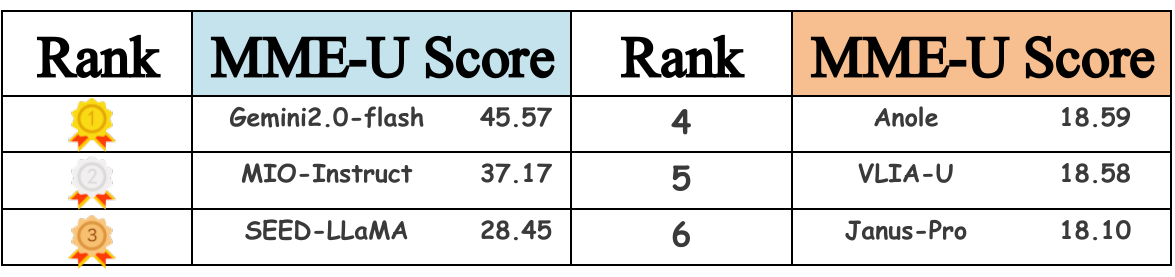}
    \caption{Leaderboard.}
    \label{fig:sub-b}
  \end{subfigure}
\caption{\textbf{A comprehensive visualization of the diverse tasks in MME-U and the leaderboard.} The figure (a) illustrates the wide-ranging nature of the tasks covered in our benchmark, which spans from traditional understanding tasks to complex mixed-modality generation challenges. Additionally, the leaderboard (b) highlights the performance rankings of various U-MLLMs in our benchmark.}
  \label{fig:teaser}
\vspace{-0.3cm}
\end{figure}
\section{Introduction}
Unlike traditional MLLMs (e.g., GPT-4V) and purely generative models (e.g., DALL-E 3), U-MLLMs~\cite{xie2024show, wang2024emu3, chen2025janus, ma2024janusflow} excel in processing mixed-modal inputs and outputs, providing enhanced flexibility and the ability to address a broader spectrum of complex tasks. Recently, closed-source U-MLLMs, such as GPT-4o and Gemini 2.0 Flash, have demonstrated exceptional generative capabilities, impressing in both instruction comprehension and image creation, as shown in Figure~\ref{fig:teaser}. These models exhibit an extraordinary grasp of image details, even surpassing proprietary generative models. However, this versatility also introduces considerable challenges in comprehensively evaluating their capabilities, primarily due to two key issues:

\begin{figure*}[t]
\vspace{-0.2cm}
    \centering
\includegraphics[width=\linewidth]{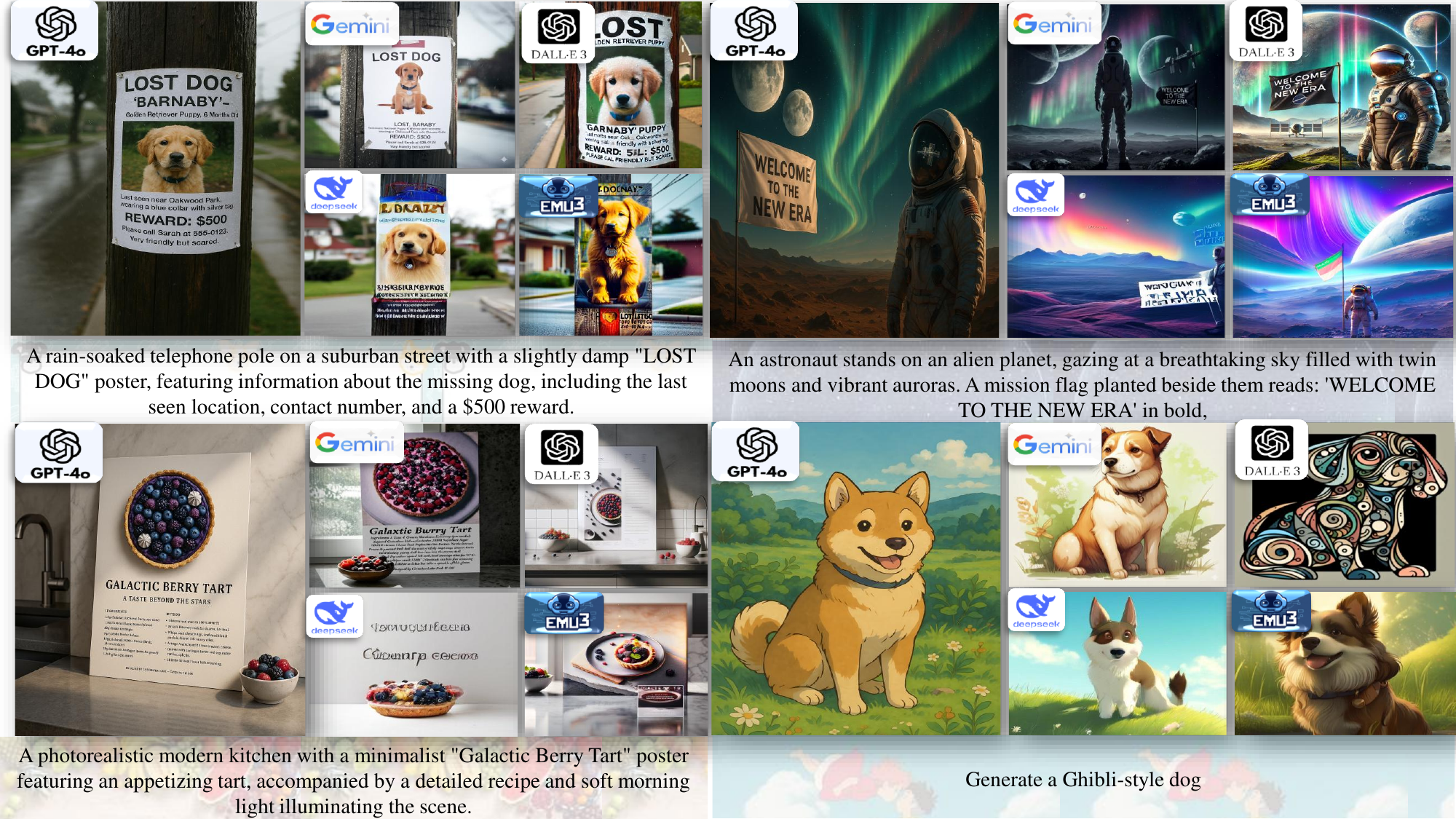}
    \caption{\textbf{Complex instruction-based image generation comparison} of results from open-source U-MLLMs (DeepSeek-Janus Flow, EMU3), closed-source U-MLLMs (GPT-4o, Gemini-2), and proprietary models (DALLE-3). The closed-source U-MLLMs have demonstrated abilities surpassing proprietary generation models, with a significantly larger gap compared to open-source models.}
    \label{fig:teaser}
\vspace{-0.2cm}
\end{figure*}

\begin{itemize}
    \item \textbf{Lack of Standardized Benchmarks for Traditional Tasks.} Existing works typically evaluate traditional generation and understanding tasks separately, using various benchmarks. However, the benchmarks chosen across studies are inconsistent, leading to unfair comparisons. Moreover, the evaluation methods differ significantly—multimodal understanding tasks may involve varied formats such as multiple-choice questions, GPT-4 scoring, or binary classification, while multimodal generation tasks may rely on metrics like CLIP score or FID. This diversity in evaluation makes it difficult to derive an intuitive and unified performance score.
    \item \textbf{Absence of Benchmarks for Mixed-Modality Generation\footnote{also termed as unify tasks}.} The most distinctive feature of U-MLLMs is their mixed-modality generation capabilities, which demonstrate the synergistic interaction between multiple modalities. For instance, image editing requires understanding textual instructions and identifying objects to be modified, while solving geometry problems involves comprehending the problem, drawing auxiliary lines, and performing logical reasoning. Despite these advanced capabilities, most methods only showcase simple cases, lacking a standardized benchmark to rigorously assess these complex mixed-modality tasks.
\end{itemize}

To address these challenges, we propose a comprehensive evaluation framework for U-MLLMs, which is shown in Figure~\ref{fig:teaser}. For \textbf{traditional generation and understanding tasks}, we sample data from 12 existing datasets, resulting in 10 tasks with 30 subtasks. On the understanding side, these tasks encompass single-image, multi-image, and video-based perception and reasoning tasks, covering a wide range of difficulties—from simple visual question-answering (VQA) to high-resolution VQA in real-world scenarios and long-video understanding. On the generation side, we include tasks such as image/video generation and editing, as well as more complex conditional image generation and image-to-video generation, aiming to cover the full spectrum of existing generative tasks. To simplify evaluation and provide a unified score, we manually reformat all understanding tasks into multiple-choice questions, reporting accuracy as the primary metric. For generation tasks, we standardize the evaluation scores and normalize them to provide a consistent metric. This approach reduces the difficulty of benchmark collection and mitigates the issue of inconsistent evaluation metrics across studies.

\textbf{For the Unified Tasks}, we constructed five tasks: (1) Image Editing and Explaining, where the model first understands complex editing instructions and edits an image; (2) Common Sense Question Answering, where the model answers a question and generates the corresponding image; (3) Auxiliary Lines, where the model draws auxiliary lines for geometry problems and then solves them; (4) SpotDiff, where the model identifies and draws the differences between two images; and (5) Visual CoT, where the model generates step-by-step strategies for navigating a maze and visualizes the next state. These tasks evaluate a model's ability to perform sequential reasoning and generate corresponding multimodal outputs at each step. All tasks are carefully formatted as multiple-choice questions to facilitate consistent, fair, and objective evaluation.

We evaluate 12 existing U-MLLMs, including Janus-Pro, EMU3, VILA-U, and MiniGPT-5. To provide context for their performance, we also compare them with specialized understanding models (e.g., Claude-3.5 Sonnet, Qwen2.5-VL) and generative models (e.g., DALL-E-2, DALL-E-3). This comprehensive evaluation not only underscores the strengths and weaknesses of U-MLLMs but also establishes a standardized benchmark for future research in this rapidly evolving field. For example, we uncover several key experimental findings, as illustrated in Figure~\ref{fig:teaser}. Currently, U-MLLMs exhibit significant variance in rankings across three dimensions, and no single model has emerged as the best performer across multiple capabilities. Moreover, the performance gap between models is substantial. Finally, the current open-sourced U-MLLMs still exhibit a significant gap in performance compared to specialized models in both understanding and generation tasks. Additionally, while many works claim to handle mixed-modality generation, our unify task tests demonstrate that the majority of existing U-MLLMs struggle to consistently and effectively process these types of tasks.

\begin{figure*}
  \centering
\includegraphics[width=1.0\linewidth]{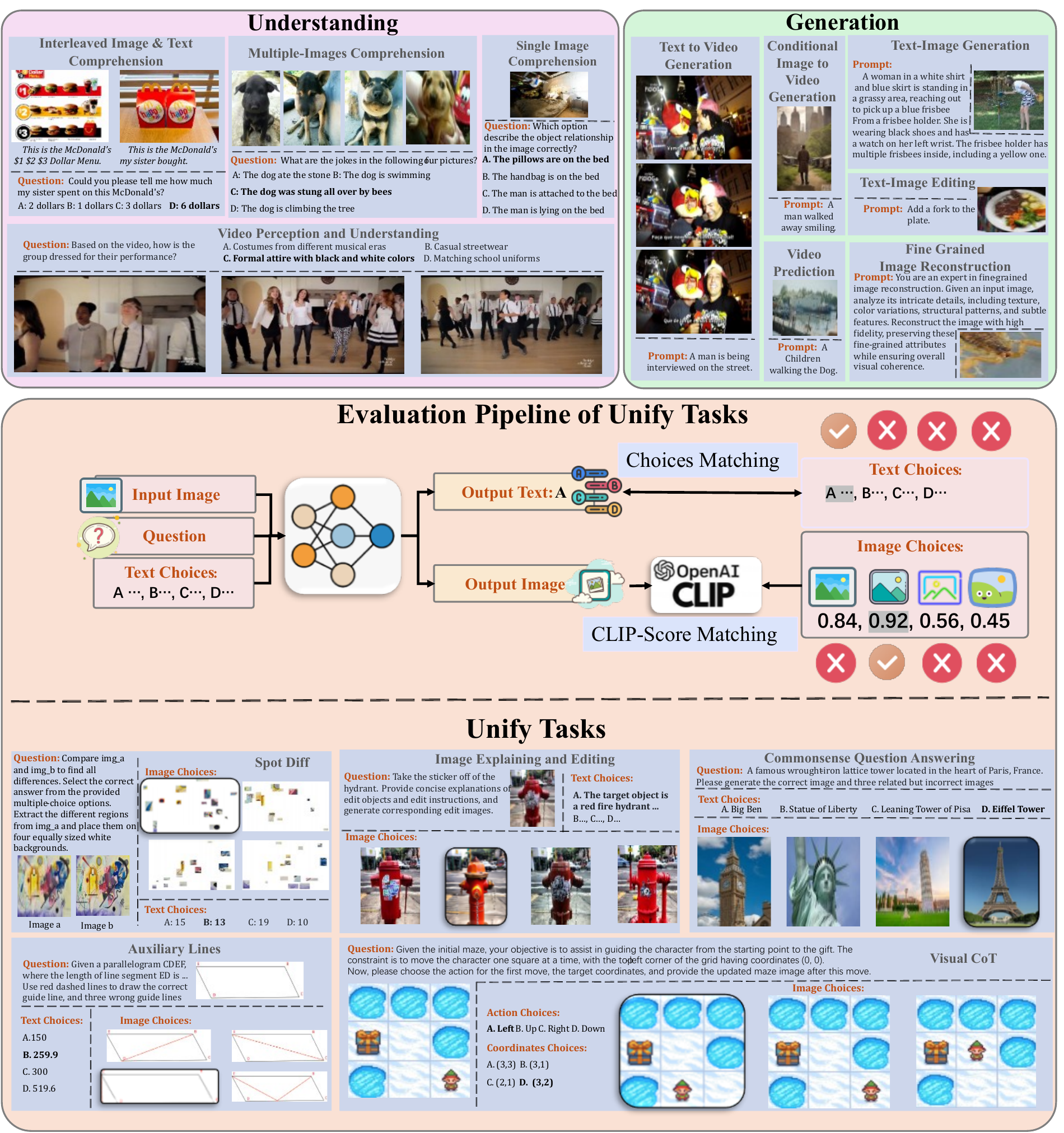}
\caption{\textbf{Diagram of our MME-Unify.} Our benchmark consists of 3 main domains, encompassing 15 subtasks to comprehensively evaluate U-MLLMs' understanding, generation, and unified capabilities. Specifically, each unify task includes at least one question, an input image, multiple text choices, and image choices. The image choices consist of a correct answer image and a set of manually crafted negative samples. During the evaluation process, we input the image, question, and text options, and the U-MLLMs are required to select the correct text answer and generate an image. The text answer is evaluated by matching it with the correct answer, while the generated image is compared with the constructed image choices. If the CLIP score between the generated image and the correct answer image is the highest, it is considered correct; otherwise, it is deemed incorrect.}  \label{fig:visualization}
\end{figure*}


%% file: sec/3_finalcopy.tex
\begin{table*}
\centering
\resizebox{\textwidth}{!}{

\begin{tabular}{lcccccccccccc}
\toprule
Benchmark & Question & Year& SIPU  & MITIU & VPU  & FIR & TIE & TIG & CIVG & TVG & VP & UT \\
\midrule
MSR-VTT~\cite{xu2016msr} & $10,000$ & CVPR 2016 & \redcross  & \redcross & \blackcheck & \redcross & \redcross & \redcross & \redcross & \redcross & \redcross & \redcross \\
MMBench~\cite{liu2023mmbench} & $3,217$ & arXiv 2023 & \blackcheck & \redcross & \redcross & \redcross & \redcross & \redcross & \redcross & \redcross & \redcross & \redcross \\
GenEval~\cite{ghosh2023geneval} & $1,200$ & arXiv 2023 & \redcross  & \redcross & \redcross & \redcross & \redcross & \blackcheck & \redcross & \redcross & \redcross & \redcross\\
MagicBrush~\cite{zhang2023magicbrush} & $10,338$ & NeurIPS 2023 & \redcross  & \redcross & \redcross & \blackcheck & \blackcheck & \redcross & \redcross & \redcross & \redcross & \redcross \\
VBench~\cite{huang2024vbench} & $1,600$ & CVPR 2024 & \redcross  & \redcross & \redcross & \redcross & \redcross & \redcross & \redcross & \blackcheck & \redcross & \redcross \\
SEED-Bench2~\cite{li2023seed2} & $19,242$ & arXiv 2024 & \blackcheck  & \blackcheck & \blackcheck & \blackcheck & \blackcheck & \blackcheck & \redcross & \redcross & \redcross & \redcross \\  
Emu-Edit~\cite{sheynin2024emu} & $5,611$ & CVPR 2024 & \redcross  & \redcross & \redcross & \blackcheck & \blackcheck & \redcross & \redcross & \redcross & \redcross & \redcross \\
TIP-I2V~\cite{wang2024tip} & $500,000$ & arXiv 2024 & \redcross  & \redcross & \redcross & \redcross & \redcross & \redcross & \blackcheck & \redcross & \blackcheck & \redcross \\
MMBench-Video~\cite{fang2025mmbench} & $2,000$ & NeurIPS 2024 & \redcross  & \redcross & \blackcheck & \redcross & \redcross & \redcross & \redcross & \redcross & \redcross & \redcross \\
MME\cite{fu2023mme} & $2,374$ & arXiv 2023 & \blackcheck  & \redcross & \redcross & \redcross & \redcross & \redcross & \redcross & \redcross & \redcross & \redcross \\
Video-MME~\cite{fu2024video} & $2,700$ & CVPR 2025 & \redcross  & \redcross & \blackcheck & \redcross & \redcross & \redcross & \redcross & \redcross & \redcross & \redcross \\
MME-RealWorld~\cite{zhang2024mme} & $29,429$ & ICLR 2025 & \blackcheck  & \redcross & \redcross & \redcross & \redcross & \redcross & \redcross & \redcross & \redcross & \redcross \\
\midrule
MME-Unify (ours) & $4,104$ & 2025 & \blackcheck  & \blackcheck & \blackcheck & \blackcheck & \blackcheck & \blackcheck & \blackcheck & \blackcheck & \blackcheck & \blackcheck \\
\bottomrule
\end{tabular}
}
\caption{\textbf{Comparison of MME-U and other Benchmark.} \textbf{SIPU:} Single Image Perception \& Understanding; \textbf{MITIU:} Multiple \& Interleaved Image-Text Understanding; \textbf{VPU:} Video Perception \& Understanding; \textbf{CIVG:} Conditional Image-to-Video Generation; \textbf{FIR:} Fine-grained Image Reconstruction; \textbf{TIE:} Text-Guided Image Editing; \textbf{TIG:} Text-to-Image Generation; \textbf{TVG:} Text-to-Video Generation; \textbf{VP:} Video Prediction; \textbf{UT:} Unified Task.}
\label{tab:comparison} 
\end{table*}

\section{MME-Unify}
This section outlines the data collection, question annotation, and evaluation strategy for MME-Unify. Figures~\ref{fig:teaser} and~\ref{fig:visualization} provide visual representations of subtasks and samples across three domains, while Table~\ref{tab:comparison} compares MME-U with existing benchmarks. MME-U categorizes U-MLLM capabilities into three areas: (1) Multimodal Understanding, (2) Multimodal Generation, and (3) Unify Capability, highlighting the diverse aspects of model performance.

\subsection{Multi-Modal Understanding}
\textbf{Data Collection.} Multimodal understanding tasks are divided into three subcategories based on visual input type:
\begin{itemize}
\item Single-Image Perception and Understanding (SIPU). Evaluates image-text pair comprehension.
\item Multi-Image \& Interleaved Text-Image Understanding (MITIU). Assesses the model's ability to handle and process multi-image and interleaved text-image inputs.
\item Video Perception and Understanding (VPU). Measures video comprehension capability.
\end{itemize}
To ensure comprehensive coverage of various image and video understanding scenarios, we collect 1,900 samples from 5 benchmarks such as MME and MMBench, encompassing over 24 tasks. This includes 1,600 perception tasks, such as OCR, diagram and table understanding, and spatial perception, along with 300 reasoning tasks, including attribute reasoning and action reasoning, with at least 50 QA pairs per sub-task. Additional details can be found in Appendix Figure~\ref{fig:subtask_disturbution} and Appendix Table~\ref{tab:task-dataset-sampling}. More visualization examples can be found in Appendix Figure~\ref{fig:Visualization_Understanding}.

\textbf{QA Pairs Reformulation.} To standardize the evaluation of the understanding task, we convert all the collcted data into multiple-choice QA pairs, with one correct option and the remaining options carefully designed to be closely related to it.  For models that can accept only single-image input, we use the first image from the multi-image input or the first frame from the video input. For models that cannot process video files (e.g., MP4 files), we uniformly sample six key frames from the video to serve as the visual input.

\textbf{Evaluation Strategy.} To fairly evaluate MLLM outputs, we apply rule-based filtering to match model responses with answer options, similar to MME-Realworld~\cite{zhang2024mme,fu2024video}. Furthermore, to eliminate positional bias inherent in multiple-choice questions, the correct answer is randomly shuffled among the four available options. We then calculate the average accuracy across all sub-tasks and derive the overall understanding score, providing a fair, robust, and unbiased evaluation of the model's performance.

\subsection{Multi-Modal Generation}
Multimodal generation involves various tasks for image and video modalities, which can be further subdivided based on application, as shown in Figure~\ref{fig:visualization}: 1. \textit{Fine-grained Image Reconstruction (FIR).} Given an original image, the model is required to restore detailed features and local textures.
    2. \textit{Text-guided Image Editing (TIE). }Edit or modify an image based on textual instructions.
    3. \textit{Text-guided Image Generation (TIG).} Given a text description, the model needs to generate an image that matches it. 
    4. \textit{Conditional Image-to-Video Generation (CIVG).} Generate a dynamic video sequence based a given image and text prompt.
    5. \textit{Text-guided Video Generation (TVG).} Generate a video sequence based on a textual description.
    6. \textit{Video Prediction (VP).} Predict subsequent frames or the complete video sequence based on the information from the first frame.

\textbf{Data Collection.} Data is collected from benchmark datasets, such as COCO~\cite{lin2014microsoft}, Emu-Edit~\cite{sheynin2024emu}, MSR-VTT~\cite{xu2016msr}, ensuring at least 200 samples for each task. For video prediction, videos are sourced from the Pexel Video website\footnote{\url{https://www.pexels.com/videos/}} and the first frame is used for prediction. Detailed data sources and sample sizes are in Appendix Table~\ref{tab:task-dataset-sampling}. More visualization examples can be found in Figure~\ref{fig:Visualization_Generation}.

\textbf{QA Pairs Reformulation.} Due to the diversity of generation tasks and their varied data sources, the collected samples contain redundant attributes and inconsistent number of images, videos, and other multimodal data. We aim to provide a streamlined, unified evaluation framework. To achieve this, we contribute the following: 
\begin{itemize}
    \item \textbf{Attribute Unification Pipeline.} First, we summarize all attributes appearing in the data, which exceed 30 types, creating significant complexity. We then manually eliminate task-irrelevant attributes and merge similar attributes across different tasks. For example, text attributes are represented as \textit{Text Prompt}, image attributes as \textit{Src Image} and \textit{Ref Image} based on their input/output roles, and video attributes as \textit{Video}. For any task where an attribute is not required, its corresponding value remains empty.
    \item \textbf{Task-Specific Prompt Engineering.} To ensure that the model can effectively generate outputs that meet the task requirements, we establish specific system prompts for each subtask. Each sample’s text prompt or src image serves as the input, while the reference image or video acts as the ground truth answer. Through standardizing attribute values and constructing tailored prompts, we convert diverse samples from different tasks into a unified format for evaluating multimodal generation tasks.
\end{itemize}

\textbf{Evaluation Strategy.} Evaluating multimodal generation tasks with a unified metric is challenging due to the diversity of subdomains and their distinct metrics (e.g., CLIP-I, CLIP-T, FVD, FID). To address this, we: (1) Perform domain-specific preliminary evaluations using standard metrics; (2) Standardize all metrics to a consistent (0, 100) scale, converting non-positive indicators into positive ones; and (3) Compute the average of standardized scores to derive the final generation score. This approach ensures cross-task comparability while maintaining domain-specific evaluation rigor. Detailed metrics and standardization methods are provided in Appendix~\ref{sec:eval_met}

\subsection{Unify Capability}
MME-U contains five unified subtasks: (1) Common Sense Question Answering (CSQ), (2) Image Editing and Explaining (IEE), (3) SpotDiff (SD), (4) Auxiliary Lines (AL), and (5) Visual CoT (VCoT). Each subtask includes at least 50 manually constructed samples and is structured with task-specific instructions and question templates that require mixed-modality input-to-output generation.

\textbf{Common Sense Question Answering.}  
This task evaluates U-MLLMs' ability to associate commonsense descriptions with visual features, such as linking ``the tomb of an ancient Egyptian pharaoh'' to a pyramid or ``China’s national treasure'' to a panda. Our approach involves:
    1. \textit{Question Construction.} Using GPT-4o, we generate riddle-like questions based on commonsense concepts, with similar but incorrect words as negative options. For example, when the answer is ``panda,'' we select ``brown bear'' or ``polar bear'' as negative options to increase difficulty.
    2. \textit{Image Collection.} We manually gather images from the internet corresponding to the correct and their negative options.
    3. \textit{Task Execution.} U-MLLMs are prompted to select the correct textual option and generate the corresponding image.
Detailed procedures and the prompt are in Figure~\ref{fig:Construction_Process}(a) and~\ref{figure:Prompt_for_CSQ}.

\textbf{Image Editing and Explanation.}  
This task evaluates U-MLLMs' ability to understand complex editing instructions and generate accurate modifications. Our methodology includes: 1. \textit{Data Collection.} We source data (source images, editing instructions, and reference images) from the Emu-Edit dataset. 2. \textit{Textual QA Construction.} Using GPT-4o, we generate accurate interpretations of editing targets and three incorrect interpretations for textual multiple-choice questions. 3. \textit{Visual QA Construction.} The correct instruction corresponds to the target image in Emu-Edit. For incorrect instructions, we input them into InstructPix2Pix \cite{brooks2022instructpix2pix} to generate negatively edited images, forming image-based multiple-choice questions. 4. \textit{Task Execution.} Given the corresponding prompt, source image, and editing instructions, the model must first produce a correct understanding of the editing target and instructions, and then generate an edited image based on that understanding. Detailed procedures and the system prompt are in Figure~\ref{fig:Construction_Process}(b) and~\ref{figure:Prompt_for_IEE}.

\textbf{SpotDiff.} When identifying differences between two similar images, humans typically need to recall the exact locations of these differences to accurately count them. This task evaluates U-MLLMs' ability to identify and recall differences between similar images, simulating human visual reasoning.
Our approach involves: 1.\textit{Data Collection:} We sample image pairs with annotated differences from the SpotDiff website\footnote{\url{https://www.allstarpuzzles.com/spotdiff}}.
    2. \textit{Textual QA Construction.} Using the annotated difference count, we create textual multiple-choice questions with three incorrect counts (±10 from the true value).
    3. \textit{Visual QA Construction.} We place the annotated difference regions from the image pair onto a white background as the correct answer, and randomly crop other areas to place them on the background as incorrect answers.
    4. \textit{Task Execution.} U-MLLMs must identify the difference regions between the two images and draw them onto the white background, while also selecting the correct difference count.
Detailed procedures in Figure~\ref{fig:Construction_Process}(c), and the system prompt is provided in Figure~\ref{figure:Prompt_for_SD}.

\textbf{Auxiliary Lines.}  
This task evaluates U-MLLMs' ability to integrate understanding and generation by solving geometric problems requiring auxiliary lines. {Our methodology includes:}  
1. \textit{Data Selection.} We filter the Geometry3K dataset for problems requiring auxiliary lines, extracting logical forms (e.g., ``Triangle(A, B, C)''), choices, and answers.  
2. \textit{Textual QA Construction.} Using GPT-4o, we generate natural language QA pairs (Question, Choices, Answer) for textual multiple-choice questions.  
3. \textit{Visual QA Construction.} We manually solve each sampled geometric problem by drawing the correct auxiliary lines on its diagram, and we construct three additional diagrams with erroneous auxiliary lines.  
4. \textit{Task Execution.} U-MLLMs must first generate a geometric diagram with auxiliary lines, and then, based on that diagram, solve the problem by selecting the correct answer.   Detailed procedures appear in Figure~\ref{fig:Construction_Process}(d), and the prompt is provided in Figure~\ref{figure:Prompt_for_AL}.

\textbf{Visual CoT.}  
This task evaluates U-MLLMs' step-by-step decision-making in maze navigation, simulating real-world problem-solving. {Our approach involves:}  
1. \textit{Maze Generation.} Using the OpenAI API, we create maze configurations of varying sizes (3×3, 4×4, 5×5) and layouts.  
2. \textit{Action Specification.} For each step, we manually define actions (Up, Right, Down, Left, Finish) and coordinates, updating the maze layout via the API.  
3. \textit{QA Construction.}  
   - \textit{Action Questions.} Options are uniformly set as Up, Right, Down, and Left, with the correct answer manually determined.  
   - \textit{Coordinate/Image Questions.} The correct answers for each step's coordinates and state image are manually defined, and negative samples are also manually specified.  
4. \textit{Task Execution.} U-MLLMs receive the initial maze state and task definition, then are prompted to generate actions, coordinates, and maze images iteratively. After the first step, we add the action, coordinate, and image from the previous decision into the system prompt as history information. The model iterates, outputting each step's decision until the target is reached\footnote{task requires an average of 3.5 steps per sample, with a minimum of two and a maximum of seven steps (as shown in Figure~\ref{fig:vcot_steps}).}. Detailed procedures appear in Figure~\ref{fig:Construction_Process}(e), and the prompts are in Figure~\ref{figure:Prompt_for_VCoT_1} and~\ref{figure:Prompt_for_VCoT_2}.

\textbf{Evaluation Strategy.} The unified tasks evaluation combines text-based and image-based multiple-choice questions across all subtasks. Our evaluation framework includes:
\begin{enumerate}
    \item \textit{Textual QA Evaluation.}  
    For image explanation and editing, we compute CLIP-T similarity between the generated explanation and each option, selecting the one with the highest similarity as correct. For other tasks, U-MLLMs directly select the correct option from the provided choices.
    
    \item \textit{Image-Based QA Evaluation.}  
    We compute CLIP-I similarity between the generated image and each candidate option, selecting the option with the highest score as the model’s prediction.
    
    \item \textit{Task-Specific Rules.}  
    For each task we calculate two accuracy metrics---\textbf{acc} and \textbf{acc+}---where \textbf{acc} is defined as the average of the text option accuracy and the image accuracy, and \textbf{acc+} represents the accuracy for samples where both the textual and image-based answers are correct. Specifically, for the Visual CoT task, each step is treated as a multiple-choice question, and the accuracy of action, accordinate and image are calculated separately, and the average of these three accuracies is calculated as \textbf{acc}, while the accuracy of successfully completing the maze is used to calculate \textbf{acc+}. The detailed calculation process can be found in the Appendix~\ref{sec:eval_met}
\end{enumerate}
We then calculate the average \textbf{acc} of all subtasks as the unified score, and the overall MME-U score is the average of the understanding, generation, and unified scores.

\begin{table*}[htbp]
\centering
\renewcommand{\arraystretch}{1.4} 
\setlength{\tabcolsep}{3pt} 
\resizebox{1\textwidth}{!}{
\Large 
\begin{tabular}{l|l|cccc|ccccccc|cccccc|c}
\hline
\textbf{Method} & \textbf{LLM} & \multicolumn{4}{c|}{\textbf{Understanding}} & \multicolumn{7}{c|}{\textbf{Generation}} & \multicolumn{6}{c|}{\textbf{Unify}} & \textbf{MME-U Score} \\
\hline
 &Task Split  & \textbf{SIPU} & \textbf{MITIU} & \textbf{VPU} & \textbf{Avg} & \textbf{CIVG} & \textbf{FIR} & \textbf{TIE} & \textbf{TIG}  & \textbf{TVG} & \textbf{VP} & \textbf{Avg} & \textbf{IEE} & \textbf{CSQ} & \textbf{AL} & \textbf{SD} & \textbf{VCoT} & \textbf{Avg} & \textbf{Avg} \\
\hline
 &  QA pairs & 1200 & 400 & 364 & 1964 & 600 & 200 & 200 & 200 & 200 & 194 & 1594 & 200 & 100 & 52 & 104 & 90 & 546 & 4104 \\
\hline
\rowcolor{gray!15} \multicolumn{20}{c}{\LARGE\textit{\textbf{Understanding Models}}} \\

SliME-7B & Vicuna-7B & 58.50 & 43.53 & 36.02 & 46.02 & - & - & - & - & - & - & - & - & - & - & - & - & - & 15.34 \\
VITA-1.5 & Qwen-7B & 70.67 & 56.00 & 56.04 & 60.89 & - & - & - & - & - & - & - & - & - & - & - & - & - & 20.30 \\
Qwen2.5-VL-Instruct & Qwen-7B & 75.08 & 53.50 & 57.14 & 61.91 & - & - & - & - & - & - & - & - & - & - & - & - & - & 20.64 \\
Claude-3.5-sonnet & - & 75.83 & 53.25 & 58.52 & 62.53  & - & - & - & - & - & - & - & - & - & - & - & - & - & 20.84 \\
GPT-4o & - & 74.01 & 54.50 & 59.34 & 62.62 & - & - & - & - & - & - & - & - & - & - & - & - & - & 20.87 \\
Gemini2.0-flash & - & \underline{80.92} & 61.75 & \underline{64.64} & \underline{69.10} & - & - & - & - & - & - & - & - & - & - & - & - & - & 23.03 \\
\hline
\rowcolor{gray!15} \multicolumn{20}{c}{\LARGE\textit{\textbf{Generative Models}}} \\
DALL-E-2 & - & - & - & - & - & - & - & - & 50.62 & - & - & 8.44 & - & - & - & - & - & - & 2.81 \\
DALL-E-3 & - & - & - & - & - & - & - & - & 51.40 & - & - & 8.57 &  &  &  &  &  &  &  2.86 \\
OmniGen & - & - & - & - & - & - & 48.82 & \underline{43.82} & 51.05 & - & - & 23.95 & - & - & - & - & - & - & 7.98 \\
CogVideoX & - & - & - & - & - & \underline{68.05} & - & - & - & \underline{69.62} & \underline{87.61} & 37.54 & - & - & - & - & - & - & 12.51 \\
\hline
\rowcolor{gray!15} \multicolumn{20}{c}{\LARGE\textit{\textbf{Unified Models}}} \\
Show-o & Phi-1.5 & 32.47 & 34.75 & 25.66 & 30.96  & - & - & - & 43.54 & - & - & 7.26 & - & - & - & - & - & - & 12.74 \\
Emu3 & LLama-8B & 45.75 & 30.50 & 23.32 & 33.19 & - & - & - & 49.08 & - & - & 8.18 & - & - & - & - & - & - & 13.79 \\
HermesFlow & Phi-1.5 & 41.49 & 33.00 & 28.32 & 34.27 & - & - & - & 46.48 & - & - & 7.75 & - & - & - & - & - & - & 14.01 \\
GILL$^*$ & OPT-6-7B & 22.18 & 6.00 & 3.56 & 10.58 & - & 50.67 & 35.71 & 46.60 & - & - & 22.16 & 24.25 & 21.29 & 8.66 & 6.67 & 1.90 & 12.55 & 15.10 \\
Janus-Flow & DeepSeek-LLM-1.5b-base & 63.17 & 32.00 & 35.16 & 43.44 & - & - & - & 32.88 & - & - & 5.48 & - & - & - & - & - & - & 16.31 \\
MiniGPT-5$^*$ & Vicuna-7B & 19.25 & 10.92 & 15.93 & 15.37 & - & 38.96 & 35.04 & 35.48 & - & - & 18.25 & 22.80 & 34.13 & 14.37 & 5.00 & 2.08 & 15.67 & 16.43 \\
Janus-Pro & DeepSeek-LLM-7b-base & 59.56 & 43.50 & 42.22 & 48.43 & - & - & - & 35.29 & - & - & 5.88 & - & - & - & - & - & - &  18.10  \\
VILA-U & LLama-7B & 51.04 & 32.25 & 36.54 & 39.95 & - & - & - & 45.10 & 49.64 & - & 15.79 & - & - & - & - & - & - & 18.58 \\
Anole$^*$ & - & 17.17 & 14.50 & 9.00 & 13.56 & - & 36.64 & 43.42 & 41.52 & - & - & 19.91 & 18.55 & 59.65 & 14.42 & 15.00 & 3.89 & 22.30 & 18.59 \\
SEED-LLaMA$^*$ & LLaMA2-Chat-13B & 49.17 & 33.00 & 36.26 & 39.48 & - & 57.00 & 42.26 & 41.96 & - & - & 23.54 & 22.00 & 51.49 & 12.50 & 22.00 & 3.61 & 22.32 & 28.45 \\
MIO-Instruct$^*$ & MIO-7B & 52.00 & 33.50 & 39.01 & 41.50 & 51.24 & 59.29 & 43.66 & 48.23 & 51.88 & 66.37 & \underline{53.45} & 24.16 & 38.50 & 8.66 & 11.50 & 0 & 16.56 & 37.17 \\
Gemini2.0-flash-exp$^*$ & - & 72.58 & \underline{68.25} & 54.90 & 65.24 & - & \underline{77.61} & 43.54 & \underline{57.56} & - & - & 29.79 & \underline{38.42} & \underline{74.75} & \underline{47.12} & \underline{26.00} & \underline{12.41} & \underline{40.74} & \underline{45.57} \\
\hline
\end{tabular}%
}
\caption{\textbf{Comparison of multimodal models on understanding, generation, unifying tasks, and overall MME-U Score.} \textbf{SIPU:} Single Image Perception \& Understanding; \textbf{MITIU:} Multiple \& Interleaved Image-Text Understanding; \textbf{VPU:} Video Perception \& Understanding; \textbf{CIVG:} Conditional Image-to-Video Generation; \textbf{FIR:} Fine-grained Image Reconstruction; \textbf{TIE:} Text-Guided Image Editing; \textbf{TIG:} Text-to-Image Generation; \textbf{TVG:} Text-to-Video Generation; \textbf{VP:} Video Prediction; \textbf{IEE:} Image Editing and Explaining; \textbf{CSQ:} Common Sense Question Answering; \textbf{AL:} Auxiliary Lines; \textbf{SD:} SpotDiff; \textbf{VCoT:} Visual CoT. $^*$ denotes U-MLLMs with the ability to generate interleaved images and texts, while `-' indicates that the model is unable to finish the corresponding task and \underline{underlined} content signifies the best performance within a single model across all methods on this task.}
\label{tab:main_result}
\medskip
\normalsize 
\end{table*}

\section{Experiment}

We evaluate a total of 22 MLLMs and U-MLLMs, including DeepSeek-Janus-Pro~\cite{chen2025janus}, DeepSeek-Janus-Flow~\cite{ma2024janusflow}, SliME~\cite{zhang2024beyond}, VITA-1.5~\cite{fu2025vita},  Gemini2.0-flash~\cite{deepmind2024gemini}, Gemini2.0-flash-exp~\cite{deepmind2024gemini},  Claude-3.5sonnet~\cite{anthropic2024claude}, Emu3~\cite{wang2024emu3}, GPT-4o~\cite{openai2024gpt4o}, OmniGen~\cite{xiao2024omnigen}, DALL-E-2~\cite{openai2024dalle}, DALL-E-3~\cite{openai2024dalle3}, CogVideoX\cite{Yang2024CogVideoXTD},  HermesFlow~\cite{hermesflow2024}, Qwen2.5-VL-Instruct~\cite{wang2024qwen2}, Show-o~\cite{xie2024show}, VILA-U~\cite{wu2024vila}, GILL~\cite{koh2023generating}, Anole~\cite{chern2024anole}, MIO-Instruct~\cite{wang2024mio}, SEED-LLaMA~\cite{ge2023making}, MiniGPT-5~\cite{zheng2023minigpt5}. Among the baselines, Chat-UniVi, Gemini2.0-flash, Claude-3.5-sonnet, GPT-4o\footnote{ Currently, the image generation API for GPT-4o is not yet available. We will incorporateit into our evaluation as soon as it becomes accessible.}, OmniGen, DALL-E-2, DALL-E-3 are specialized understanding models or generative models. Notably, GILL, Anole, MIO-Instruct, SEED-14B, MiniGPT-5 and Gemini2.0-flash-exp can generate interleaved images and texts. Some MLLMs also can generate arbitrarily interlaced modalities, but they are not available as open-source code or model weights yet, such as PUMA~\cite{fang2024puma}, VITRON~\cite{fei2024vitron} and TextHarmony~\cite{Zhao2024HarmonizingVT}.

\begin{table*}[htbp]
\centering
\renewcommand{\arraystretch}{1.5} 
\setlength{\tabcolsep}{3pt} 
\resizebox{1\textwidth}{!}{
\Large 
\begin{tabular}{l|cccc|cccc|cccc|cccc|ccccc|cc}
\hline
\textbf{Method} & \multicolumn{4}{c|}{\textbf{IEE}} & \multicolumn{4}{c|}{\textbf{CSQ}} & \multicolumn{4}{c|}{\textbf{AL}} & \multicolumn{4}{c|}{\textbf{SD}} & \multicolumn{5}{c|}{\textbf{VCoT}} & \multicolumn{2}{c}{\textbf{Unify Score}} \\
\hline
\textbf{Metric} & \textbf{Text Acc} & \textbf{Image Acc} & \textbf{Acc} & \textbf{Acc+} & \textbf{Text Acc} & \textbf{Image Acc} & \textbf{Acc} & \textbf{Acc+} & \textbf{Text Acc} & \textbf{Image Acc} & \textbf{Acc} & \textbf{Acc+} & \textbf{Text Acc} & \textbf{Image Acc} & \textbf{Acc} & \textbf{Acc+} & \textbf{Action Acc} & \textbf{Coordinate Acc} & \textbf{Image Acc} & \textbf{Acc} & \textbf{Acc+} & \textbf{Acc} & \textbf{Acc+} \\
\hline
GILL & 21.00 & 27.50 & 24.25 & 8.00 & 14.75 & 27.82 & 21.29 & 4.95 & 7.69 & 9.62 & 8.66 & 1.92 & 0 & 13.33 & 6.67 & 0 & 0.69 & 0 & 5.00 & 1.90 & 0 & 12.55 & 2.98 \\
MiniGPT-5 & 21.50 & 24.00 & 22.80 & 5.00 & 29.70 & 38.56 & 34.13 & 15.81 & 5.66 & 23.08 & 14.37 & 3.84 & 4.00 & 6.00 & 5.00 & 2.00 & 2.08 & 1.25 & 2.92 & 2.08 & 0 & 15.67 & 5.33 \\
MIO-Instruct & 24.12 & 24.19 & 24.16 & 7.00 & 77.00 & 0 & 38.50 & 0 & 17.31 & 0 & 8.66 & 0 & 23.00 & 0 & 11.50 & 0 & 0 & 0 & 0 & 0 & 0 & 16.56 & 1.40 \\
Anole & 17.00 & 20.10 & 18.55 & 3.23 & 70.30 & 49.00 & 59.65 & 38.00 & 15.38 & 13.46 & 14.42 & 3.84 & 17.00 & 13.00 & 15.00 & 2.00 & 3.47 & 0.69 & 7.50 & 3.89 & 0 & 22.30 & 9.17 \\
SEED-LLaMA & 19.00 & 25.00 & 22.00 & 4.50 & 56.44 & 46.53 & 51.49 & 37.62 & 13.46 & 11.54 & 12.50 & 3.84 & 23.00 & 21.00 & 22.00 & 4.00 & 4.17 & 2.64 & 4.03 & 3.61 & 0 & 22.32 & 9.99 \\
Gemini2.0-flash-exp & \underline{33.33} & \underline{43.50} & \underline{38.42} & \underline{11.11} & \underline{83.17} & \underline{63.37} & \underline{74.75} & \underline{66.33} & \underline{59.61} & \underline{34.62} & \underline{47.12} & \underline{30.77} & \underline{28.00} & \underline{24.00} & \underline{26.00} & \underline{5.00} & \underline{17.64} & \underline{10.14} & \underline{9.44} & \underline{12.41} & 0 & \underline{40.74} & \underline{22.64} \\
\hline
\end{tabular}%
}
\caption{\textbf{Comparison of U-MLLMs on various unify tasks and overall unify Score.}}
\label{tab:result_unify}
\end{table*}

\begin{figure*}[t]
  \centering
  \begin{subfigure}[b]{0.33\textwidth}
    \centering
    \includegraphics[width=\textwidth,height=3.8cm]{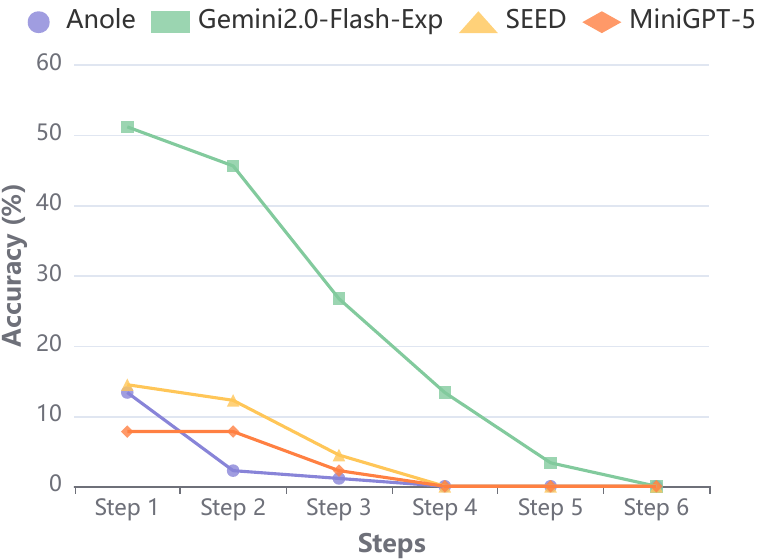}
    \caption{Action Accuracy.}
  \end{subfigure}
  \hfill
  \begin{subfigure}[b]{0.33\textwidth}
    \centering
    \includegraphics[width=\textwidth,height=3.8cm]{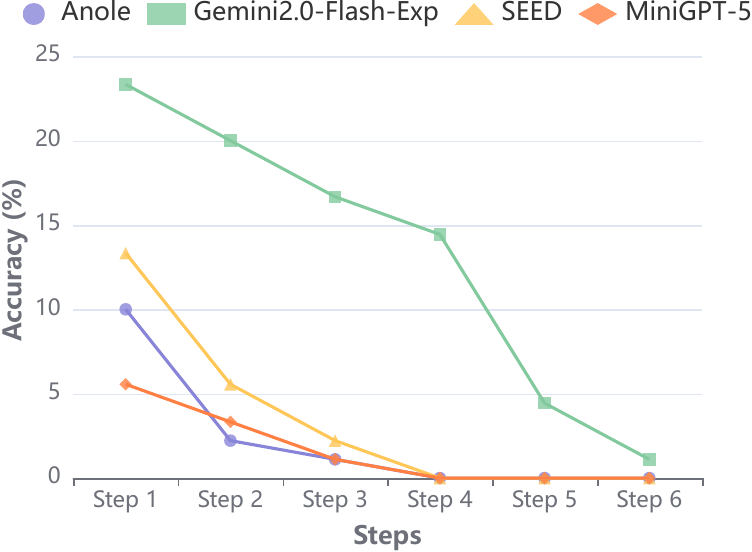}
    \caption{Coordinate Accuracy.}
  \end{subfigure}
  \hfill
  \begin{subfigure}[b]{0.33\textwidth}
    \centering
    \includegraphics[width=\textwidth,height=3.8cm]{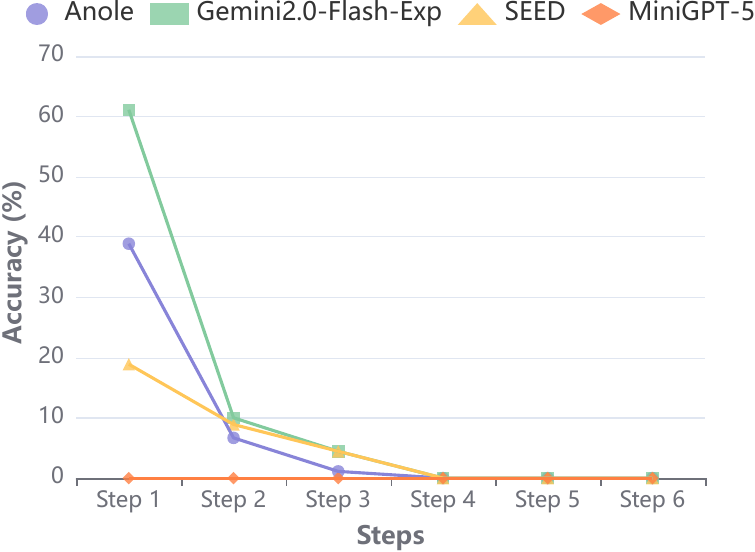}
    \caption{Image Accuracy.}
  \end{subfigure}
  \caption{\textbf{Accuracy distribution across different dimensions on visual cot task.} (a) action, (b) location, and (c) image.}
  \label{fig:vcot}
\end{figure*}

\subsection{Results}
The evaluation results of various MLLMs in MME-U, as shown in Table~\ref{tab:main_result}, indicate that Gemini2.0-flash-exp achieves the highest MME-U score at 45.57. Although compared to MIO-Instruct it does not encompass all subtasks, it demonstrates very balanced performance across understanding, generation, and unify tasks, unlike other models that may exhibit deficiencies in certain test dimensions. It is evident that, compared to traditional MLLMs or generative models, U-MLLMs are capable of handling a wider range of tasks, including more complex image-text interleaved reasoning. However, overall, the development of U-MLLMs is still in its early stages, and even the best-performing models only achieve scores of around 40 on MME-U. Next, we will provide a separate analysis of understanding, generation, and unify tasks.

\textbf{Understanding.} It is evident that Gemini2.0-flash-exp\cite{deepmind2024gemini} demonstrates the best understanding capability among U-MLLMs, while also being a closed-source model. For open-source models, the two U-MLLMs with the best understanding capabilities are Janus-Flow~\cite{ma2024janusflow} and Janus-Pro~\cite{chen2025janus}. These models utilize two separate vision encoders to handle generation and understanding tasks independently, thus overcoming the limitations of tokenizers like VQGAN~\cite{yu2021vector}, which are not well-suited for extracting image understanding features. In contrast, models like Emu3~\cite{wang2024emu3} and Show-o~\cite{xie2024show}, which use a single tokenizer for all image tasks, perform poorly on understanding tasks and still show a significant performance gap compared to currently available open-source MLLMs of similar size. However, our experiments also show that models like Janus-Pro perform poorly on generation tasks. They even fail to support multimodal generation, scoring zero on unified tasks. Therefore, how to strike a balance between understanding and generation capabilities, or whether the two capabilities can indeed complement each other, remains an open question. We also see potential in bridging this gap in understanding capabilities by leveraging existing U-MLLMs alongside strong MLLM baselines. For instance, MIO-instruct~\cite{wang2024mio} achieves impressive understanding results through extensive training data, including video, audio, image-text pairs, and a complex three-stage training pipeline. This suggests that U-MLLMs may require a broader variety or larger volume of data for training.

\begin{figure*}[t]
  \centering
  \begin{subfigure}[b]{0.24\textwidth}
    \centering
    \includegraphics[width=\textwidth,height=3.75cm]{ICCV2025-Author-Kit/MME-U_Figures/t2i/t2i_gt.pdf}
    \caption{Ground Truth.}
  \end{subfigure}
  \begin{subfigure}[b]{0.24\textwidth}
    \centering
    \includegraphics[width=\textwidth,height=3.75cm]{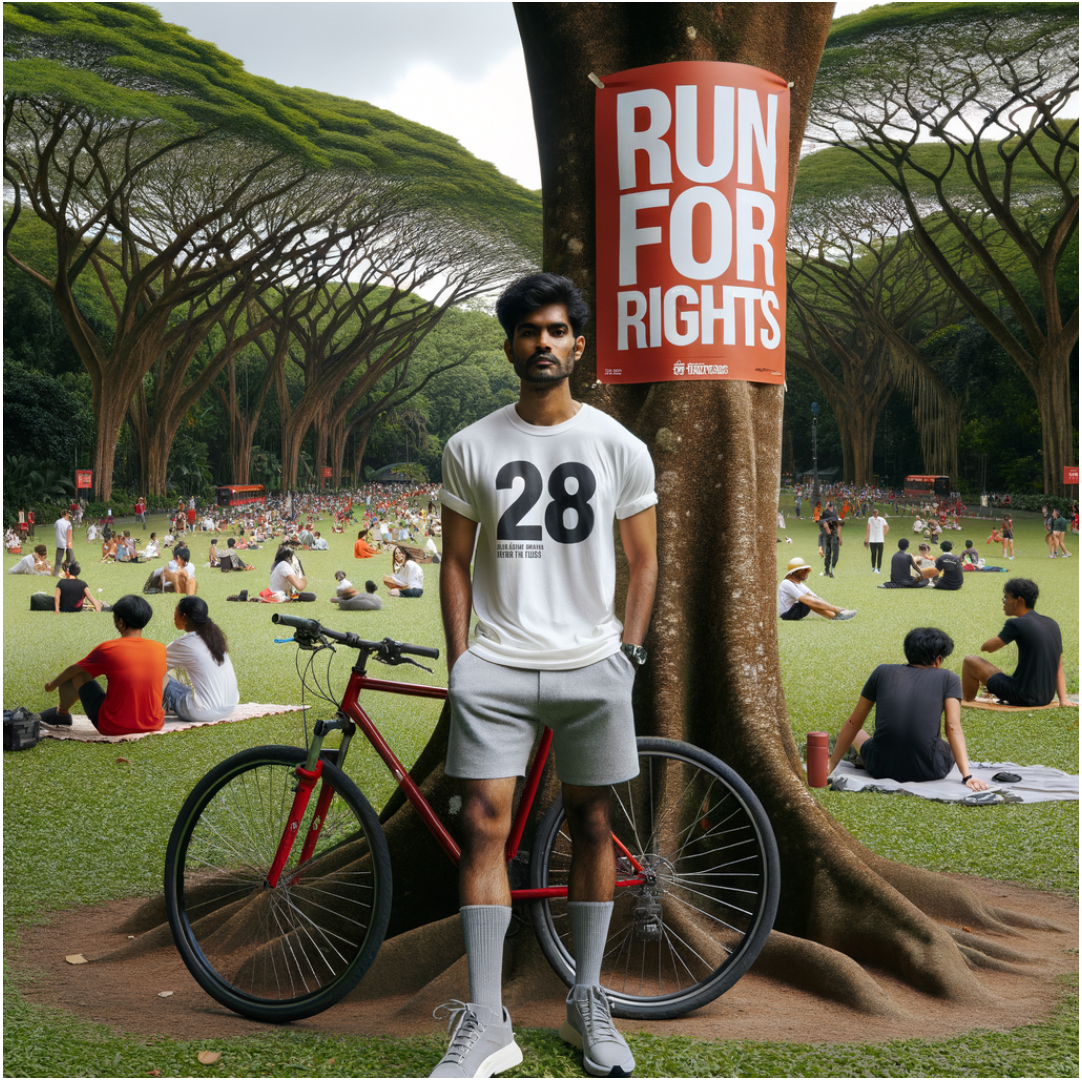}
    \caption{DALLE-2.}
  \end{subfigure}
  \hfill
  \begin{subfigure}[b]{0.24\textwidth}
    \centering
    \includegraphics[width=\textwidth,height=3.75cm]{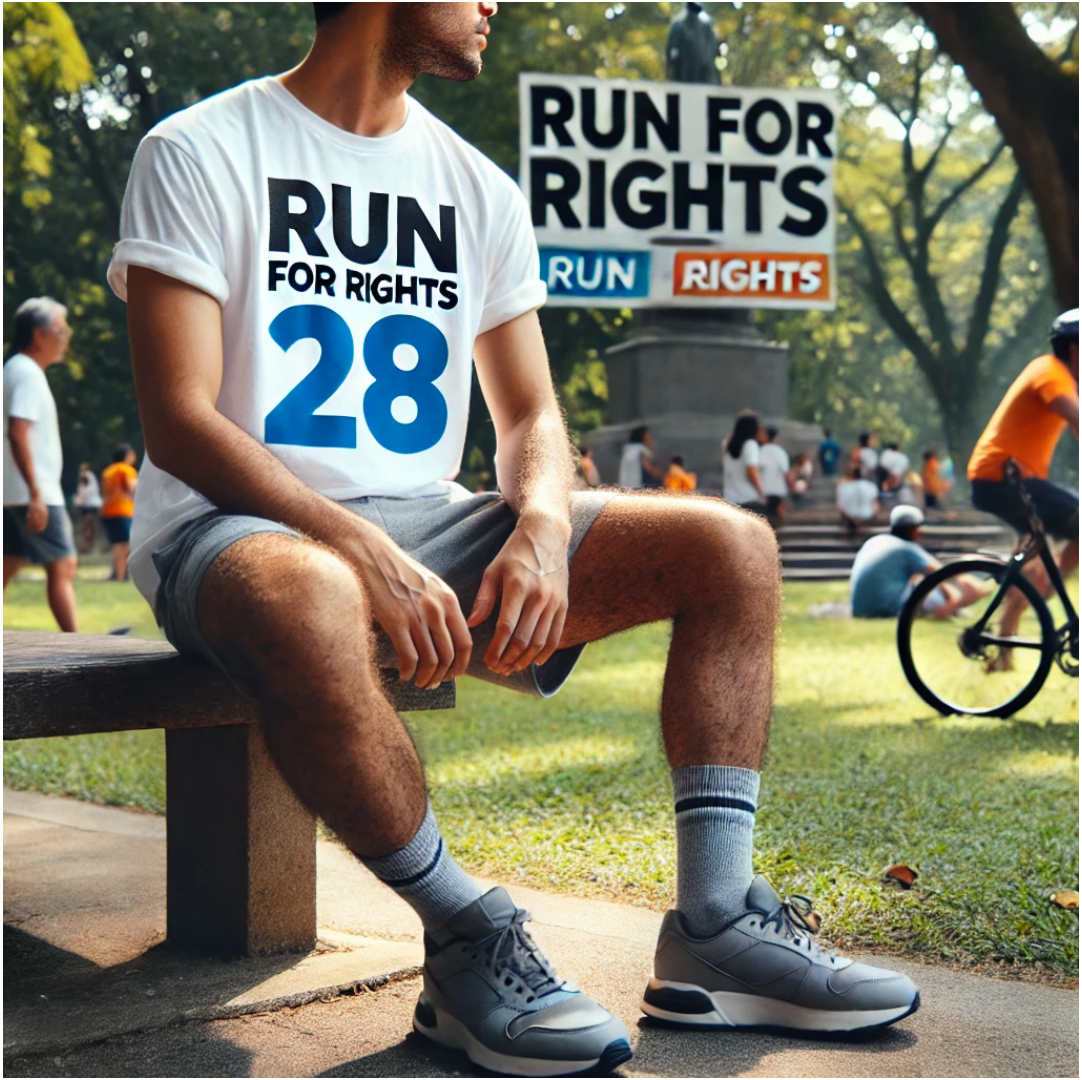}
    \caption{DALLE-3.}
  \end{subfigure}
  \hfill
  \begin{subfigure}[b]{0.24\textwidth}
    \centering
    \includegraphics[width=\textwidth,height=3.75cm]{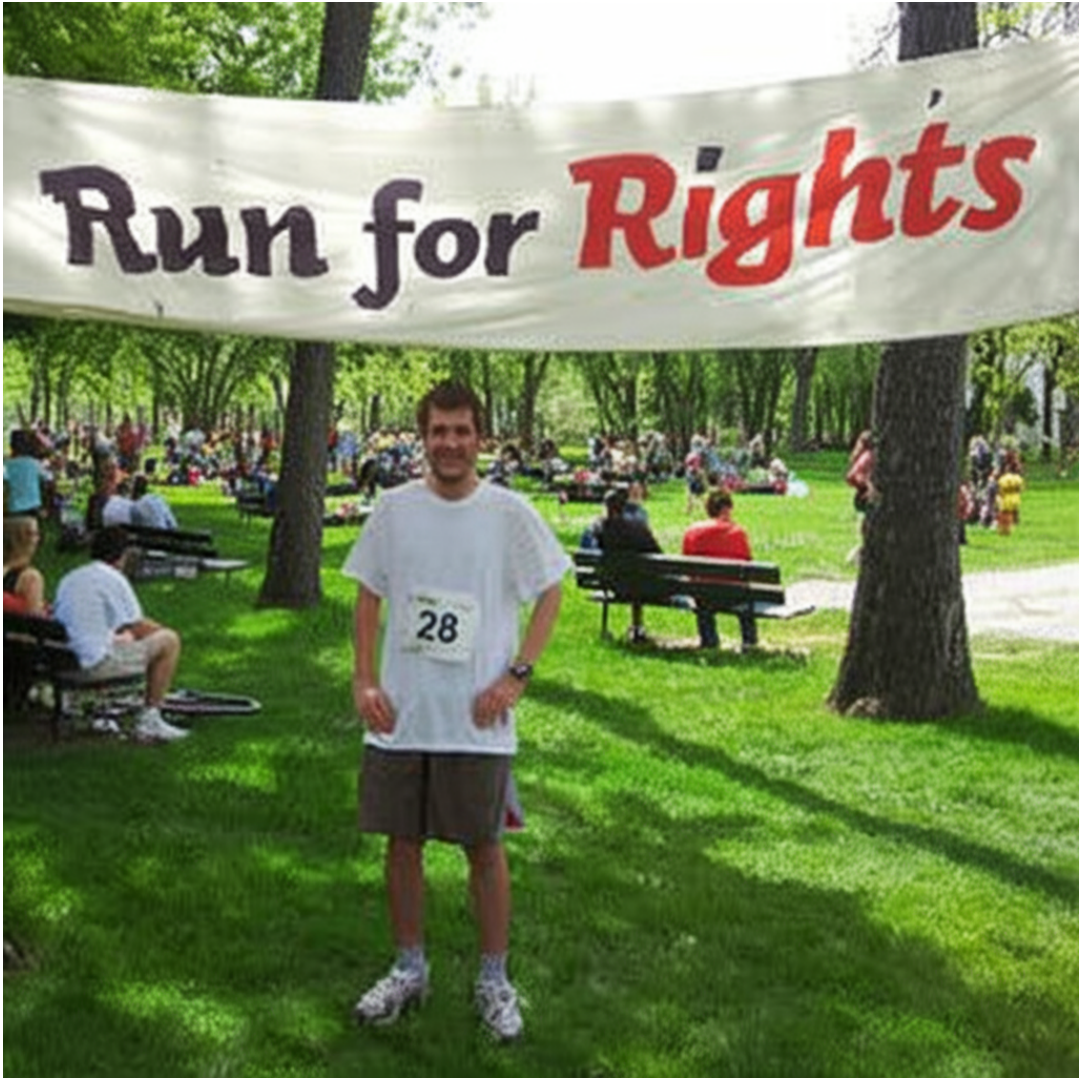}
    \caption{Gemini2.0-flash-exp.}
  \end{subfigure}
  
  \hfill
\begin{subfigure}[b]{0.24\textwidth}
    \centering
    \includegraphics[width=\textwidth,height=3.75cm]{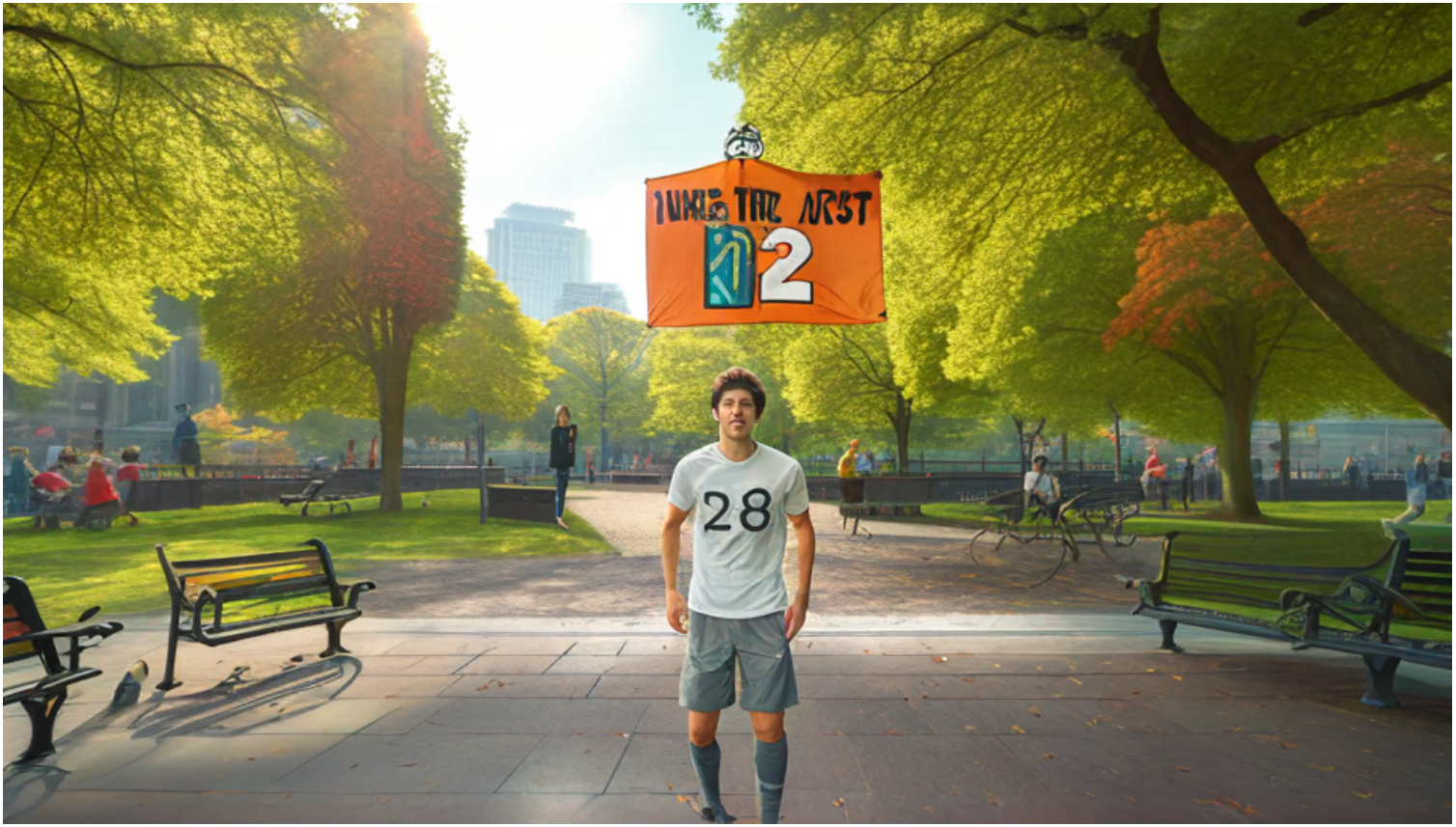}
    \caption{EMU3.}
  \end{subfigure}
    \begin{subfigure}[b]{0.24\textwidth}
    \centering
    \includegraphics[width=\textwidth,height=3.75cm]{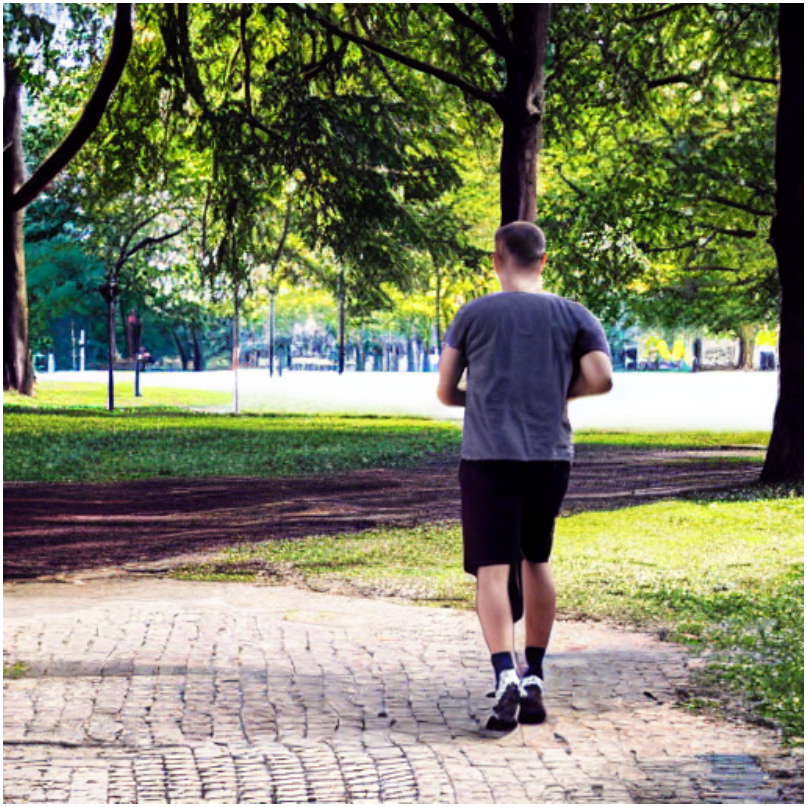}
    \caption{GILL.}
  \end{subfigure}
  \hfill
  \begin{subfigure}[b]{0.24\textwidth}
    \centering
    \includegraphics[width=\textwidth,height=3.75cm]{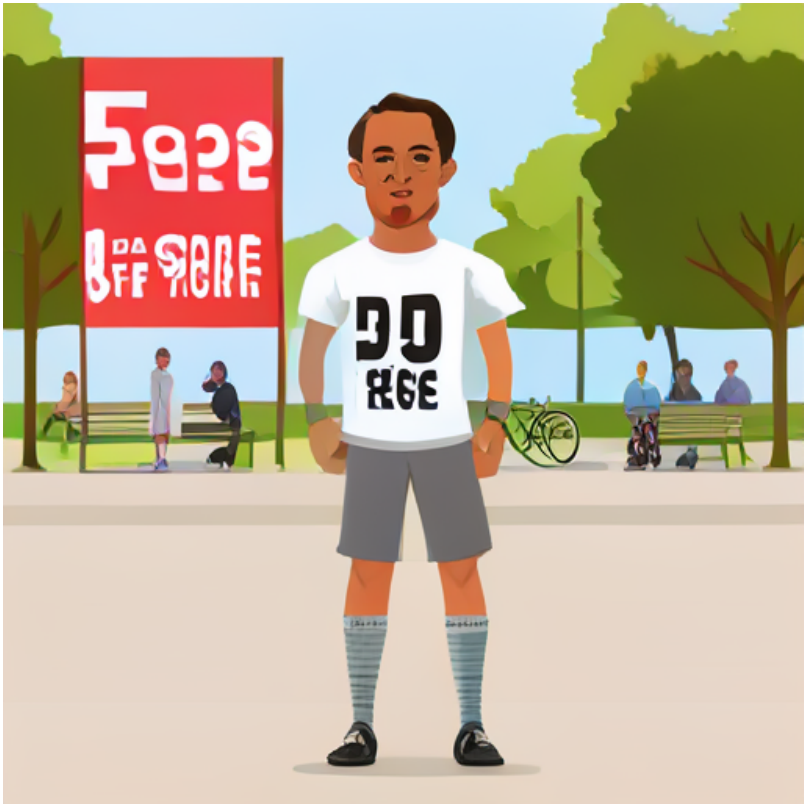}
    \caption{HermesFlow.}
  \end{subfigure}
  \hfill
  \begin{subfigure}[b]{0.24\textwidth}
    \centering
    \includegraphics[width=\textwidth,height=3.75cm]{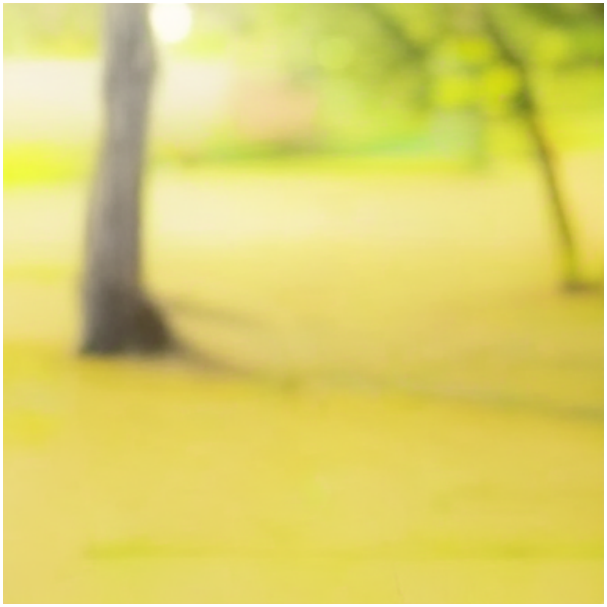}
    \caption{Janus-Pro.}
  \end{subfigure}
  
  \hfill

    \begin{subfigure}[b]{0.24\textwidth}
    \centering
    \includegraphics[width=\textwidth,height=3.75cm]{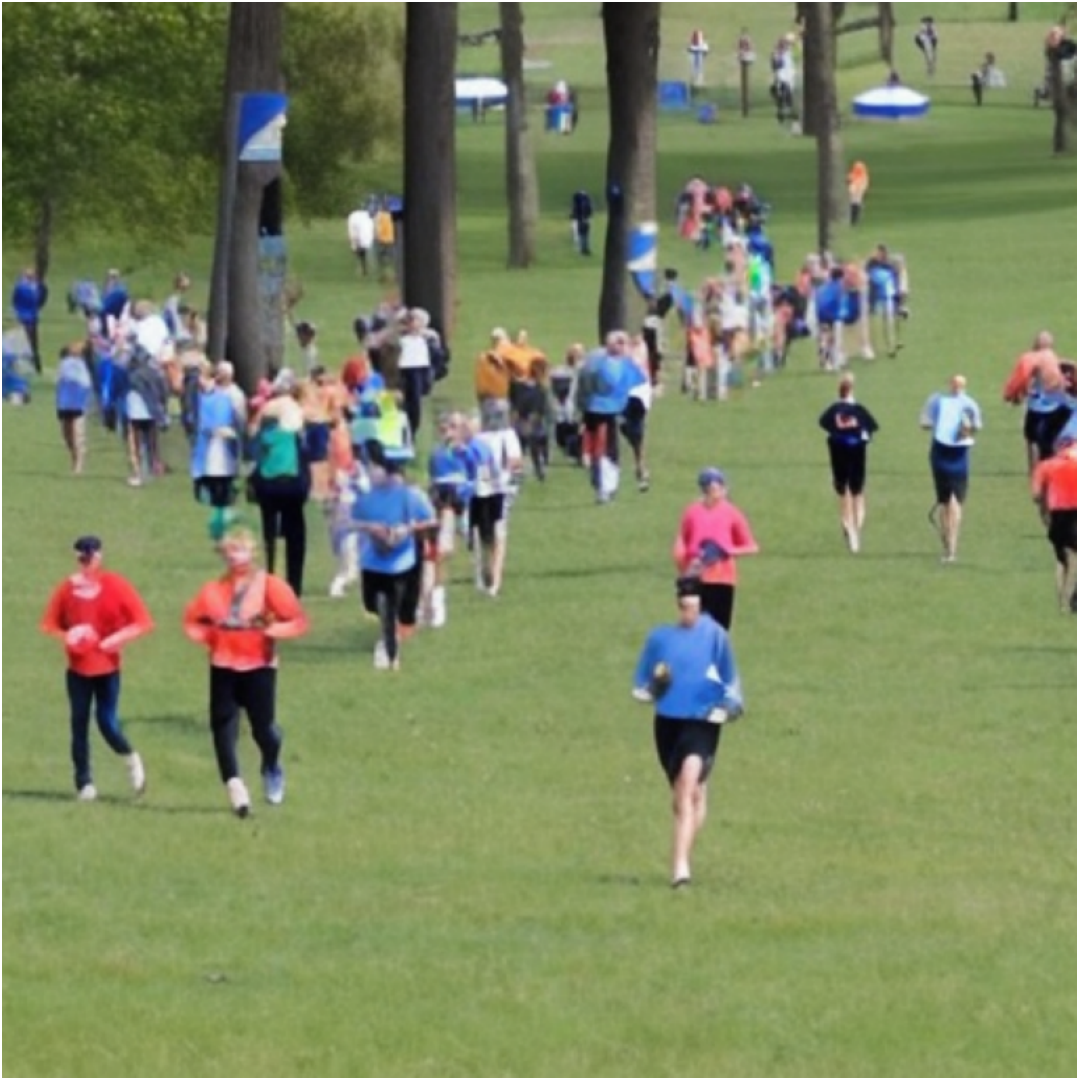}
    \caption{MiniGPT-5.}
  \end{subfigure}
    \begin{subfigure}[b]{0.24\textwidth}
    \centering
    \includegraphics[width=\textwidth,height=3.75cm]{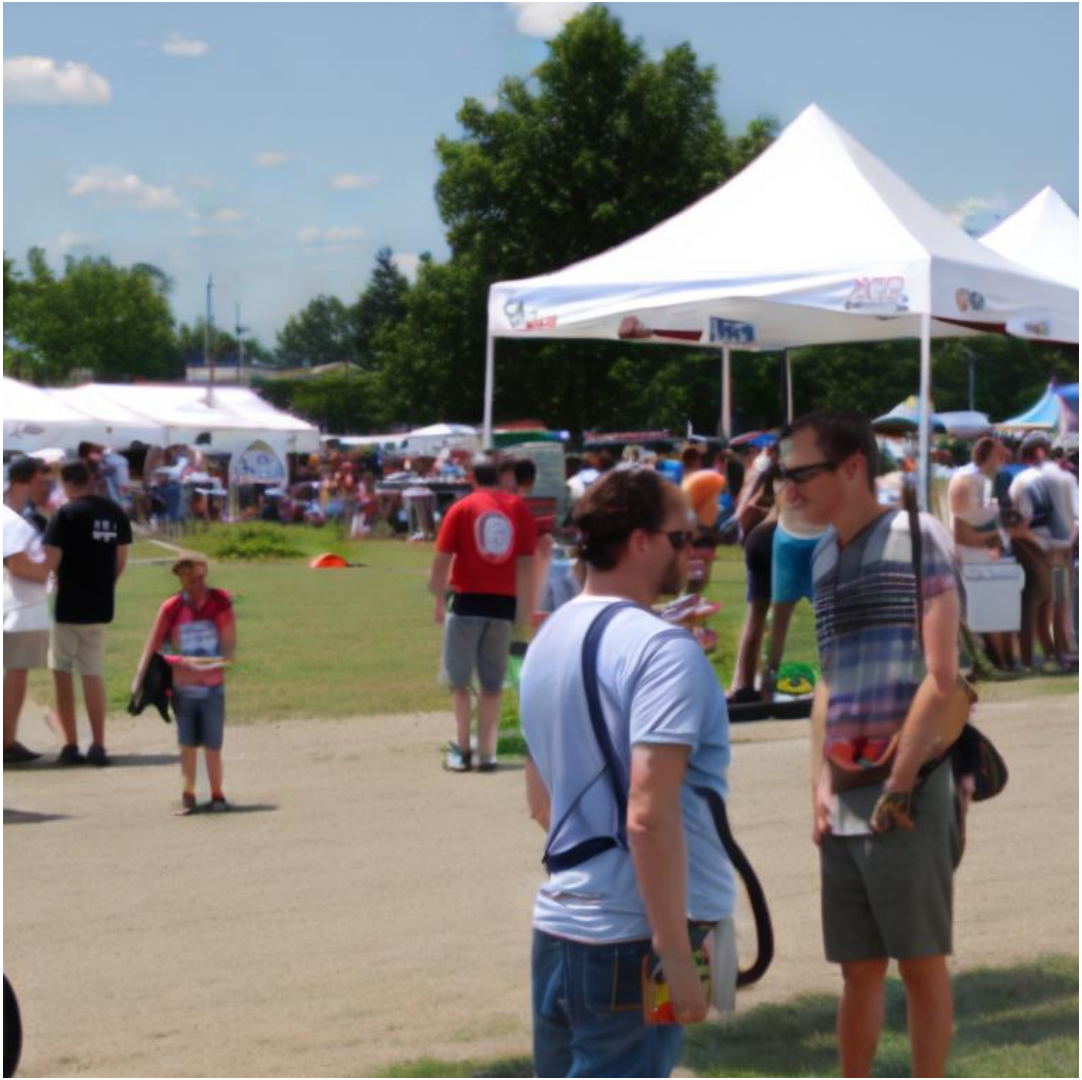}
    \caption{MIO-Instruct.}
  \end{subfigure}
  \hfill
  \begin{subfigure}[b]{0.24\textwidth}
    \centering
    \includegraphics[width=\textwidth,height=3.75cm]{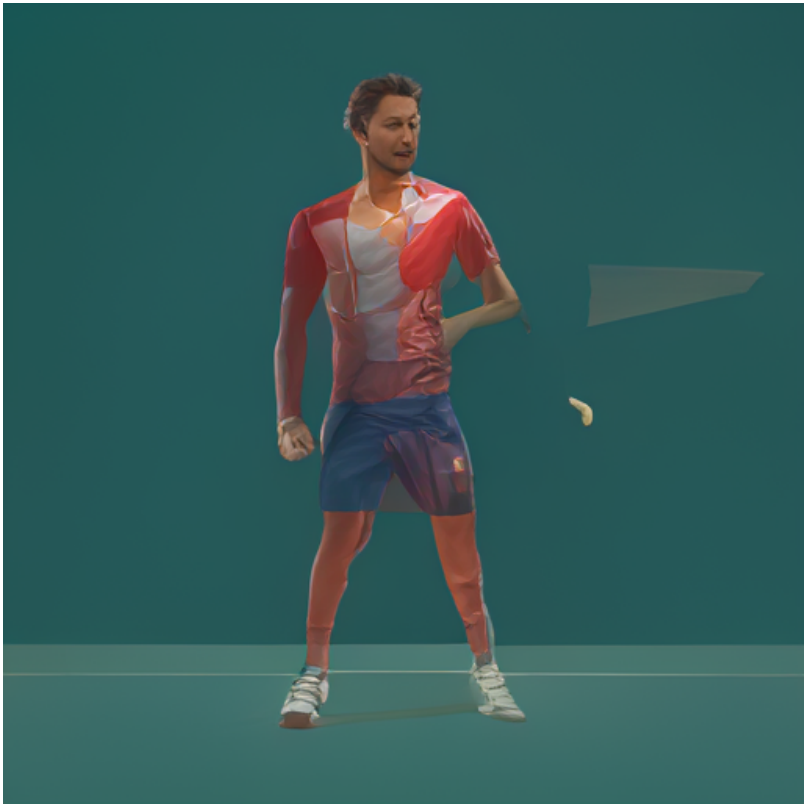}
    \caption{Show-O.}
  \end{subfigure}
  \hfill
  \begin{subfigure}[b]{0.24\textwidth}
    \centering
    \includegraphics[width=\textwidth,height=3.75cm]{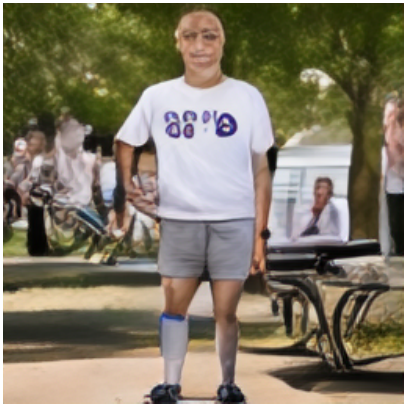}
    \caption{Vila-u}
  \end{subfigure}
  \caption{\textbf{The generated results from various models in the text-to-image generation task}, based on the following text prompt: \textit{A man is standing in a park with a 'Run for Rights' banner in the background. He is wearing a white t-shirt with the number 28 on it, grey shorts, and grey socks with black shoes. The park is filled with people, some sitting on benches, and there is a bicycle leaning against a tree}.}
  \label{fig:t2i_results}
\end{figure*}

\textbf{Generation.}
We compare the performance gap between various U-MLLMs and current state-of-the-art generative models such as DALLE-3. It is evident that, compared to understanding capabilities, the gap in generation tasks is not as significant. For the simplest TIG task, Gemini-2.0-flash-exp even outperforms the best generative model DALLE-3 by six points, while U-MLLMs such as EMU3, HermesFlow, and GILL all achieve an average score above 48. However, it is clear that most U-MLLMs still do not perform well on video generation tasks. Notably, although the original paper for Emu3 mentions its capability for video generation, the corresponding checkpoints have not been released. It's clear that the open-source community still has a long way to go before U-MLLMs that support video generation become widely available. Detailed results on the generation tasks can be found in Table~\ref{tab:main_generation}. In Figure~\ref{fig:t2i_results}, we showcase the generation results from various models using the following text prompt: ``A man is standing in a park with a 'Run for Rights' banner in the background. He is wearing a white t-shirt with the number 28 on it, grey shorts, and grey socks with black shoes. The park is filled with people, some sitting on benches, and there is a bicycle leaning against a tree.'' It is evident that most generated images, such as those from VILA-U, Show-O, and Janus-Pro, fail to capture key details from the caption, such as the number on the jersey or specific text. In contrast, the results from EMU3 more closely resemble the textual description, while MIO-Instruct's outputs are more aligned with realistic scenes (we hypothesize this is because MIO-Instruct was trained on a large amount of real-world data, enhancing its ability to generate lifelike images). However, when it comes to image detail, current open-source U-MLLMs still lag significantly behind dedicated generative models.

\textbf{Unify Capability.}
Our systematic unify task testing shows that, while U-MLLMs have indeed expanded the potential for such tasks compared to traditional understanding/generation models, their performance remains insufficient. For each unify task in Table~\ref{tab:main_result}, we require the models to generate the correct image and perform correct reasoning. Under these conditions, even for simple tasks such as answering common questions and generating images, the best open-sourced model (Anole) only achieves an accuracy of 59.65\% and accuracy-plus of 38\% (Table~\ref{tab:result_unify}). In other tasks, no open-sourced model is able to surpass the 30\% accuracy. It is worth noting that models perform even worse on tasks like Visual CoT, which require multi-step image generation and reasoning. No model is able to successfully complete tasks involving multiple steps. This finding underscores the importance of our MME-U, as relying solely on case studies to demonstrate a model's mix-modality generation capabilities is clearly insufficient. We will further analyze these models' performance, weaknesses, and provide examples in the analysis section.

\subsection{Analysis and Findings}
\textbf{Trade-off Between Basic and Unified Capabilities.} The experimental results reveal that current U-MLLMs face a significant challenge in balancing their fundamental abilities—such as understanding and generating performance—with the demands of unified tasks that require integrating multiple modalities. For instance, models like GILL, Anole, and MiniGPT-5 are designed to handle unified tasks but tend to exhibit relatively poor performance on basic tasks, which results in lower overall scores when compared to some non-unified MLLMs. On the other hand, while MIO-Instruct demonstrates high performance in basic understanding and generation, its capability to interleave image and text generation effectively is notably deficient. This imbalance suggests that the current training paradigms may not be adequately aligning the learning objectives for basic and unified capabilities within a single framework.

\textbf{Detailed Analysis of Model Performance on Unify Tasks.} In Table~\ref{tab:result_unify}, we provide a detailed analysis of different models' performance on unify tasks, focusing on text reasoning accuracy and image generation accuracy. It is clear that MIO-Instruct exhibits stronger understanding capabilities than generation abilities (as confirmed by the results in Table~\ref{tab:main_result}). As a result, many of its tasks show high text reasoning performance, particularly in commonsense QA, where its text reasoning accuracy reaches 76.24\%. However, it fails to generate a correct image, completely missing the potential for mutual reinforcement between generation and understanding. In contrast, other models show comparable performance in both text reasoning and image reasoning evaluation criteria, but their overall results are not impressive. Notably, for visual CoT tasks, despite our efforts to simplify the questions into multiple-choice format, none of the models have been able to correctly complete multi-step reasoning and generation tasks.

\textbf{Poor Instruction Following Ability for Image Generation.} There are two main issues with the current models in image generation: \textit{1. Uncontrolled Style Generation.} In Figure~\ref{fig:vcot_fig}, we present the intermediate state images generated by different models in the VCoT task. Only the Anole and Gemini2.0-flash-exp models are able to generate images with a style similar to the initial image. In contrast, other models produce images with a clear style bias, which do not align well with our state diagrams.
\textit{2. Difficulty Understanding Complex Instructions.} Many models, such as MIO-Instruct, struggle with following complex instructions, such as generating auxiliary lines based on the original question. These models fail to generate images with auxiliary lines, often requiring multiple attempts to generate a relevant image, and the resulting images often bear little resemblance to the original reference. However, for simpler instructions, like generating an image of a dog, these models are able to execute the task correctly.

\textbf{Inadequate Visual CoT Capability in Unified Models.} In Figure~\ref{fig:vcot}, we further illustrate the challenges of the Visual CoT task. The accuracy of U-MLLMs declines as the number of steps in the VCoT task increases. Errors made in earlier steps compound over time, making it increasingly difficult for models to generate correct actions, coordinates, and images. This cascading error effect highlights a fundamental limitation in maintaining consistent reasoning across multi-step tasks. At the same time, this example further emphasizes the high requirements of our unify tasks for both generation and understanding capabilities. For instance, although Anole demonstrates relatively strong image accuracy in Figure~\ref{fig:vcot}, its weaker understanding abilities result in less effective action selection. This ultimately leads to worse final results compared to the other two baselines.

Due to space limitations, we have included additional in-depth analyses in Appendix~\ref{sec:app_exp}, which contains detailed visualizations of the U-MLLMs' generation results, as well as specific examples from the unify tasks.

\section{Conclusion and Limitation}
The MME-U benchmark framework presented here serves as a foundational step towards evaluating U-MLLMs on a diverse array of tasks encompassing multimodal understanding, generation, and their integration. This benchmark reveals the current landscape of U-MLLMs, highlighting their capabilities and areas for improvement. While these models demonstrate proficiency in handling various multimodal tasks, they struggle with balancing understanding and generation, handling complex instructions, and performing well on unify tasks. Moreover, current U-MLLMs exhibit significant inconsistencies in aligning textual instructions with their visual outputs, highlighting the need for further research to improve multimodal reasoning and generation integration. However, this study simplifies the evaluation of unify tasks by framing image generation as multiple-choice questions, which may allow model ``hacking''. For instance, SEED-generated images may not meet style standards but achieve high similarity scores, inflating accuracy metrics. Future work will incorporate MLLM or CLIP scores for stricter evaluation. 


%% file: sec/X_suppl.tex
\clearpage
\setcounter{page}{1}
\maketitlesupplementary

\section{Related Works}
\label{sec:formatting}
\textbf{Unified Multimodal Large Language Models.} Building on the success of MLLMs \cite{wang2024qwen2,fu2025vita,yu2025aligning,zhang2024beyond}, recent studies U-MLLMs, which can understand and generate multiple modalities in an end-to-end manner. Some approaches have adopted a unified training objective, projecting both text and images into a discrete token space and employing a next-token prediction loss function for training~\cite{wang2024mio,wu2024vila,team2024chameleon}. This training method and framework are notably straightforward. However, using discrete image tokens (e.g., extracted from VQVAE image features) may not be optimal for image understanding tasks. Therefore, works like Janus-Flow~\cite{chen2025janus}, Janus-Pro~\cite{chen2025janus}, among others, have employed different vision encoders such as VQVAE for image generation and SigLIP for image comprehension, significantly enhancing the understanding capabilities of U-MLLMs. Additionally, other methods have found that diffusion training is more suitable for image generation. Thus, adopting diffusion-based training for image generation and next-token prediction for text generation aims to strengthen the image generation capabilities further~\cite{xie2024show,zhou2024transfusion}. Recent research has also explored fine-tuning U-MLLMs to further enhance their performance on unified tasks~\cite{li2025imagine}. However, despite the rapid advancements of U-MLLMs, there remains a lack of comprehensive benchmarks for systematically and fairly evaluating their capabilities in understanding, generation, and multimodal synthesis tasks.

\textbf{Benchmarks for Understanding.}
With the rapid development of MLLMs, several concurrent works~\cite{fu2024mme} have proposed various benchmarks to evaluate the models' capabilities in multimodal comprehension tasks, such as single-image perception and understanding~\cite{fu2023mme,zhang2024mme} (e.g., MME series), interleaved image \& text understanding, and video understanding~\cite{fang2025mmbench} (e.g., MMBench-Video, Video-MME). Additionally, some benchmarks focus on multimodal safety~\cite{zhang2025mm} or mathematical reasoning~\cite{yan2024errorradar}. These benchmarks differ in coverage and metrics.

\textbf{Benchmarks for Generation.} 
Various benchmarks have been proposed to assess multi-modal generation capabilities ~\cite{wang2024tip, sheynin2024emu, 5206848, xu2016msr, ku2024imagenhub, li2023seed2}, including tasks like image reconstruction~\cite{5206848}, image editing~\cite{sheynin2024emu, ku2024imagenhub}, and conditional image \& video generation~\cite{wang2024tip, Lin2014MicrosoftCC}. However, these benchmarks mainly focus on individual tasks within single modalities, failing to capture the full scope of multi-modal comprehension and generation. While some benchmarks, such as SEED-Bench-2~\cite{li2023seed2}, provide hierarchical evaluation for both understanding and generation, they do not assess unified tasks, and the range of tasks is limited.

\begin{figure*}  
\centering
\includegraphics[width=1.0\linewidth]{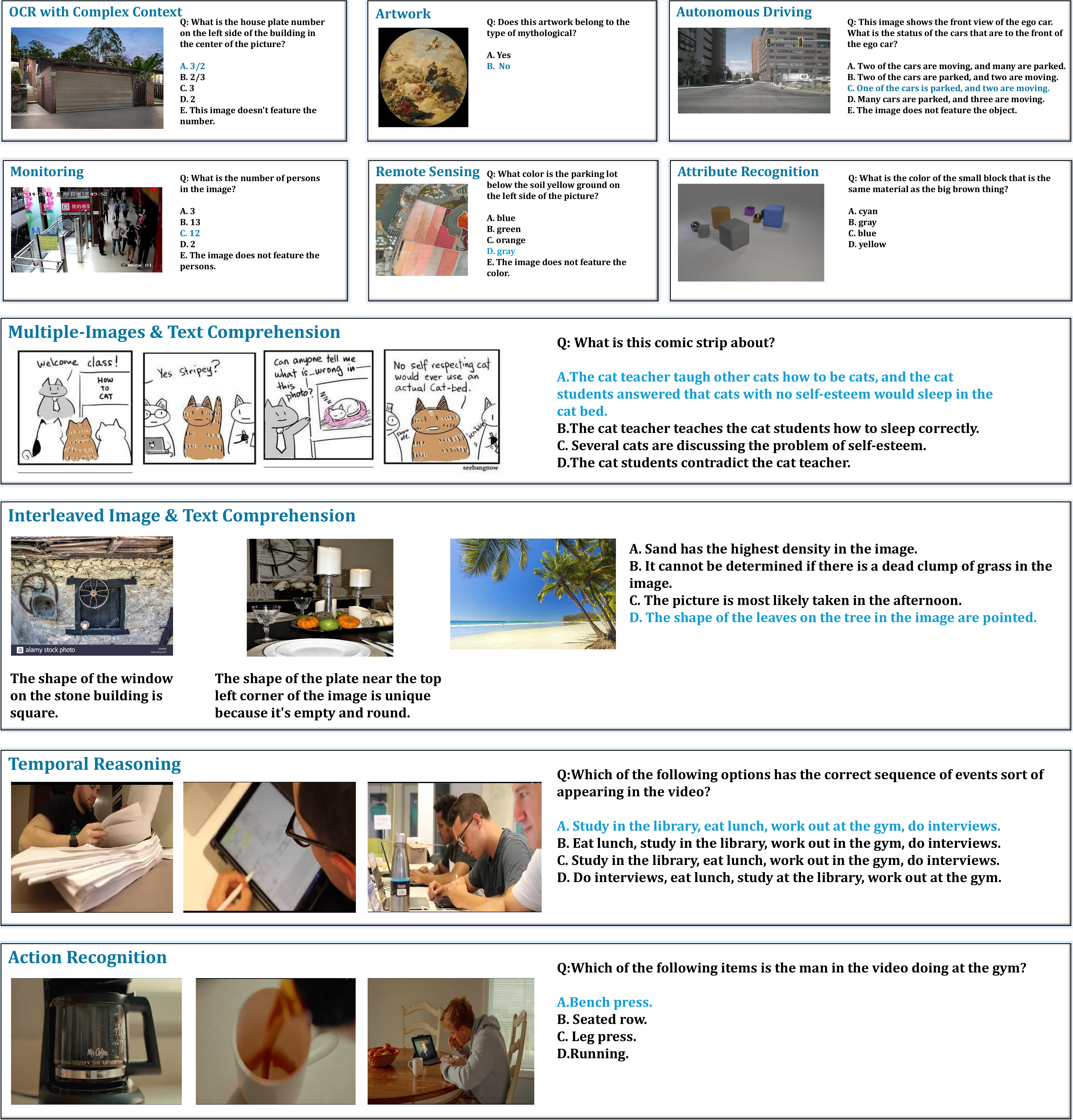}
\caption{Data samples from understanding task, which includes single-image perception and reasoning, multi-image and image-text interlaced perception and reasoning, video perception and reasoning, etc.}
\label{fig:Visualization_Understanding}
\end{figure*}

\begin{figure*}  
\centering
\includegraphics[width=1.0\linewidth]{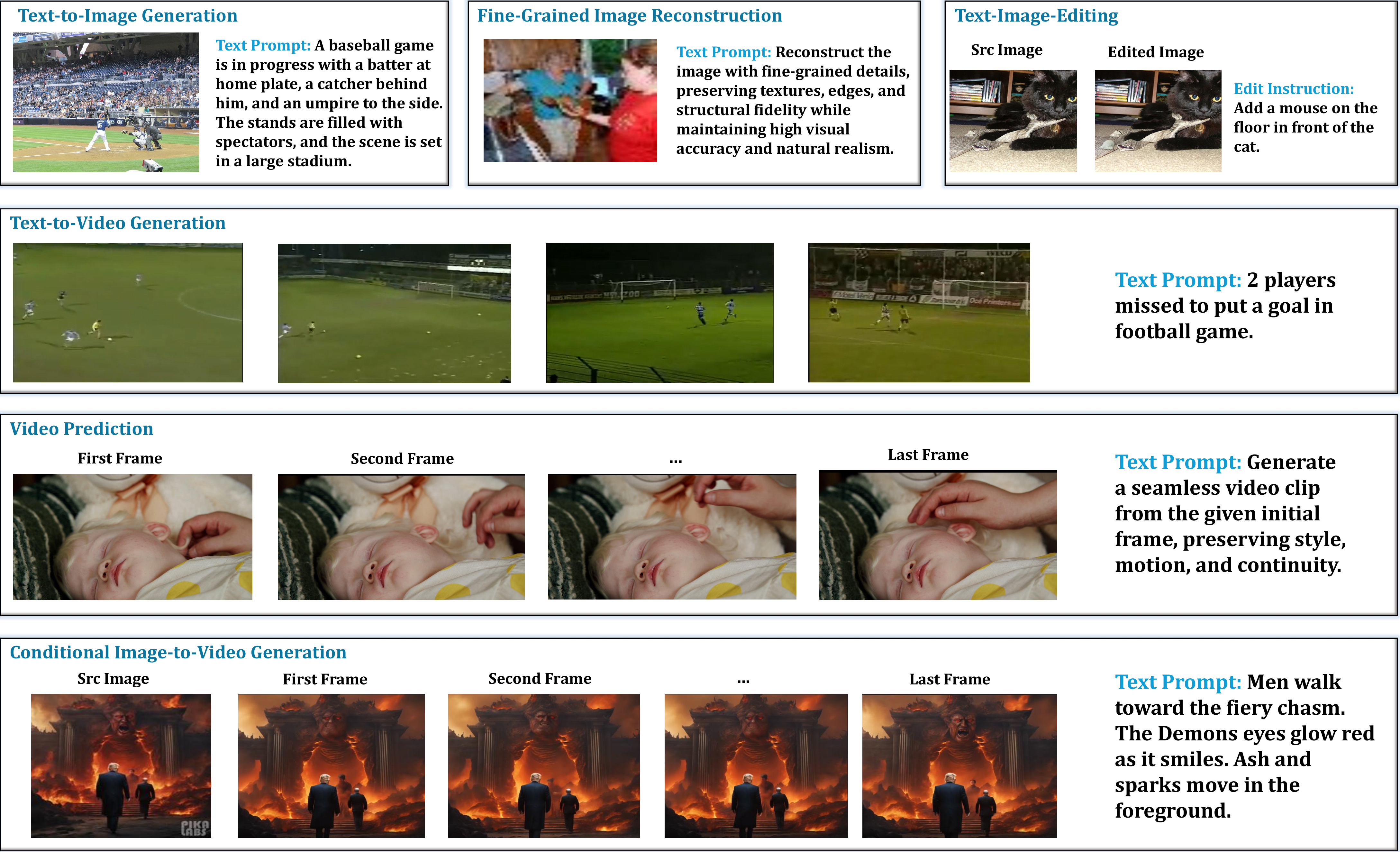}
\caption{Data samples from generation task. It includes subtasks such as Text-to-Image Generation, Text-to-Image Editing, Fine-Grained Image Reconstruction, Text-to-Video Generation, conditional Image-to-Video Generation, and Video Prediction.}
\label{fig:Visualization_Generation}
\end{figure*}

\section{Evaluation Metrics}\label{sec:eval_met}
\subsection{Understanding Score}
Let the three subtasks in the Understanding Task be formally defined as follows:
\[
T = \{\text{SIPU}, \text{MITIU}, \text{VPU}\}.
\]

For each subtask \(t \in T\), let \(Q_t\) represent the set of multiple-choice questions, where each question \(q \in Q_t\) has exactly one correct answer. To evaluate correctness, we define the indicator function for each question as follows:
\[
\mathbb{I}_t(q) =
\begin{cases}
1, & \text{if the selected answer for } q \text{ is correct}, \\
0, & \text{otherwise}.
\end{cases}
\]

The accuracy for subtask \(t\) is given by:
\[
\text{acc}_t = \frac{1}{\lvert Q_t \rvert} \sum_{q \in Q_t} \mathbb{I}_t(q).
\]

Since equal weights are assigned to each subtask, the Understanding Score (US) is computed as the arithmetic mean of the accuracies across all subtasks:
\[
\text{US} = \frac{1}{3} \sum_{t \in T} \text{score}_t, \quad T = \{\text{SIPU}, \text{MITIU}, \text{VPU}\}.
\]

\subsection{Generation Score}

The generative task comprises six subtasks:
\[
T = \{\text{CIVG}, \text{TVG}, \text{VP}, \text{FIR}, \text{TIE}, \text{TIG}\}.
\]
All metric scores are normalized to the range \([0,100]\).

\textbf{Normalization of FVD and FID Scores.} Let \( s \) denote the raw FVD or FID value for a sample, where \( s \in [1,1000] \) and lower values indicate better performance. The normalized score \( S \) is computed as:
\[
S = 100\left(1 - \frac{s-1}{1000-1}\right) = 100\left(1 - \frac{s-1}{999}\right).
\]
This ensures:
\begin{itemize}
    \item \( S = 100 \) when \( s = 1 \) (best performance),
    \item \( S = 0 \) when \( s = 1000 \) (worst performance).
\end{itemize}
If all raw scores across models are identical, each normalized score is set to 100 to maintain consistency in evaluation and prevent division by zero in the normalization process.

\textbf{Score Calculation for CIVG and TVG.} The subtask score for \( t \in \{\text{CIVG}, \text{TVG}\} \) is given by:
\[
\text{score}_{t} = \frac{\text{FVD}_{\text{norm}}^{(t)} + \text{FID}_{\text{norm}}^{(t)} + \text{CLIPSIM}^{(t)}}{3}.
\]

\textbf{Score Calculation for VP.} The VP subtask score is determined using the following formula:
\[
\text{score}_{\text{VP}} = \frac{\text{FVD}_{\text{norm}}^{(\text{VP})} + \text{FID}_{\text{norm}}^{(\text{VP})}}{2}.
\]

\textbf{Score Calculation for FIR, TIE, and TIG.} For FIR (Fine-Grained Image Reconstruction), the evaluation metric is LPIPS. To ensure higher values indicate better performance, the score is defined as:
\[
\text{score}_{\text{FIR}} = 1 - \text{LPIPS}.
\]

For both TIE (Text Image Editing) and TIG (Text-to-Image Generation), two metrics are used: CLIP-I and CLIP-T. The score for each subtask is computed as the average of these two metrics:
\[
\text{score}_{\text{TIE}} = \frac{\text{CLIP-I}_{\text{TIE}} + \text{CLIP-T}_{\text{TIE}}}{2},
\]
\[
\text{score}_{\text{TIG}} = \frac{\text{CLIP-I}_{\text{TIG}} + \text{CLIP-T}_{\text{TIG}}}{2}.
\]

\textbf{Overall Generation Score.} The overall Generation Score (GS) is the arithmetic mean of all six subtask scores:
\[
\text{GS} = \frac{1}{6} \sum_{t \in T} \text{score}_t, \quad T = \{\text{CIVG}, \text{TVG}, \text{VP}, \text{FIR}, \text{TIE}, \text{TIG}\}.
\]

\subsection{Unify Score}
Let the Unify Task consist of the subtasks 
\[
T = \{\text{IEE}, \text{CSQ}, \text{AL}, \text{SD}, \text{VCoT}\}.
\]
For each subtask \(t \in T\), denote by \(S_t\) the set of samples.
\subsubsection{Subtasks IEE, CSQ, AL, SD}
For a given subtask \(t \in \{\text{IEE}, \text{CSQ}, \text{AL}, \text{SD}\}\) and for each sample \(s \in S_t\), there are two questions:
\begin{enumerate}
    \item A text-based multiple-choice question.
    \item An image-based multiple-choice question.
\end{enumerate}
Define the indicator functions for the text and image responses as follow:
\[
\mathbb{I}_t^{\text{text}}(s) =
\begin{cases}
1, & \text{if the text answer for } s \text{ is correct}, \\
0, & \text{otherwise},
\end{cases}
\]
\[
\mathbb{I}_t^{\text{img}}(s) =
\begin{cases}
1, & \text{if the image answer for } s \text{ is correct}, \\
0, & \text{otherwise}.
\end{cases}
\]
Then, the text accuracy and image accuracy for subtask \(t\) are, respectively,
\[
\text{acc}_t^{\text{text}} = \frac{1}{|S_t|} \sum_{s \in S_t} \mathbb{I}_t^{\text{text}}(s), \quad
\text{acc}_t^{\text{img}} = \frac{1}{|S_t|} \sum_{s \in S_t} \mathbb{I}_t^{\text{img}}(s).
\]
The overall accuracy for subtask \(t\) is then defined as the average of the two:
\[
\text{acc}_t = \frac{\text{acc}_t^{\text{text}} + \text{acc}_t^{\text{img}}}{2}.
\]

Additionally, we define \(\text{acc}_t^{+}\) to represent the accuracy for samples where both the textual and image-based answers are correct:
\[
\text{acc}_t^{+} = \frac{1}{|S_t|} \sum_{s \in S_t} \mathbb{I}_t^{\text{text}}(s) \cdot \mathbb{I}_t^{\text{img}}(s).
\]
\subsubsection{Subtask VCoT}
For the VCoT subtask, each sample \(s \in S_{\text{VCoT}}\) represents a maze navigation task composed of \(K_s\) sequential steps. For each step \(k \in \{1, 2, \dots, K_s\}\), there are multiple-choice questions evaluating the model's prediction of:
\begin{enumerate}
    \item An action.
    \item A coordinate.
    \item An image.
\end{enumerate}

\textbf{Calculation of \(\mathbf{acc}_{\text{VCoT}}\):}
Let \(N_{\text{steps}} = \sum_{s \in S_{\text{VCoT}}} K_s\) be the total number of steps across all samples in the VCoT subtask.
Define the indicator functions for the correctness of action, coordinate, and image predictions for step \(k\) of sample \(s\) as follow:
\[
\mathbb{I}_{\text{VCoT}}^{\text{action}}(s, k) =
\begin{cases}
1, & \text{if the action prediction for step } k \\
   & \text{of sample } s \text{ is correct}, \\
0, & \text{otherwise}.
\end{cases}
\]

\[
\mathbb{I}_{\text{VCoT}}^{\text{coord}}(s, k) =
\begin{cases}
1, & \text{if the coordinate prediction for step } k \\
   & \text{of sample } s \text{ is correct}, \\
0, & \text{otherwise}.
\end{cases}
\]

\[
\mathbb{I}_{\text{VCoT}}^{\text{img}}(s, k) =
\begin{cases}
1, & \text{if the image prediction for step } k \\
   & \text{of sample } s \text{ is correct}, \\
0, & \text{otherwise}.
\end{cases}
\]
Calculate the average accuracy for each prediction type across all steps:
\[
\text{acc}_{\text{VCoT}}^{\text{action}} = \frac{1}{N_{\text{steps}}} \sum_{s \in S_{\text{VCoT}}} \sum_{k=1}^{K_s} \mathbb{I}_{\text{VCoT}}^{\text{action}}(s, k),
\]
\[
\text{acc}_{\text{VCoT}}^{\text{coord}} = \frac{1}{N_{\text{steps}}} \sum_{s \in S_{\text{VCoT}}} \sum_{k=1}^{K_s} \mathbb{I}_{\text{VCoT}}^{\text{coord}}(s, k),
\]
\[
\text{acc}_{\text{VCoT}}^{\text{img}} = \frac{1}{N_{\text{steps}}} \sum_{s \in S_{\text{VCoT}}} \sum_{k=1}^{K_s} \mathbb{I}_{\text{VCoT}}^{\text{img}}(s, k).
\]
The overall \(\mathbf{acc}_{\text{VCoT}}\) metric is the arithmetic mean of these three component accuracies:
\[
\mathbf{acc}_{\text{VCoT}} = \frac{ \text{acc}_{\text{VCoT}}^{\text{action}} + \text{acc}_{\text{VCoT}}^{\text{coord}} + \text{acc}_{\text{VCoT}}^{\text{img}} }{3}.
\]

\textbf{Calculation of \(\mathbf{acc+}_{\text{VCoT}}\):}
Define an indicator function for the full correctness of a single step \(k\) in sample \(s\):
\[
\mathbb{I}_{\text{step\_all\_correct}}(s, k) = \mathbb{I}_{\text{VCoT}}^{\text{action}}(s, k) \times \mathbb{I}_{\text{VCoT}}^{\text{coord}}(s, k) \times \mathbb{I}_{\text{VCoT}}^{\text{img}}(s, k).
\]
This function is 1 if all three predictions for step \(k\) are correct, and 0 otherwise.

Now, define the indicator function for the perfect completion of sample \(s\):
\[
\mathbb{I}_{\text{VCoT}}^{\text{sample\_perfect}}(s) =
\begin{cases}
1, & \text{if } \mathbb{I}_{\text{step\_all\_correct}}(s, k) = 1 \\
   & \text{for all } k \in \{1, 2, \dots, K_s\}, \\
0, & \text{otherwise}.
\end{cases}
\]
The \(\mathbf{acc+}_{\text{VCoT}}\) metric is the proportion of perfectly completed samples:
\[
\mathbf{acc+}_{\text{VCoT}} = \frac{1}{|S_{\text{VCoT}}|} \sum_{s \in S_{\text{VCoT}}} \mathbb{I}_{\text{VCoT}}^{\text{sample\_perfect}}(s).
\]

\subsection*{Unify Scores} 
The \textbf{Unify Score (Unify-S)} is defined as the arithmetic mean of the \(\mathbf{acc}_t\) metrics across all subtasks:
\[
\text{Unify-S} = \frac{1}{|T|} \sum_{t \in T} \mathbf{acc}_t,
\]

\subsection{MME-U Score}

The MME-U Score is computed as the arithmetic mean of the Understanding Score (US), Generation Score (GS), and Unify Score (Unify-S):

\[
\text{MME-U} = \frac{1}{3} \left( \text{US} + \text{GS} + \text{Unify-S} \right).
\]

where:
\begin{itemize}
    \item \(\text{US}\) is the Understanding Score,
    \item \(\text{GS}\) is the Generation Score,
    \item \(\text{Unify-S}\) is the Unify Score.
\end{itemize}

Each component score is calculated as described in their respective sections.

\begin{figure}
  \centering
  \includegraphics[width=1.0\linewidth]{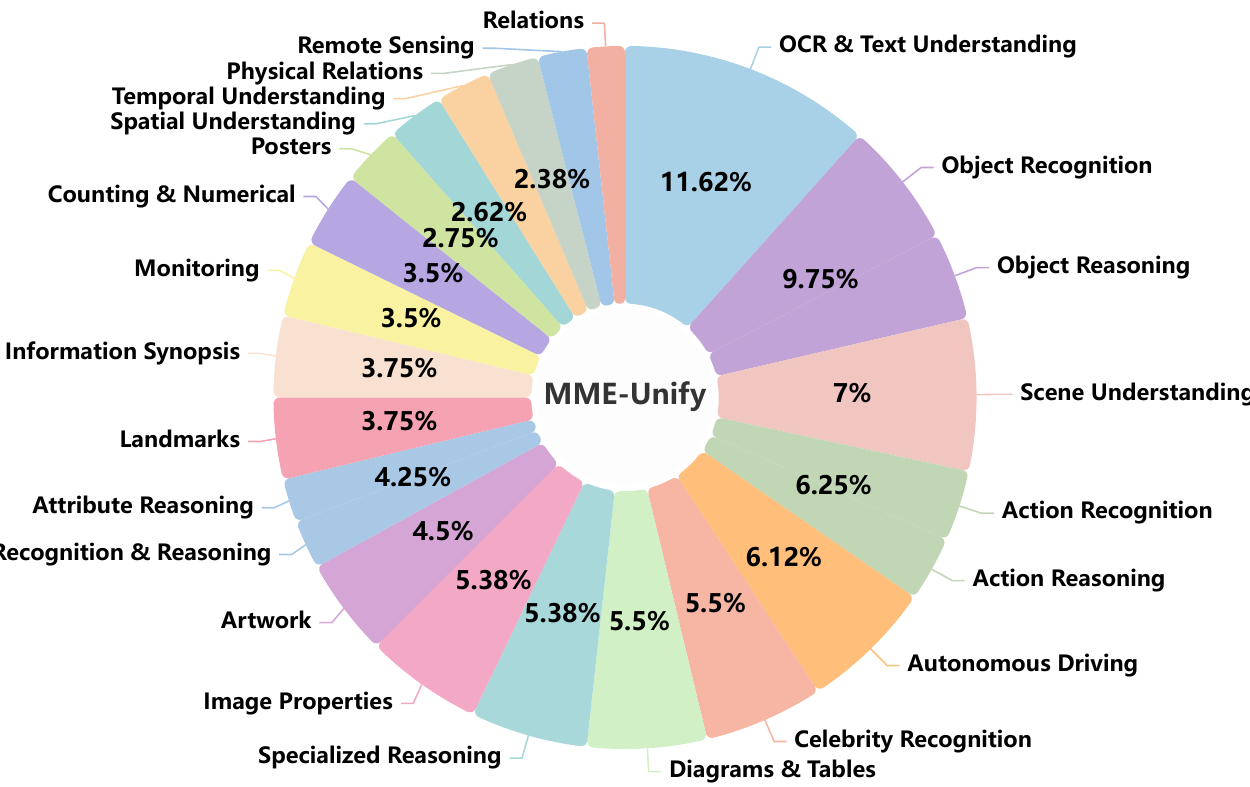}
  \caption{\textbf{An overview of real-life scenarios included in the Understanding Task.} The scores in the bars represent the proportion of the number of samples of the corresponding scenario to the total number of samples of the task.}
  \label{fig:subtask_disturbution}
\end{figure}

\begin{figure}
  \centering
  \includegraphics[width=1.0\linewidth]{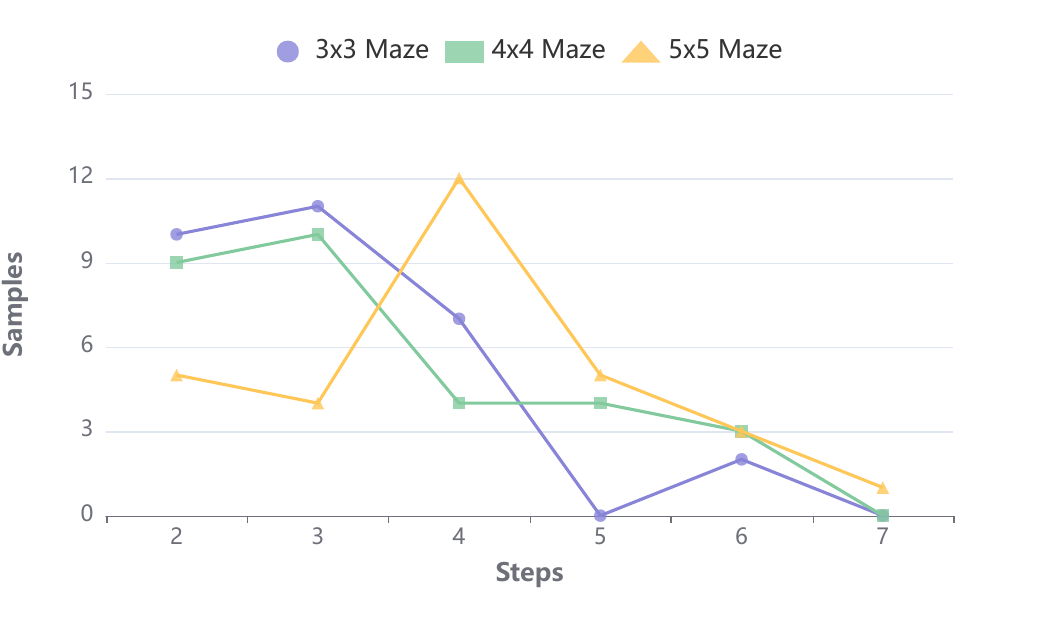}
  \caption{Distribution of steps required for samples of mazes of different sizes in the Visual CoT task.}
  \label{fig:vcot_steps}
\end{figure}

\begin{figure*}  
\centering
\includegraphics[width=1.0\linewidth]{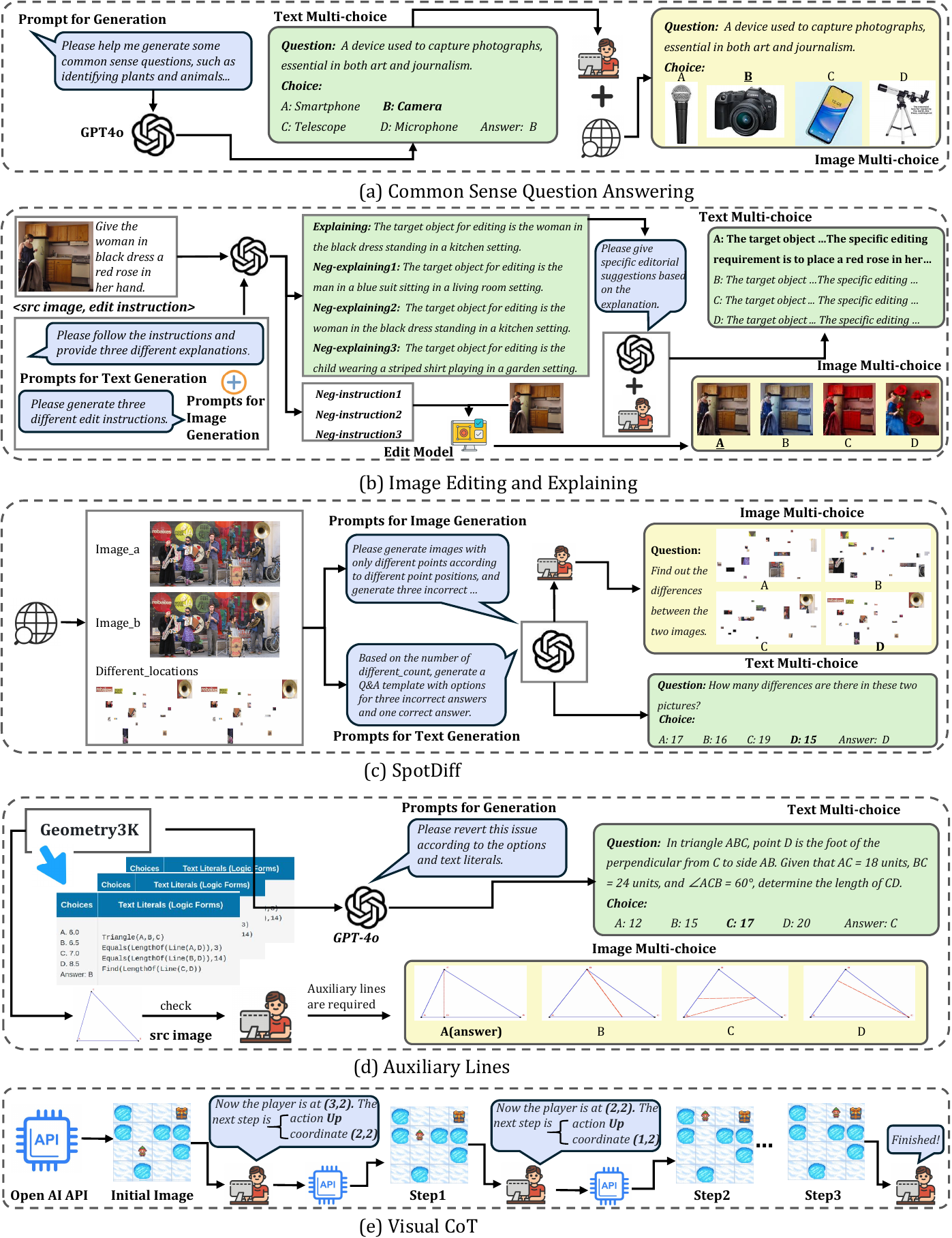}
\caption{The overall construction process for five unified tasks, which consists of (a) Common Sense Question Answering, (b) Image Editing and Explaining, (c)SpotDiff, (d) Auxiliary Lines, and (e) Visual CoT.}
\label{fig:Construction_Process}
\end{figure*}

\begin{figure*}  
\centering
\includegraphics[width=0.97\linewidth]{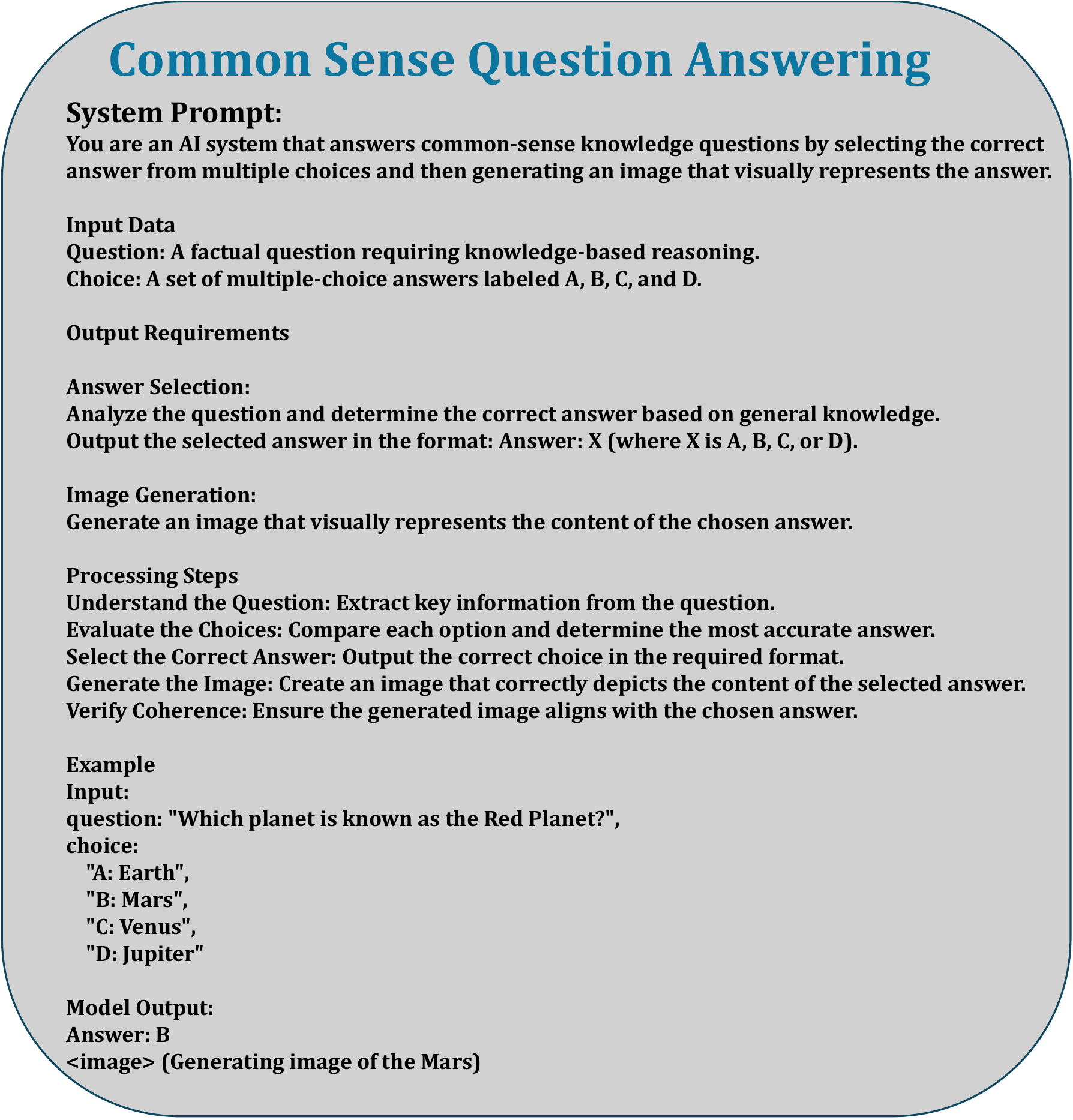}
\caption{System prompt for Common Sense Question Answering task.}
\label{figure:Prompt_for_CSQ}
\end{figure*}

\begin{figure*}  
\centering
\includegraphics[width=0.95\linewidth]{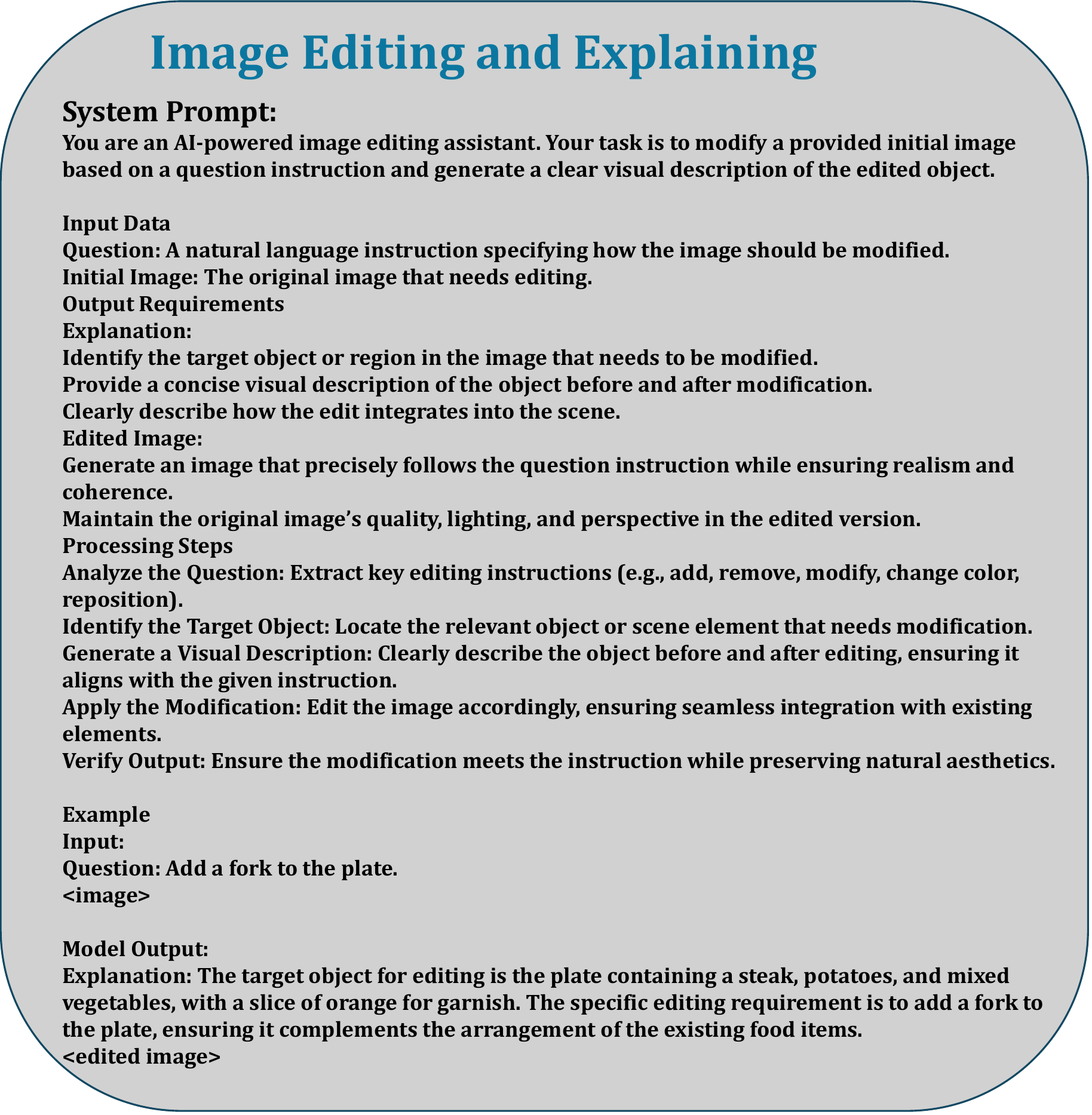}
\caption{Systemp prompt for Image Editing and Explaining task}
\label{figure:Prompt_for_IEE}
\end{figure*}

\begin{figure*}  
\centering
\includegraphics[width=0.9\linewidth]{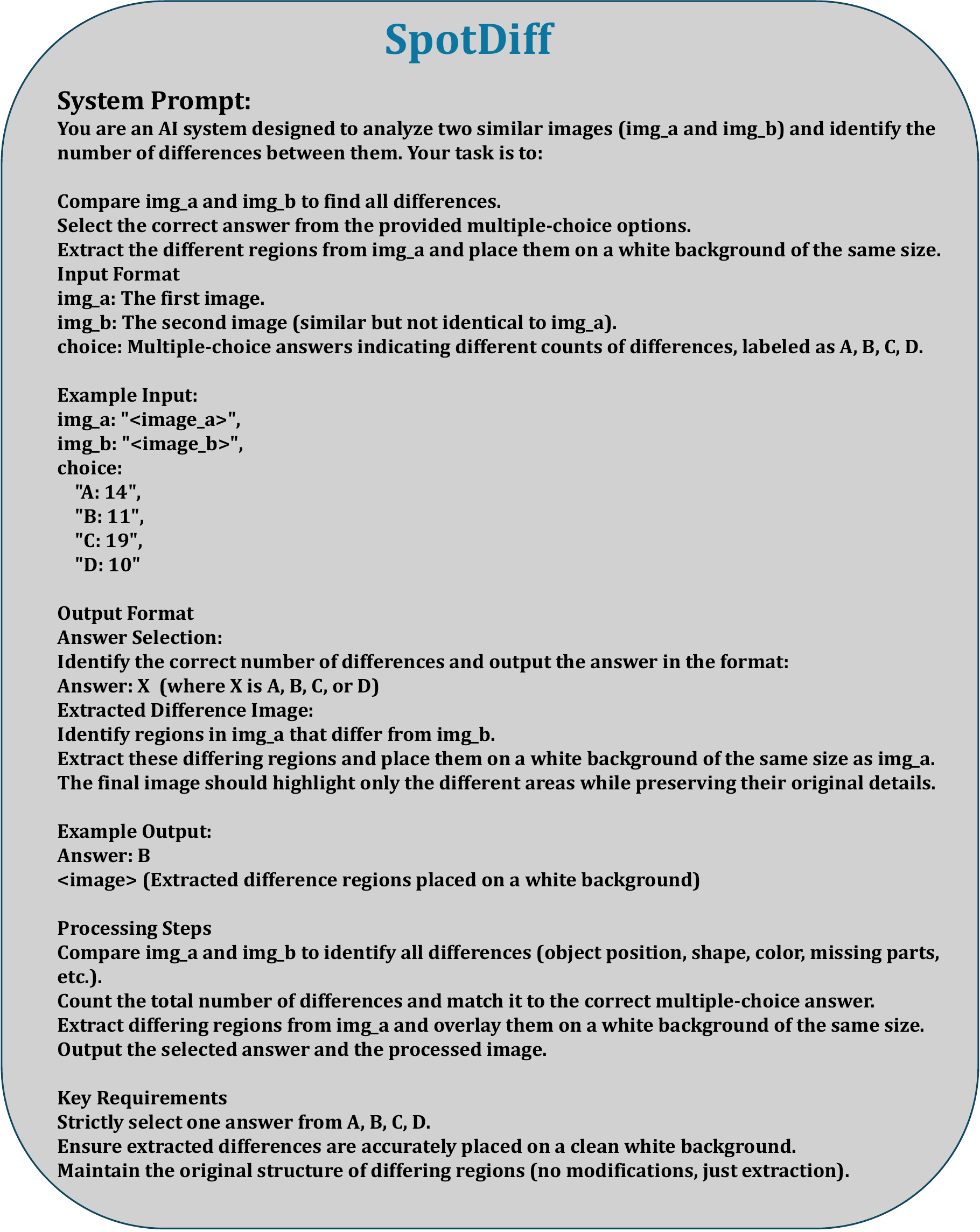}
\caption{System prompt for SpotDiff task.}
\label{figure:Prompt_for_SD}
\end{figure*}

\begin{figure*}  
\centering
\includegraphics[width=0.9\linewidth]{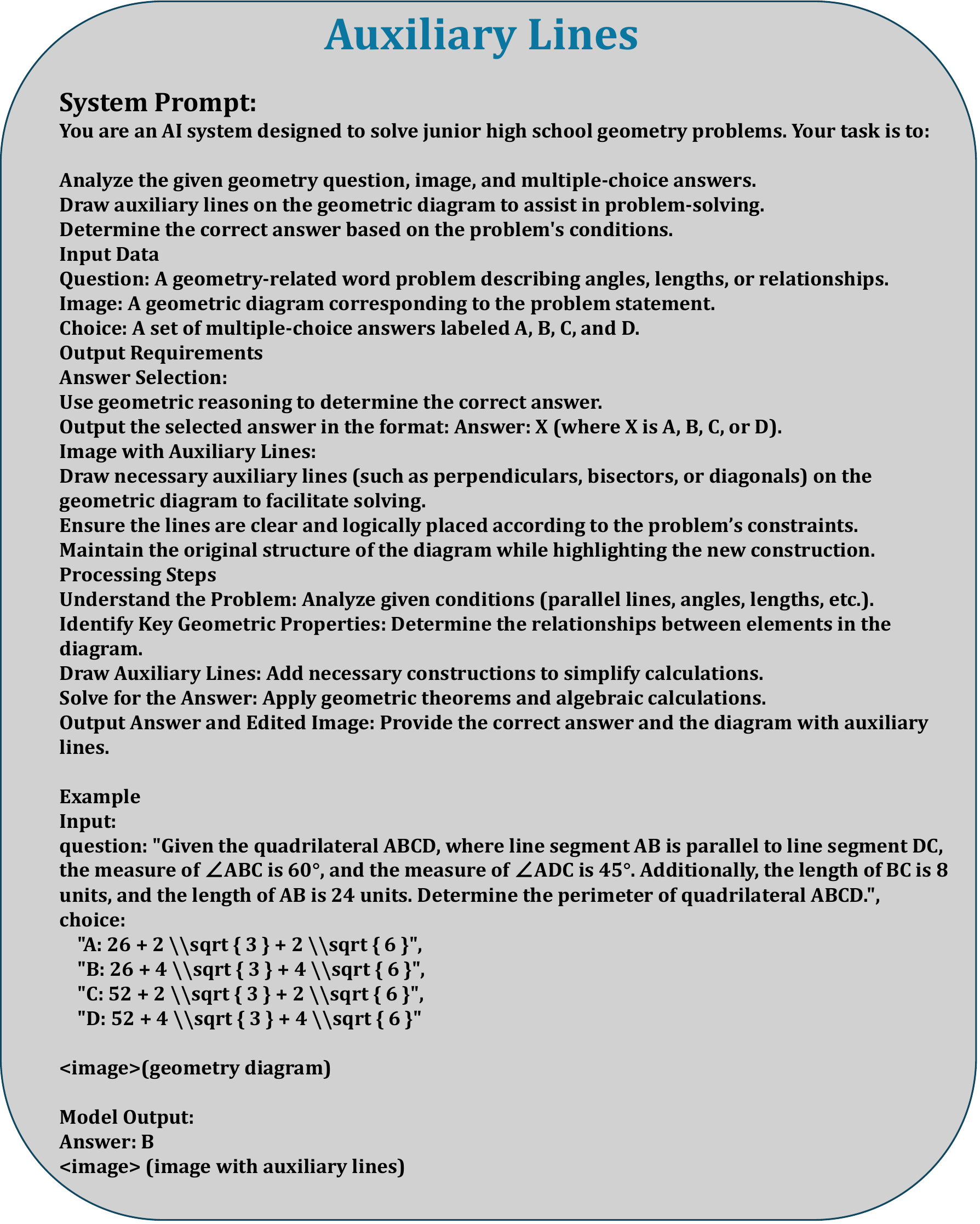}
\caption{System prompt for Auxiliary Lines task.}
\label{figure:Prompt_for_AL}
\end{figure*}

\begin{figure*}  
\centering
\includegraphics[width=0.95\linewidth]{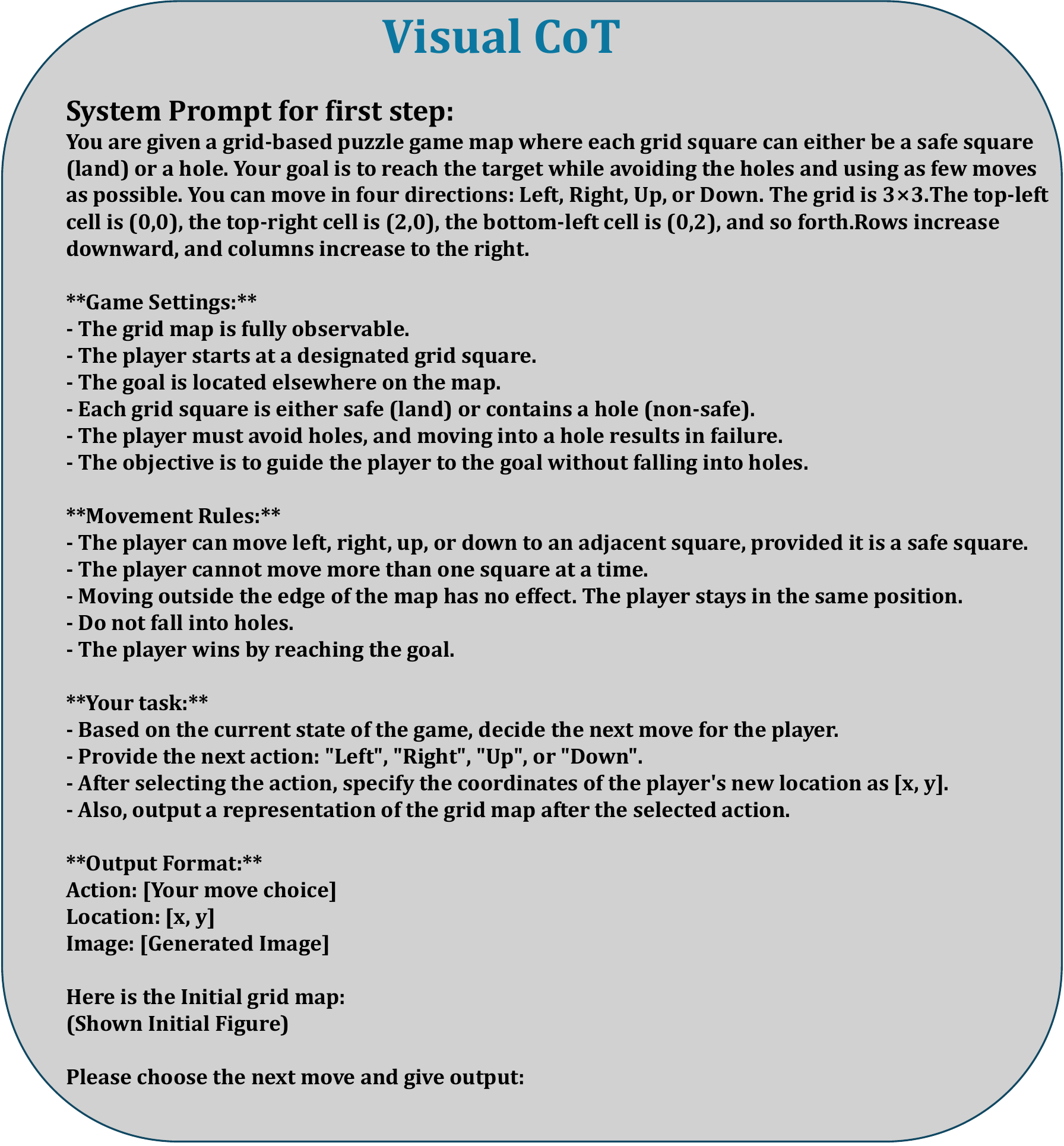}
\caption{Systemp prompt for Visual CoT task in the first step.}
\label{figure:Prompt_for_VCoT_1}
\end{figure*}

\begin{figure*}  
\centering
\includegraphics[width=0.9\linewidth]{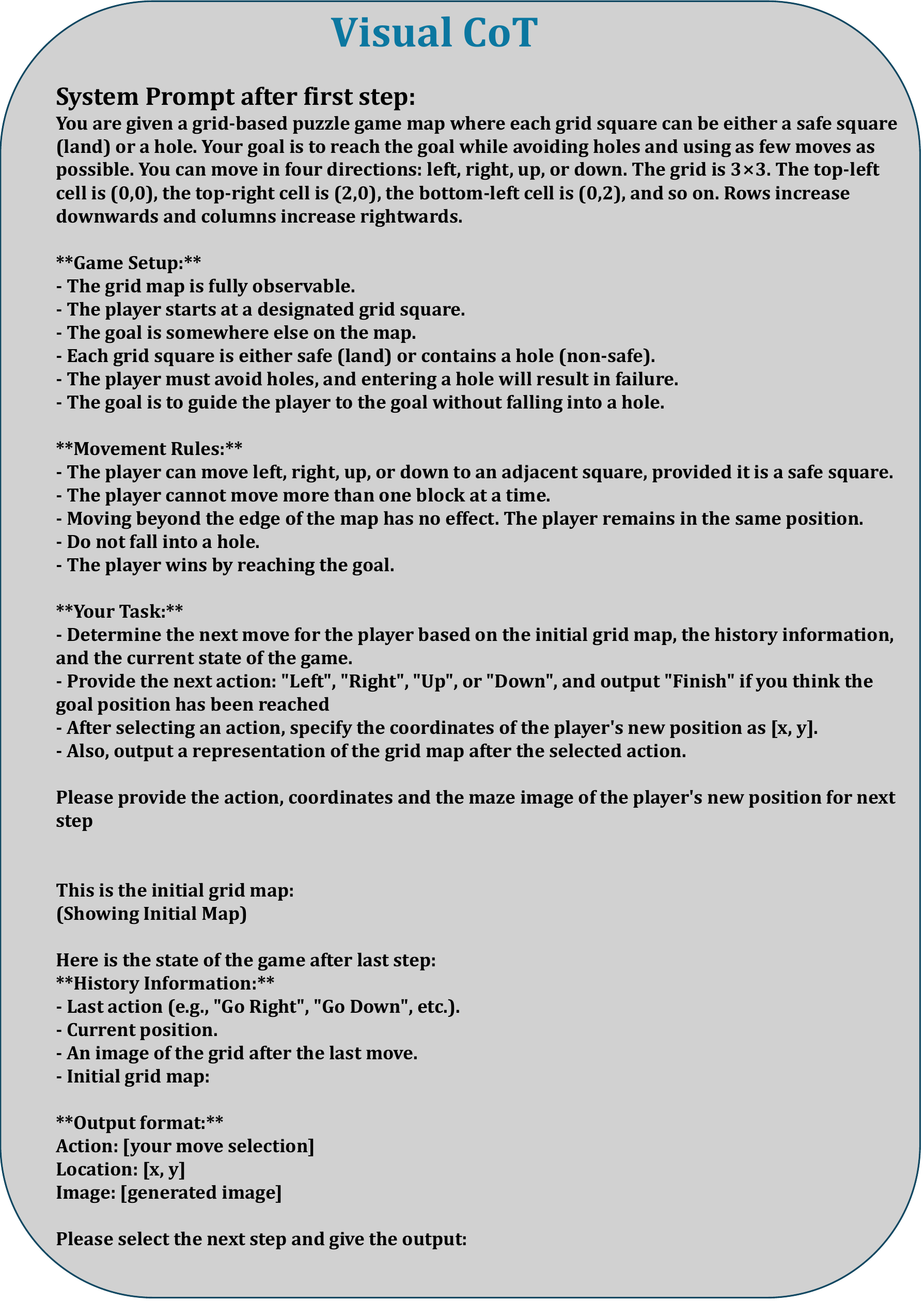}
\caption{Systemp prompt for Visual CoT task after first step.}
\label{figure:Prompt_for_VCoT_2}
\end{figure*}

\begin{figure*}[t]
  \centering
  \begin{subfigure}[b]{0.19\textwidth}
    \centering
    \includegraphics[width=\textwidth,height=3.5cm]{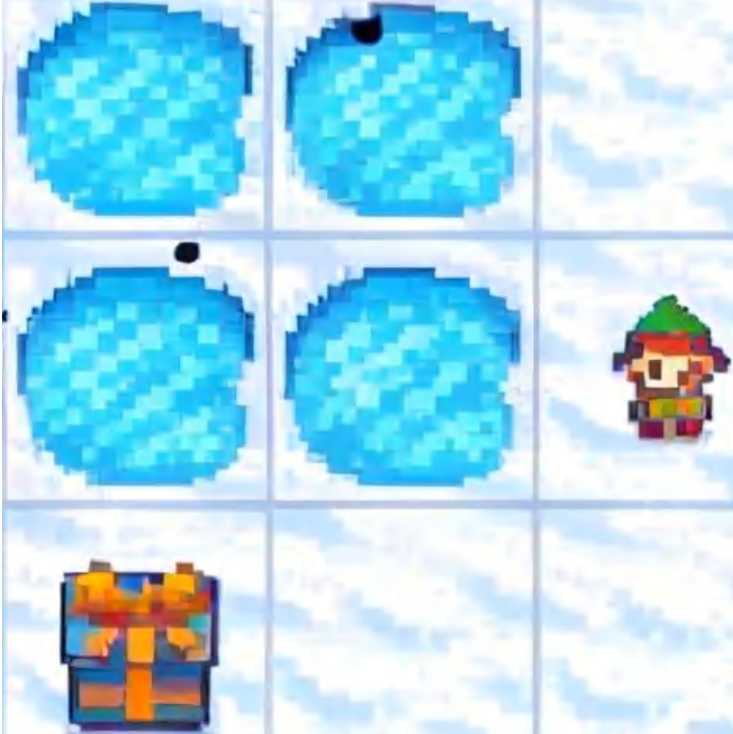}
    \caption{Anole.}
  \end{subfigure}
  \hfill
  \begin{subfigure}[b]{0.19\textwidth}
    \centering
    \includegraphics[width=\textwidth,height=3.52cm]{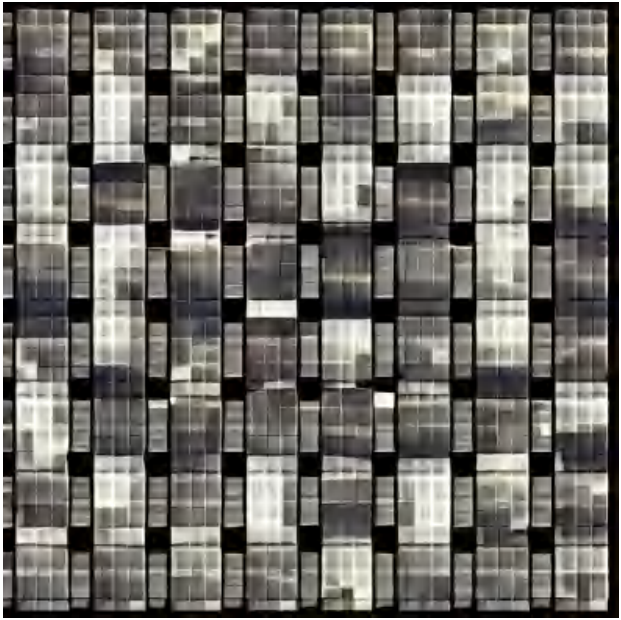}
    \caption{GILL.}
  \end{subfigure}
  \hfill
  \begin{subfigure}[b]{0.19\textwidth}
    \centering
    \includegraphics[width=\textwidth,height=3.52cm]{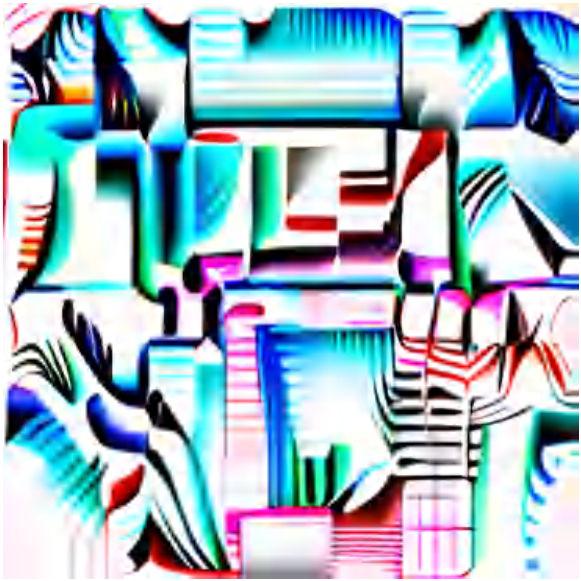}
    \caption{SEED.}
  \end{subfigure}
  \begin{subfigure}[b]{0.19\textwidth}
    \centering
    \includegraphics[width=\textwidth,height=3.52cm]{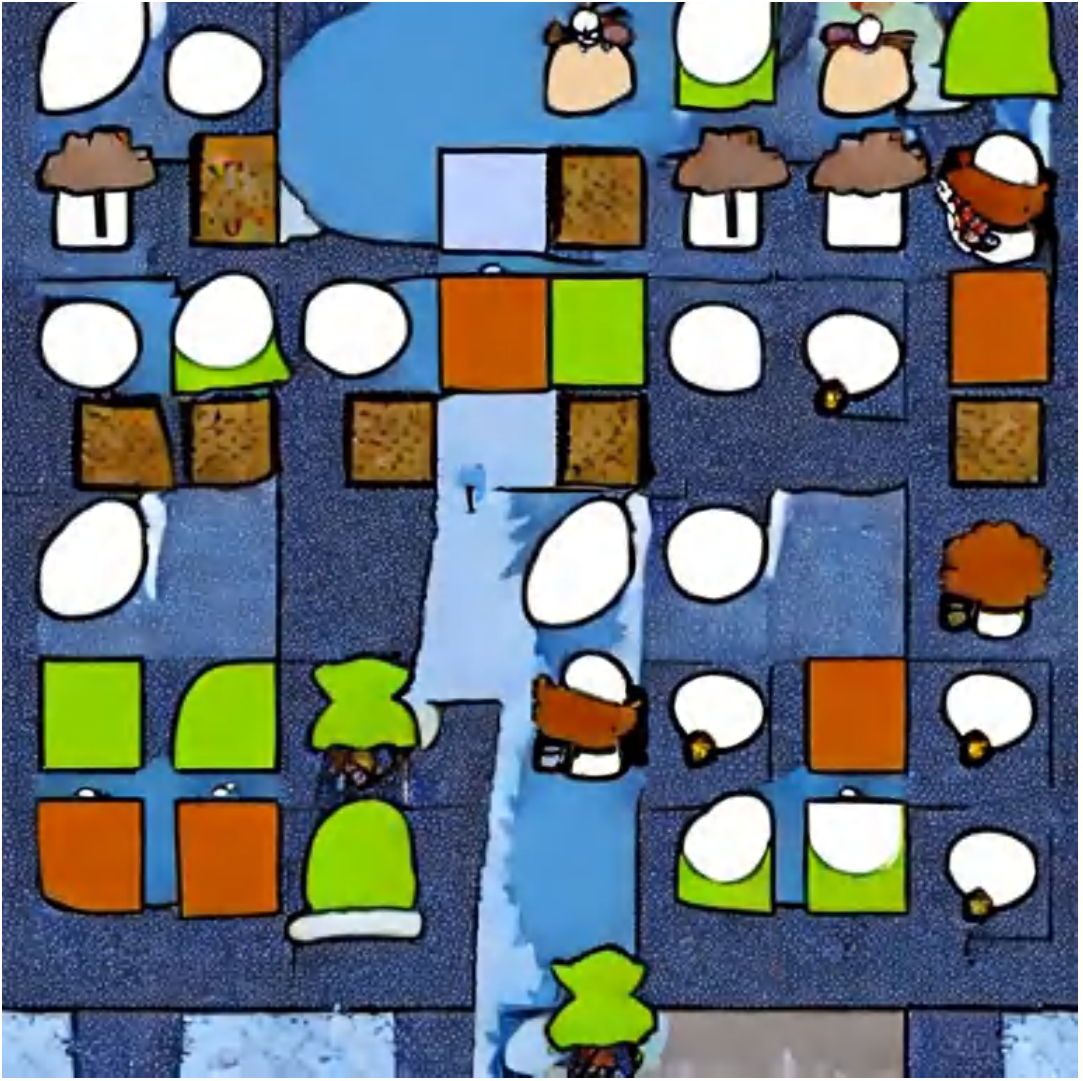}
    \caption{MiniGPT-5.}
  \end{subfigure}
  \begin{subfigure}[b]{0.19\textwidth}
    \centering
    \includegraphics[width=\textwidth,height=3.52cm]{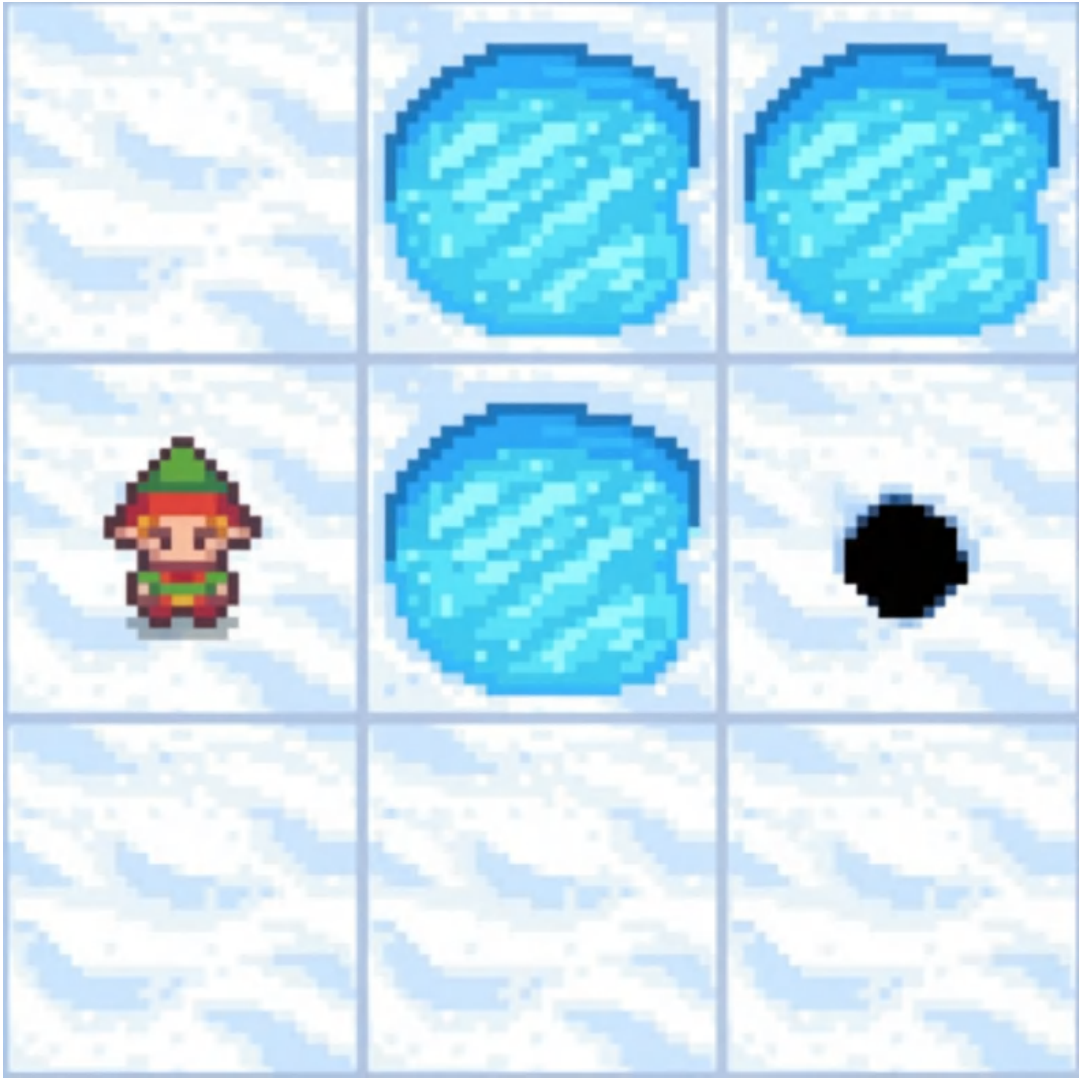}
    \caption{Gemini2.0-flash-exp.}
  \end{subfigure}
  \caption{\textbf{Intermediate process images generated by different models in VCoT.} The figure illustrates the intermediate outputs of various models in the VCoT (Visual Composition Task), showing distinct approaches in processing and generating visual content. The models shown include (a) Anole, (b) GILL, (c) SEED, (d) MiniGPT-5, and (e) Gemini-2.0-flash-exp, each producing unique visual patterns and compositions.}
  \label{fig:vcot_fig}
\end{figure*}

\begin{figure*}[t]
  \centering
  \begin{subfigure}[b]{0.24\textwidth}
    \centering
    \includegraphics[width=\textwidth,height=3.75cm]{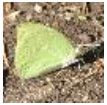}
    \caption{Source Image.}
  \end{subfigure}
  \hfill
  \begin{subfigure}[b]{0.24\textwidth}
    \centering
    \includegraphics[width=\textwidth,height=3.75cm]{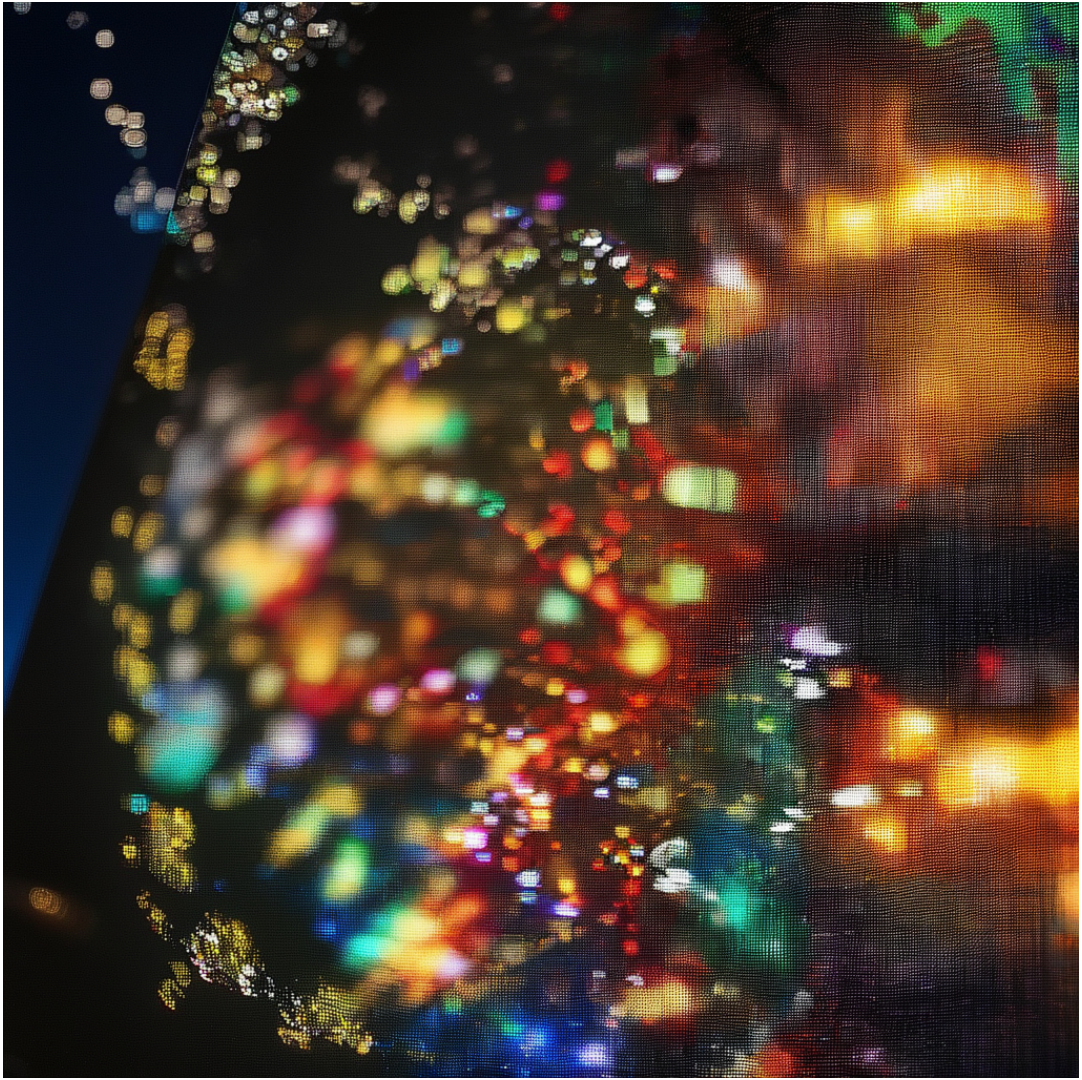}
    \caption{OmniGen.}
  \end{subfigure}
  \hfill
  \begin{subfigure}[b]{0.24\textwidth}
    \centering
    \includegraphics[width=\textwidth,height=3.75cm]{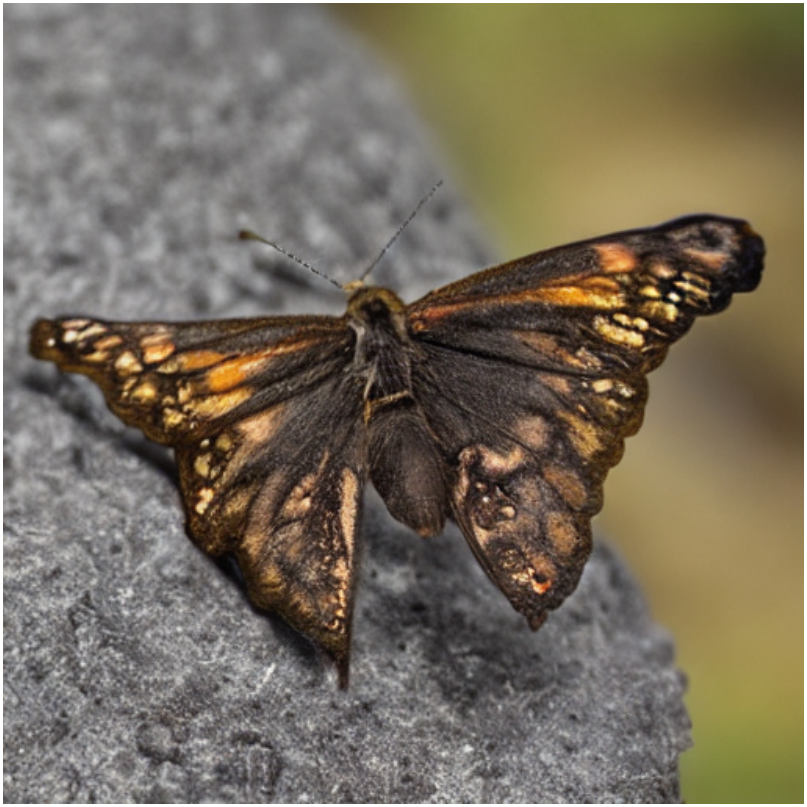}
    \caption{GILL.}
  \end{subfigure}
  \hfill
  \begin{subfigure}[b]{0.24\textwidth}
    \centering
    \includegraphics[width=\textwidth,height=3.75cm]{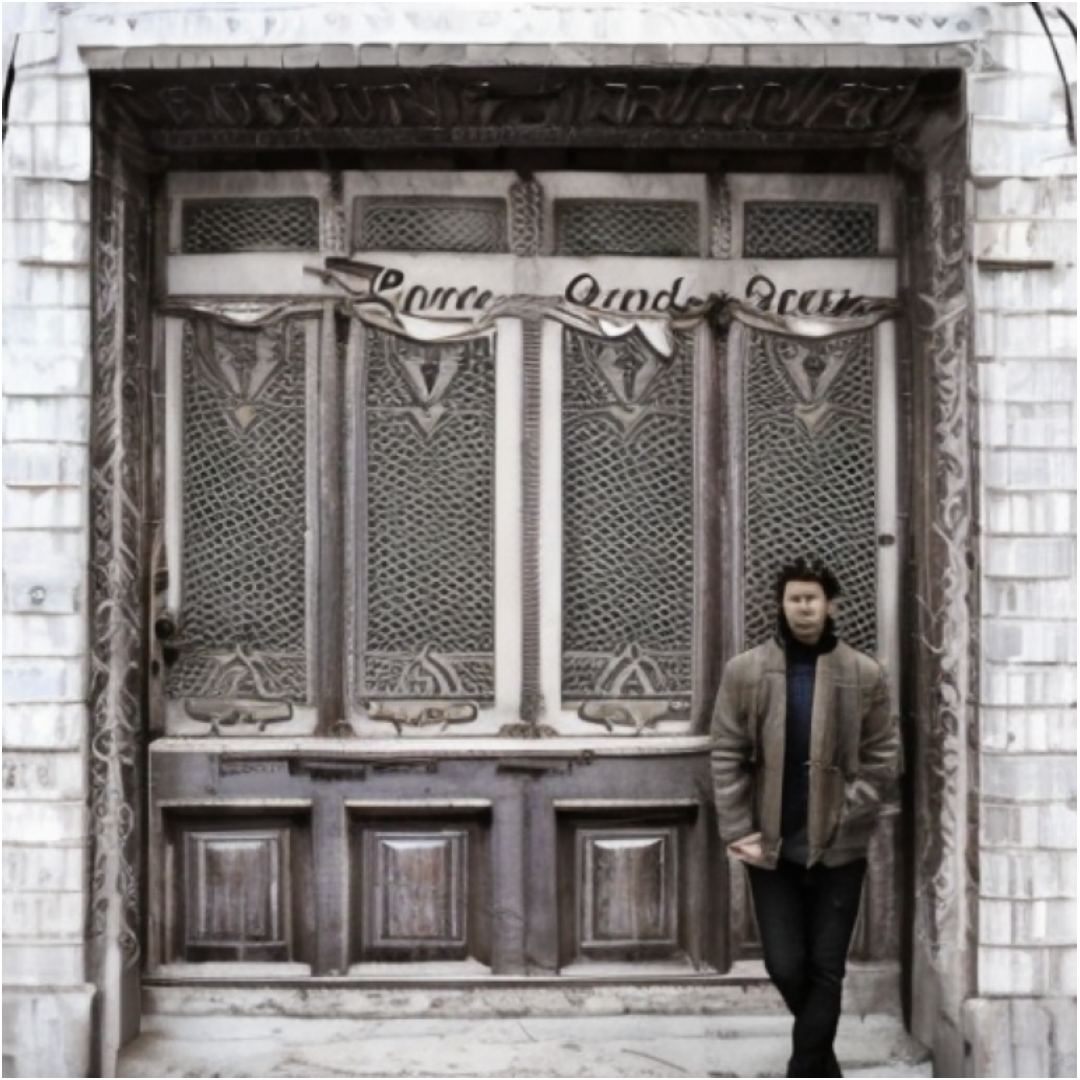}
    \caption{MiniGPT-5.}
  \end{subfigure}
  
  \vspace{0.5cm}
  \begin{subfigure}[b]{0.24\textwidth}
    \centering
    \includegraphics[width=\textwidth,height=3.75cm]{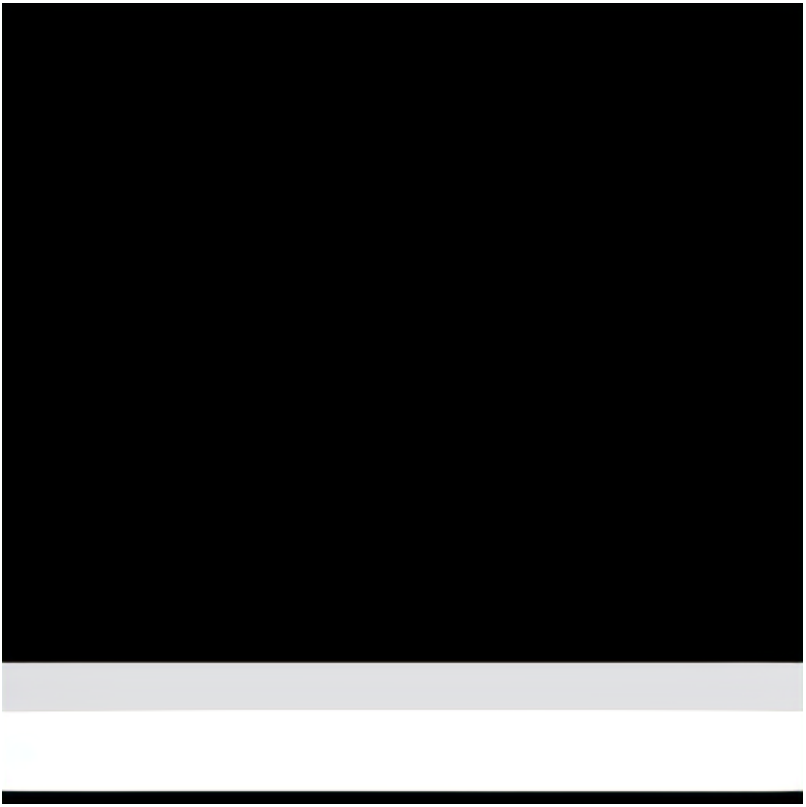}
    \caption{Anole.}
  \end{subfigure}
  \hfill
  \begin{subfigure}[b]{0.24\textwidth}
    \centering
    \includegraphics[width=\textwidth,height=3.75cm]{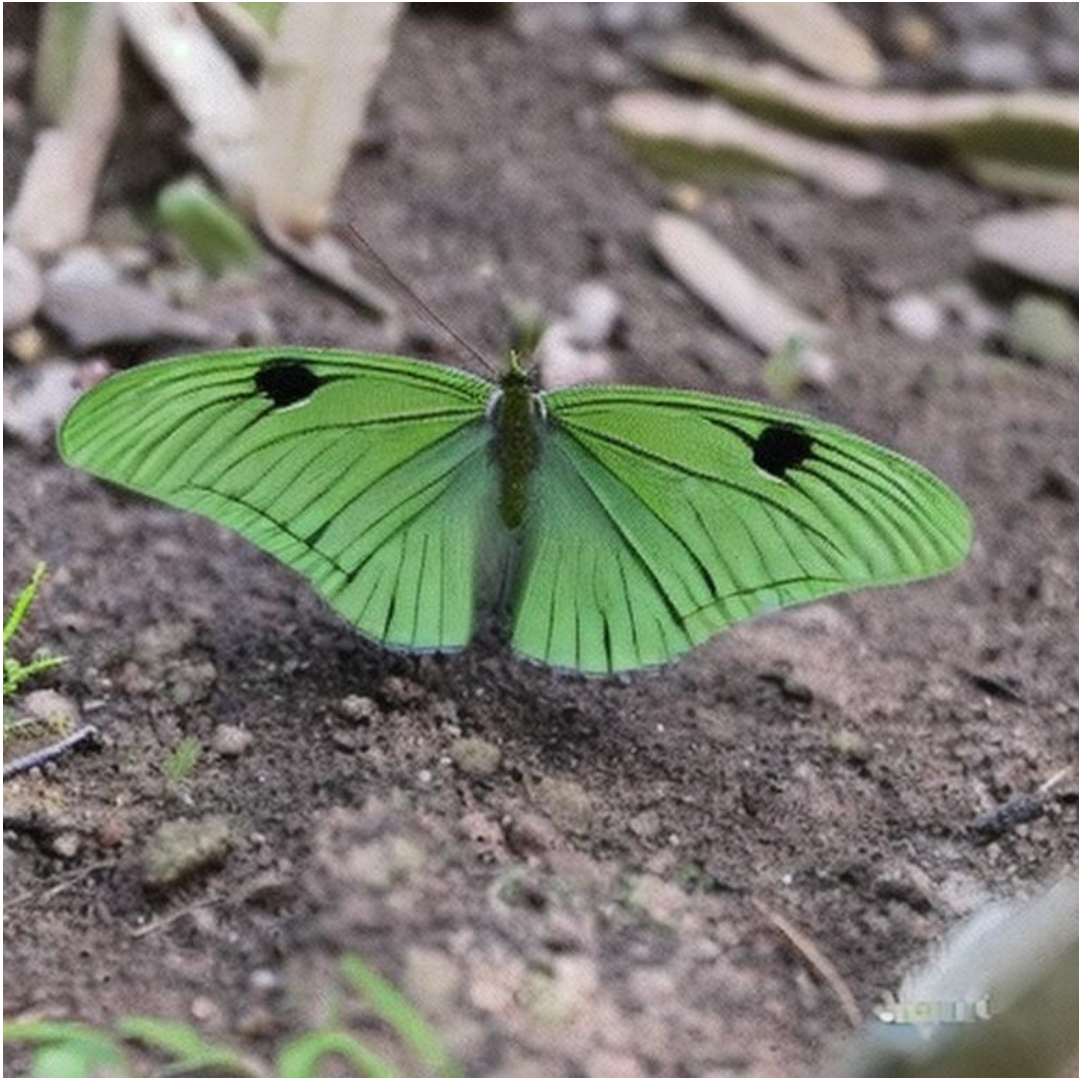}
    \caption{SEED-LLaMA.}
  \end{subfigure}
  \hfill
  \begin{subfigure}[b]{0.24\textwidth}
    \centering
    \includegraphics[width=\textwidth,height=3.75cm]{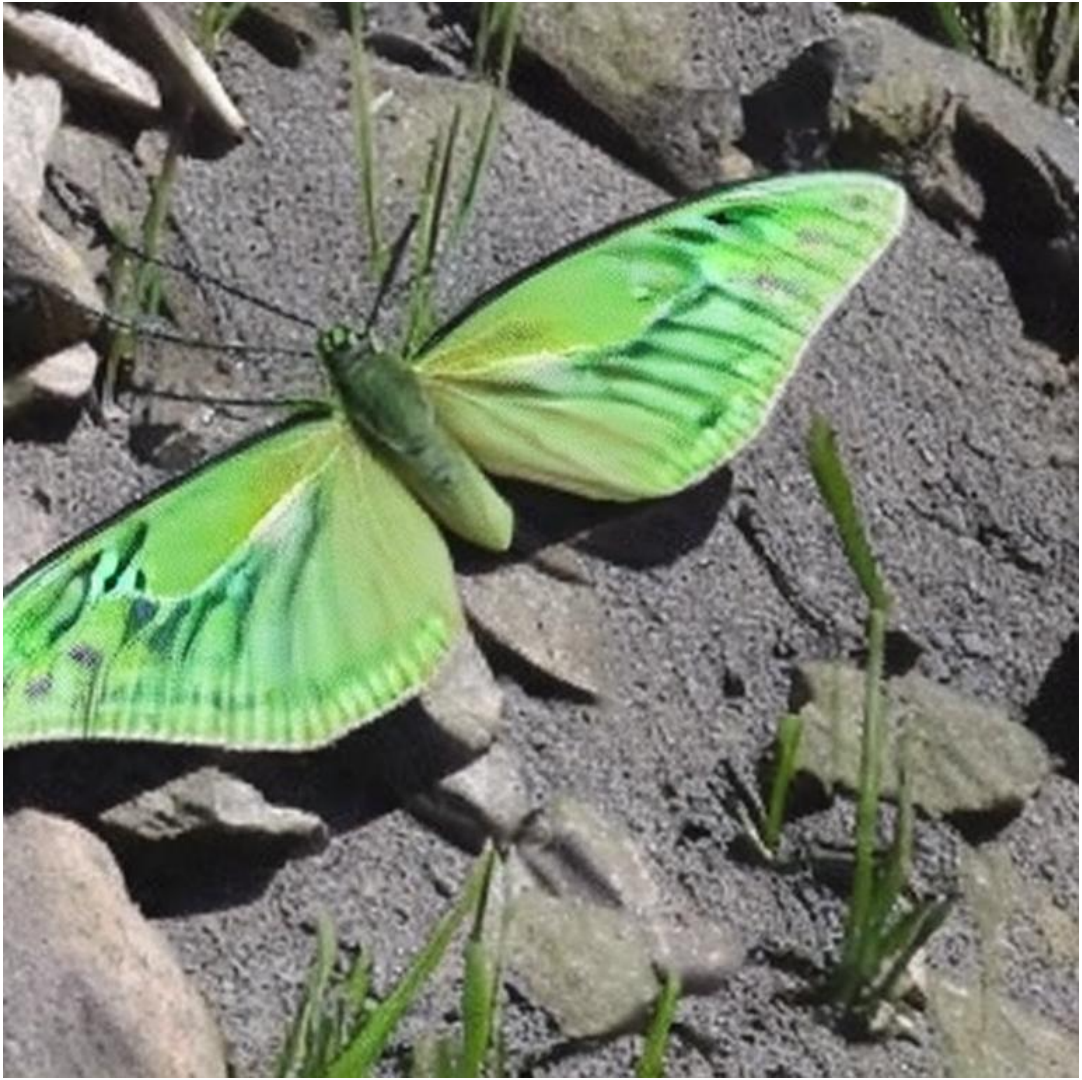}
    \caption{MIO-Instruct.}
  \end{subfigure}
  \hfill
  \begin{subfigure}[b]{0.24\textwidth}
    \centering
    \includegraphics[width=\textwidth,height=3.75cm]{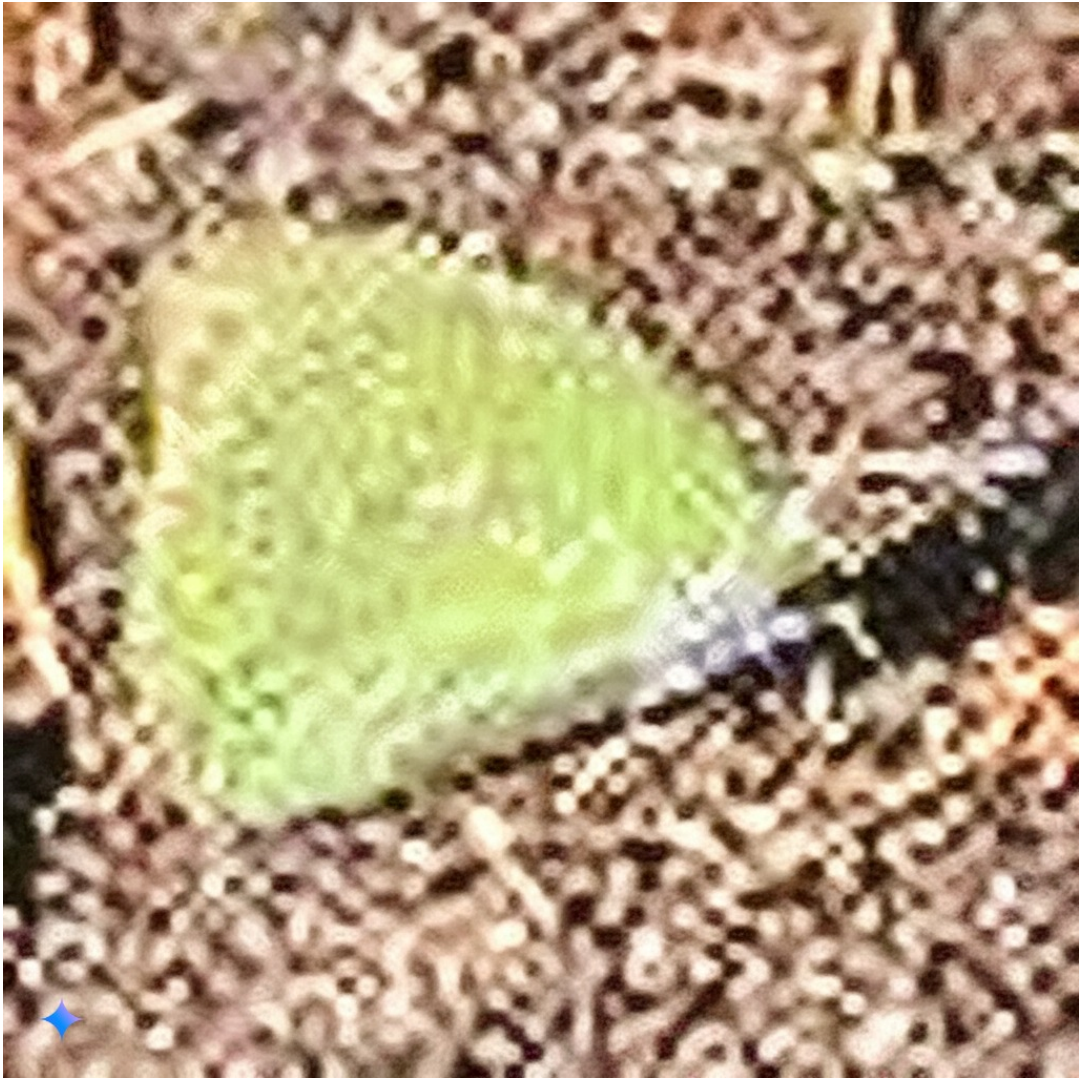}
    \caption{Gemini2.0-flash-exp.}
  \end{subfigure}
  
  \caption{\textbf{The generated results from various models in the fine-grained image reconstruction task}, based on the following text prompt: \textit{Reconstruct high-fidelity images from degraded inputs, preserving fine-grained details, textures, and structural integrity with perceptual realism}.}
  \label{fig:fir_results}
\end{figure*}

\begin{figure*}[t]
  \centering
  \begin{subfigure}[b]{0.24\textwidth}
    \centering
    \includegraphics[width=\textwidth,height=3.75cm]{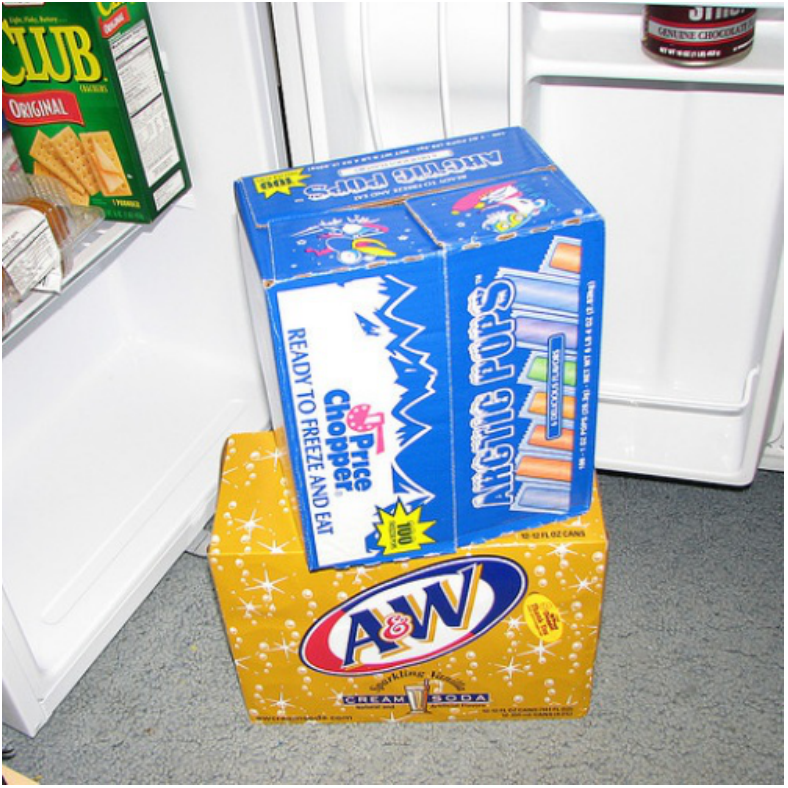}
    \caption{Source Image.}
  \end{subfigure}
  \hfill
  \begin{subfigure}[b]{0.24\textwidth}
    \centering
    \includegraphics[width=\textwidth,height=3.75cm]{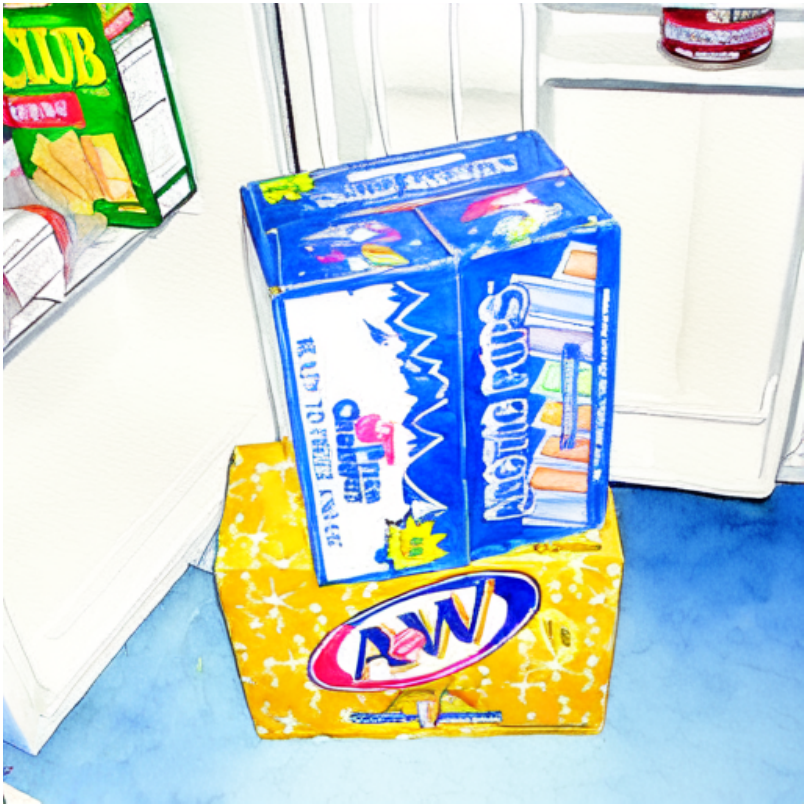}
    \caption{Ground Truth.}
  \end{subfigure}
  \hfill
  \begin{subfigure}[b]{0.24\textwidth}
    \centering
    \includegraphics[width=\textwidth,height=3.75cm]{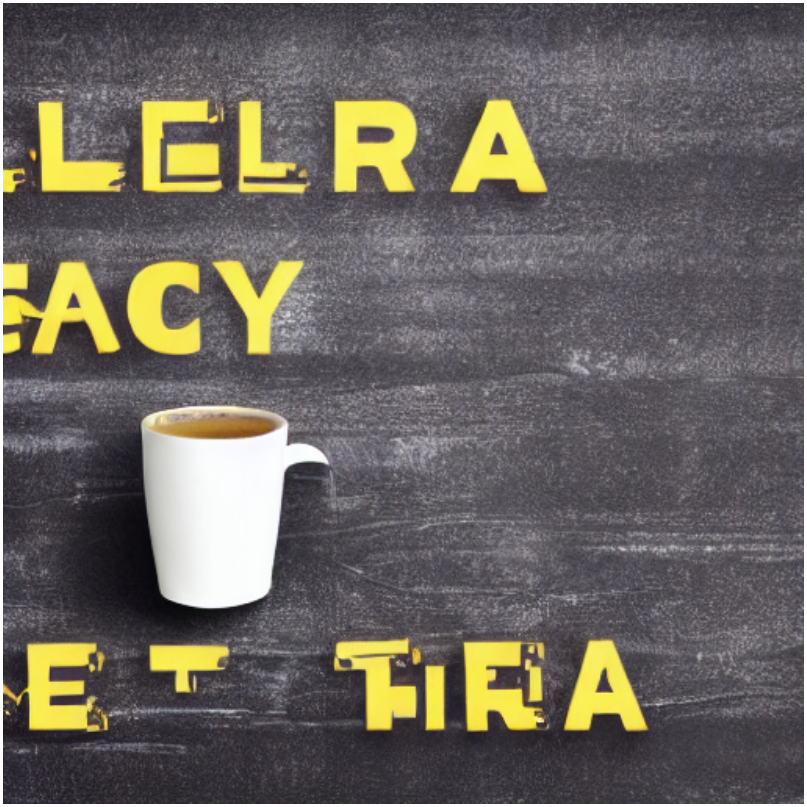}
    \caption{GILL.}
  \end{subfigure}
  \hfill
  \begin{subfigure}[b]{0.24\textwidth}
    \centering
    \includegraphics[width=\textwidth,height=3.75cm]{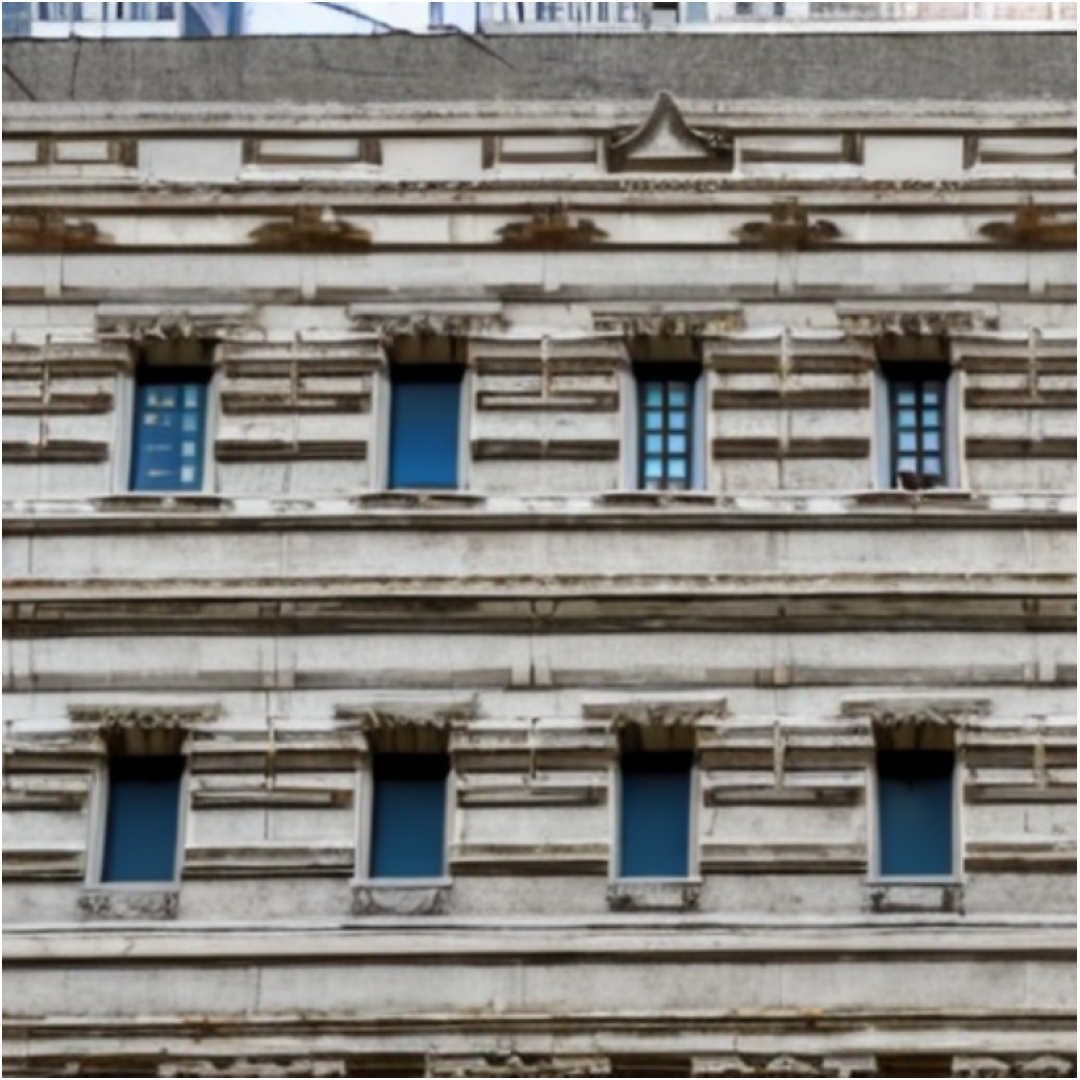}
    \caption{MiniGPT-5.}
  \end{subfigure}
  
  \vspace{0.5cm}
  \begin{subfigure}[b]{0.24\textwidth}
    \centering
    \includegraphics[width=\textwidth,height=3.75cm]{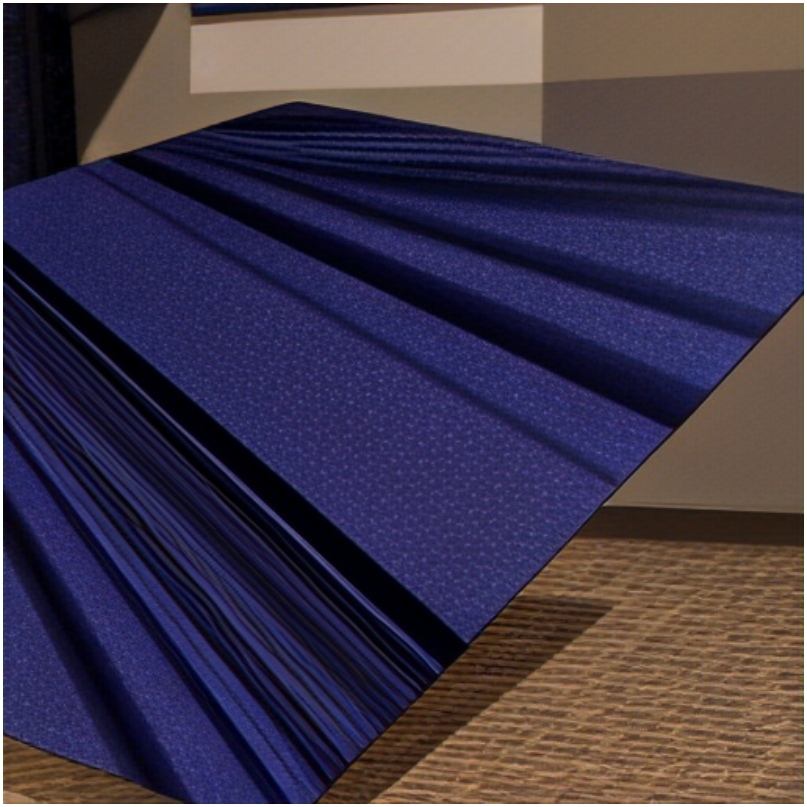}
    \caption{Anole.}
  \end{subfigure}
  \hfill
  \begin{subfigure}[b]{0.24\textwidth}
    \centering
    \includegraphics[width=\textwidth,height=3.75cm]{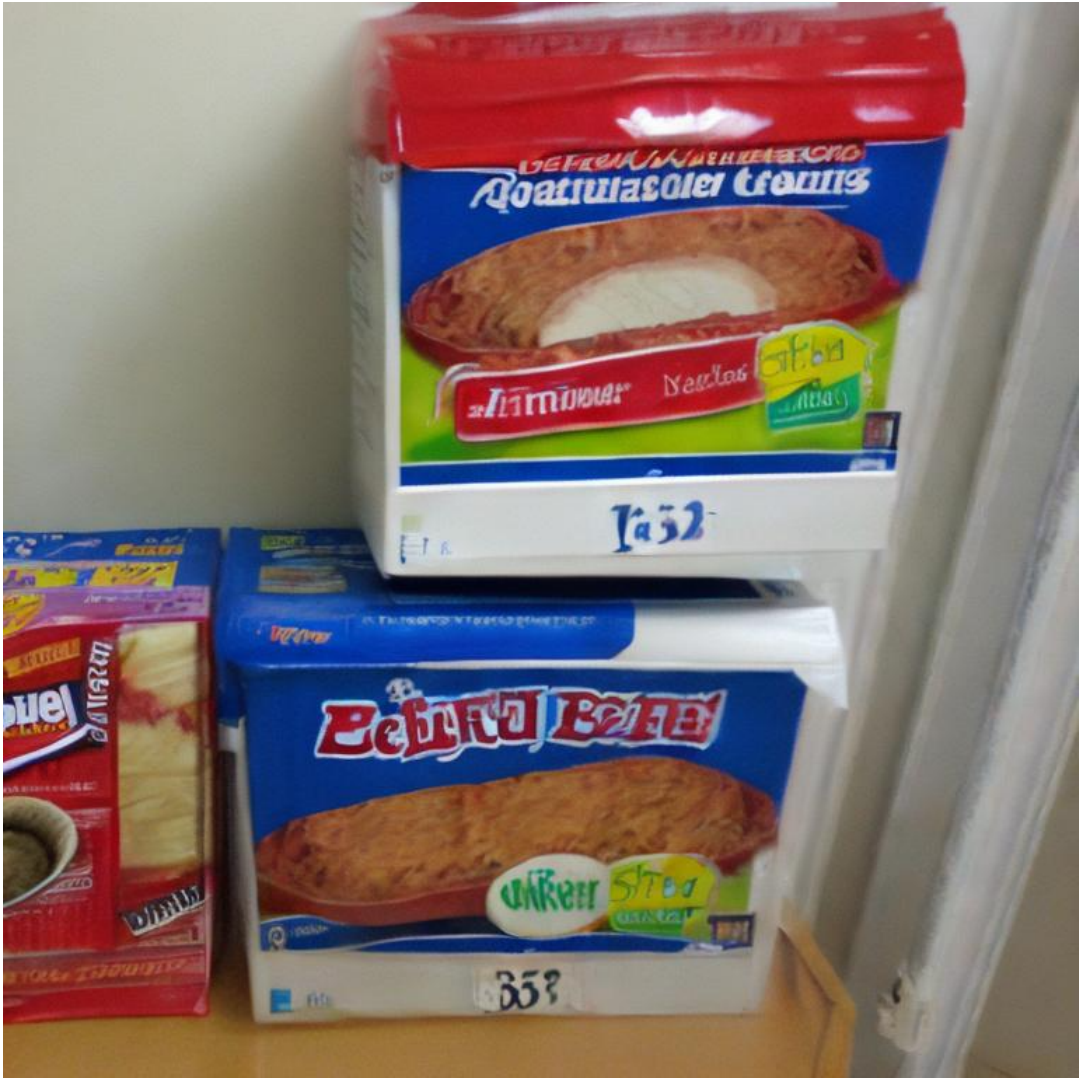}
    \caption{SEED-LLaMA.}
  \end{subfigure}
  \hfill
  \begin{subfigure}[b]{0.24\textwidth}
    \centering
    \includegraphics[width=\textwidth,height=3.75cm]{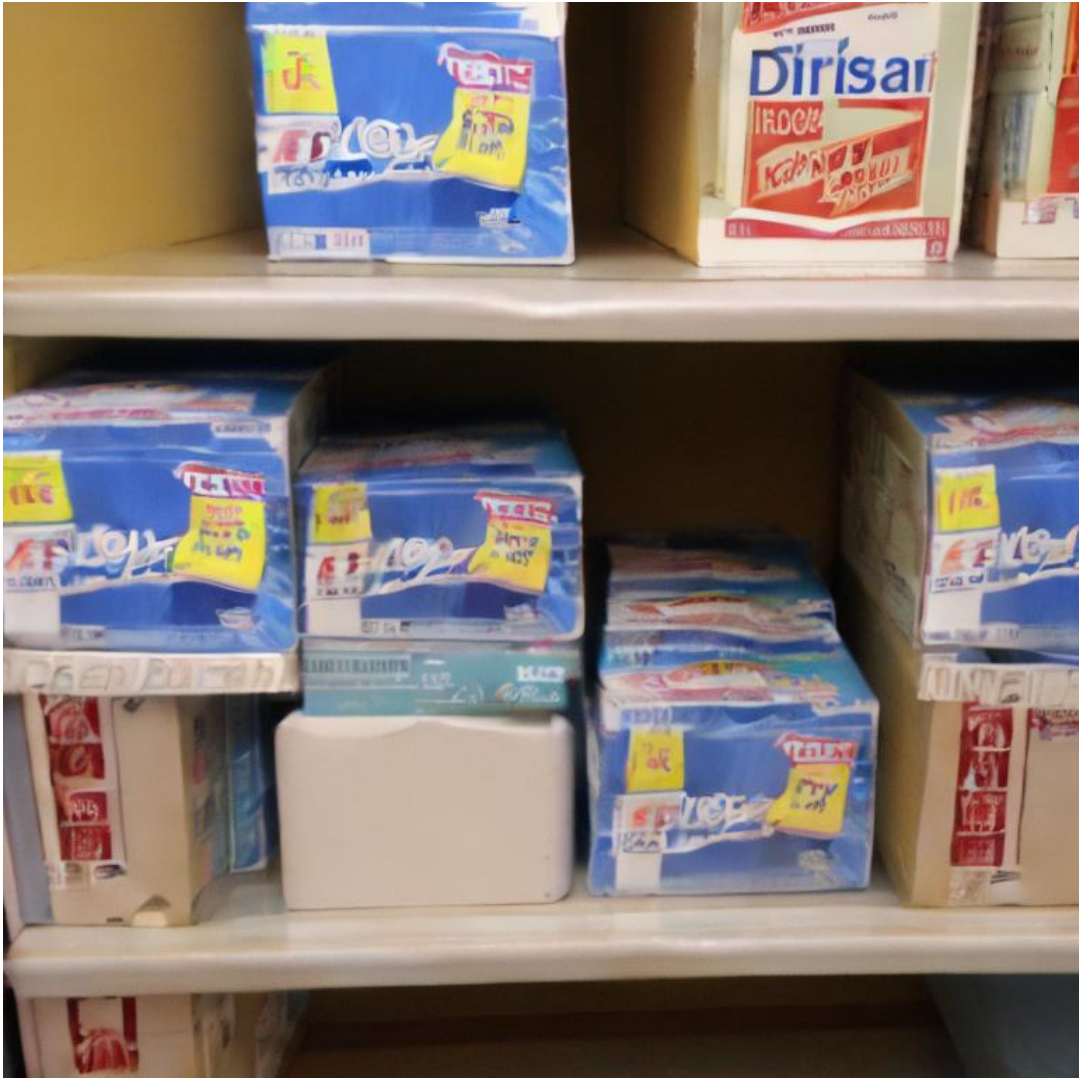}
    \caption{MIO-Instruct.}
  \end{subfigure}
  \hfill
  \begin{subfigure}[b]{0.24\textwidth}
    \centering
    \includegraphics[width=\textwidth,height=3.75cm]{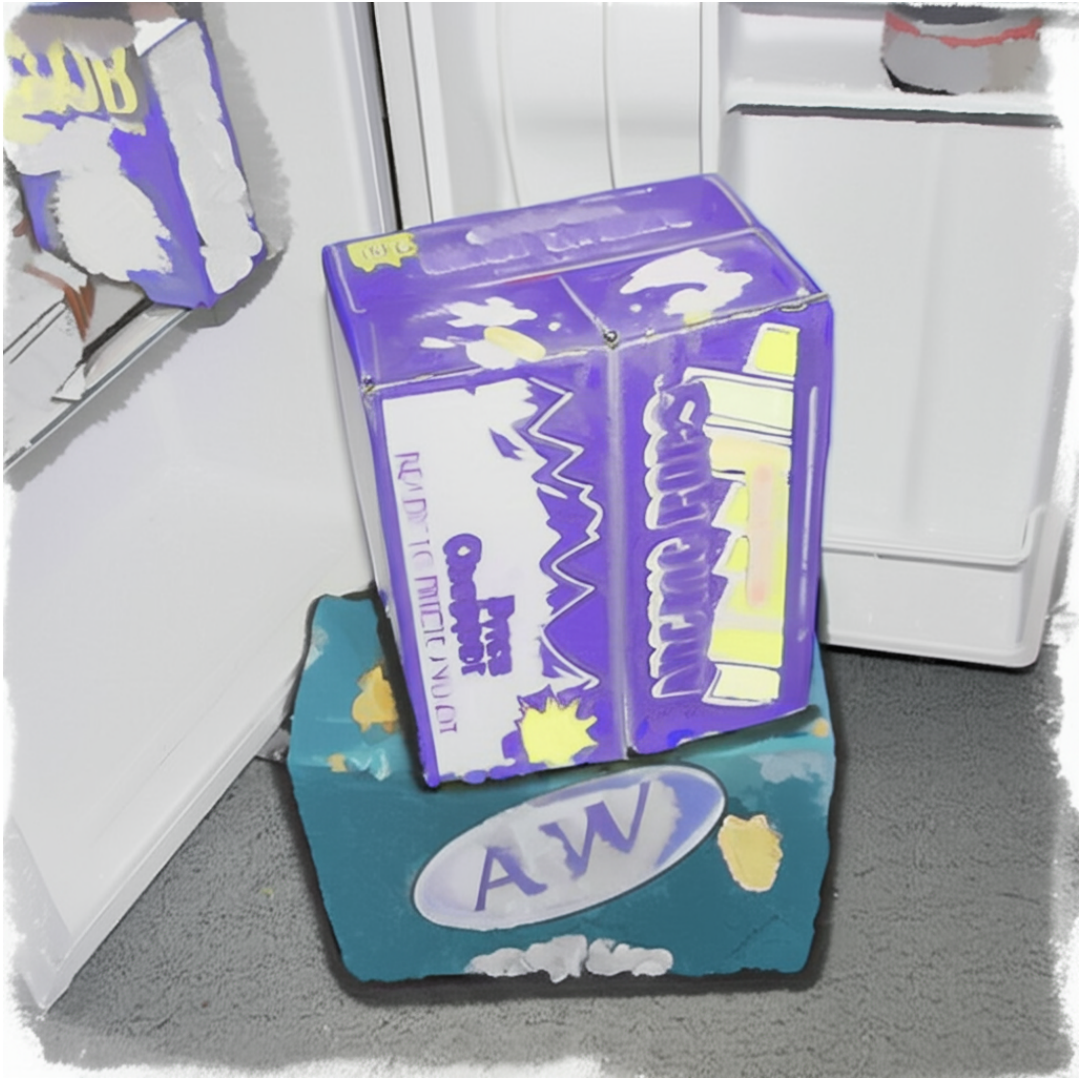}
    \caption{Gemini2.0-flash-exp.}
  \end{subfigure}
  
  \caption{\textbf{The generated results from various models in the text-guided image editing task}, based on the following text prompt: \textit{Change this image into a watercolor art}.}
  \label{fig:tie_results}
\end{figure*}

\begin{figure*}[t]

  \centering
  \begin{subfigure}[b]{0.24\textwidth}
    \centering
    \includegraphics[width=\textwidth,height=3.75cm]{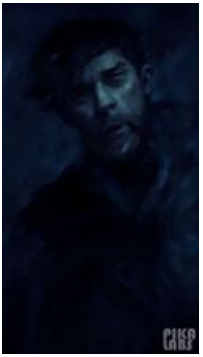}
    \caption{Source Image.}
  \end{subfigure}
   \vspace{0.5cm}

  \centering
  \begin{subfigure}[b]{0.96\textwidth}
    \centering
    \includegraphics[width=\textwidth,height=3.75cm]{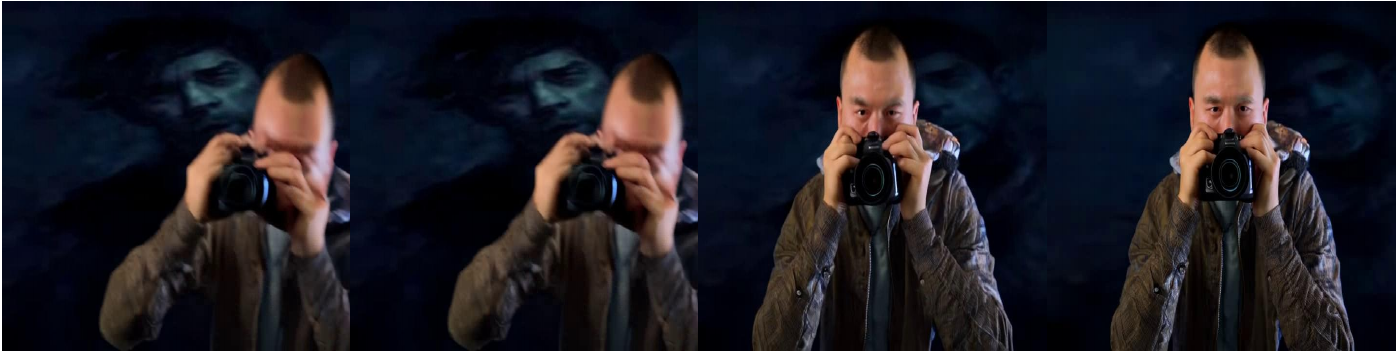}
    \caption{CogVideoX.}
  \end{subfigure}
  
  \vspace{0.5cm}
  \begin{subfigure}[b]{0.96\textwidth}
    \centering
    \includegraphics[width=\textwidth,height=3.75cm]{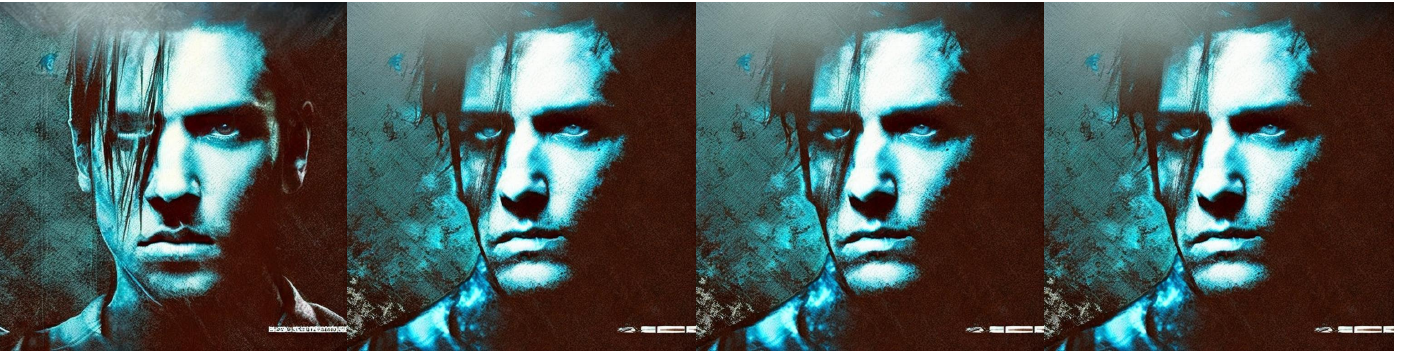}
    \caption{MIO-Instruct.}
  \end{subfigure}  
  \caption{\textbf{The generated results from various models in the conditional image-to-video generation task}, based on the following text prompt: \textit{The man is so tired. -camera zoom in}.}
  \label{fig:cvig_results}
\end{figure*}

\begin{figure*}[t]
  \centering
  \begin{subfigure}[b]{0.96\textwidth}
    \centering
    \includegraphics[width=\textwidth,height=3.75cm]{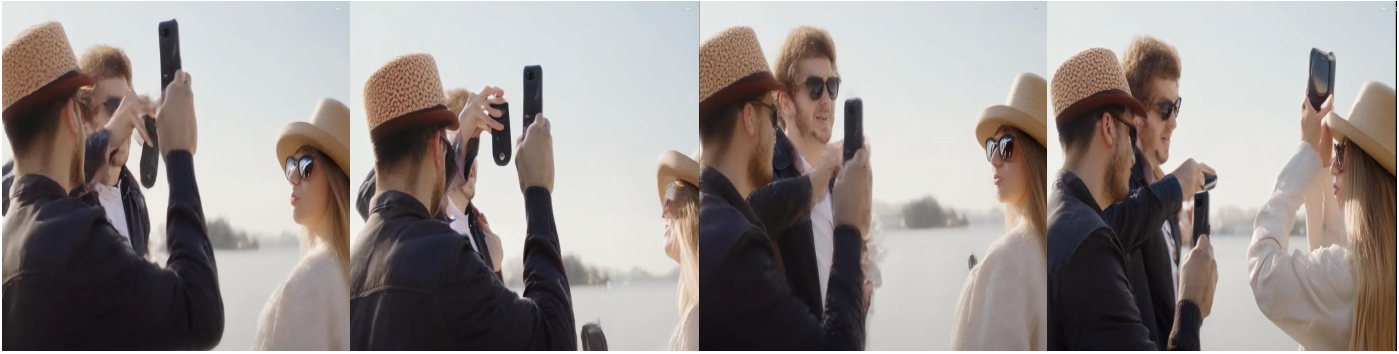}
    \caption{CogVideoX.}
  \end{subfigure}
  
  \vspace{0.5cm}
  \begin{subfigure}[b]{0.96\textwidth}
    \centering
    \includegraphics[width=\textwidth,height=3.75cm]{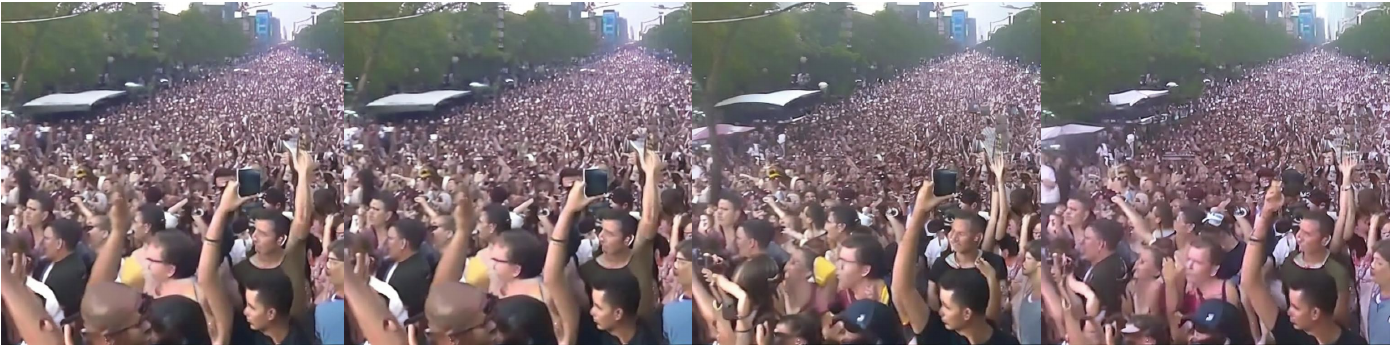}
    \caption{MIO-Instruct.}
  \end{subfigure}  
  \caption{\textbf{The generated results from various models in the text-to-video generation task}, based on the following text prompt: \textit{Men wearing sunglasses and women with hats take photographs of themselves}.}
  \label{fig:tvg_results}
\end{figure*}

\begin{figure*}  
\centering
\includegraphics[width=0.8\linewidth]{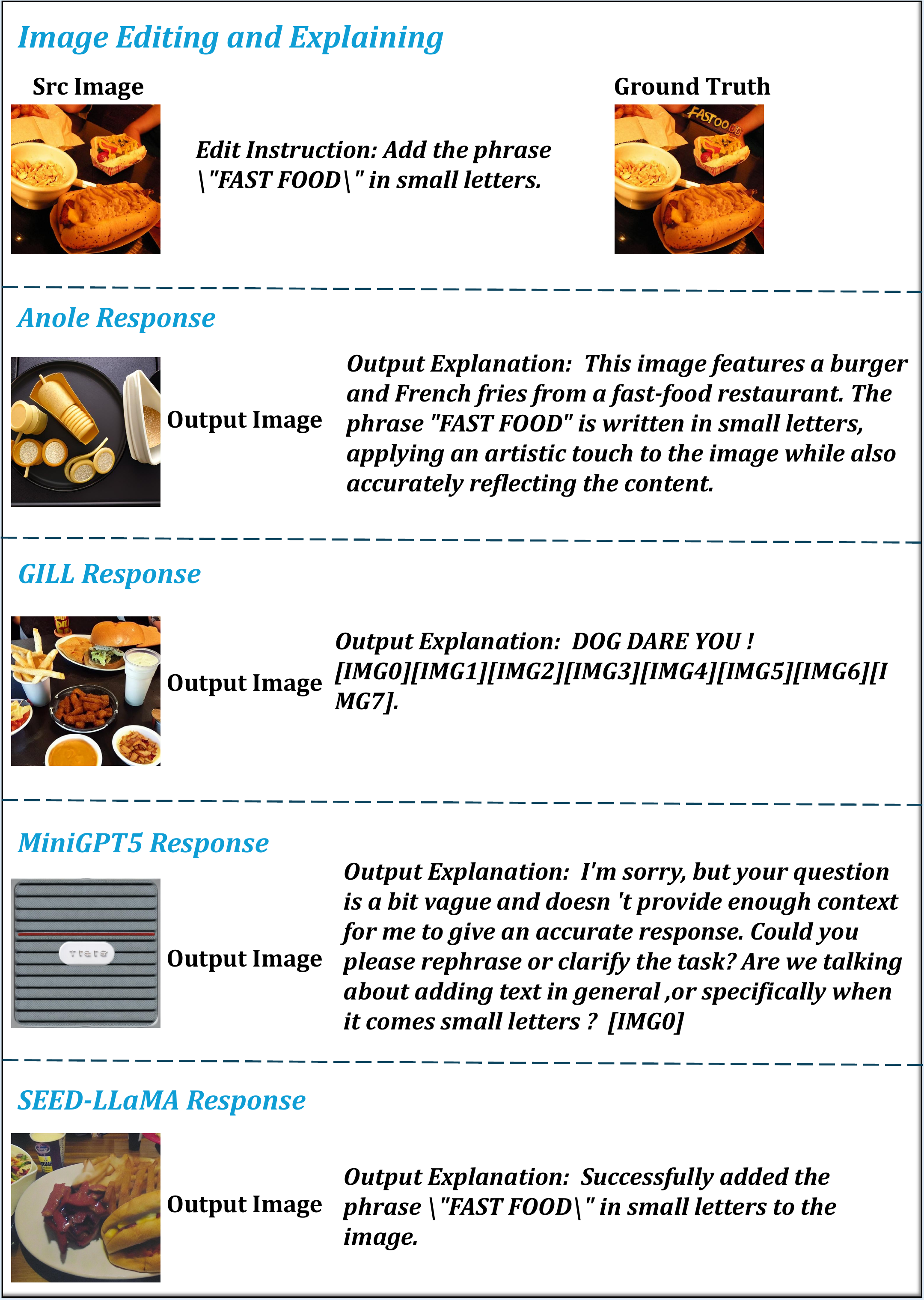}
\caption{The generated results from various models in the image editing and explaining task.}
\label{figure:iee_results}
\end{figure*}

\begin{figure*}  
\centering
\includegraphics[width=0.6\linewidth]{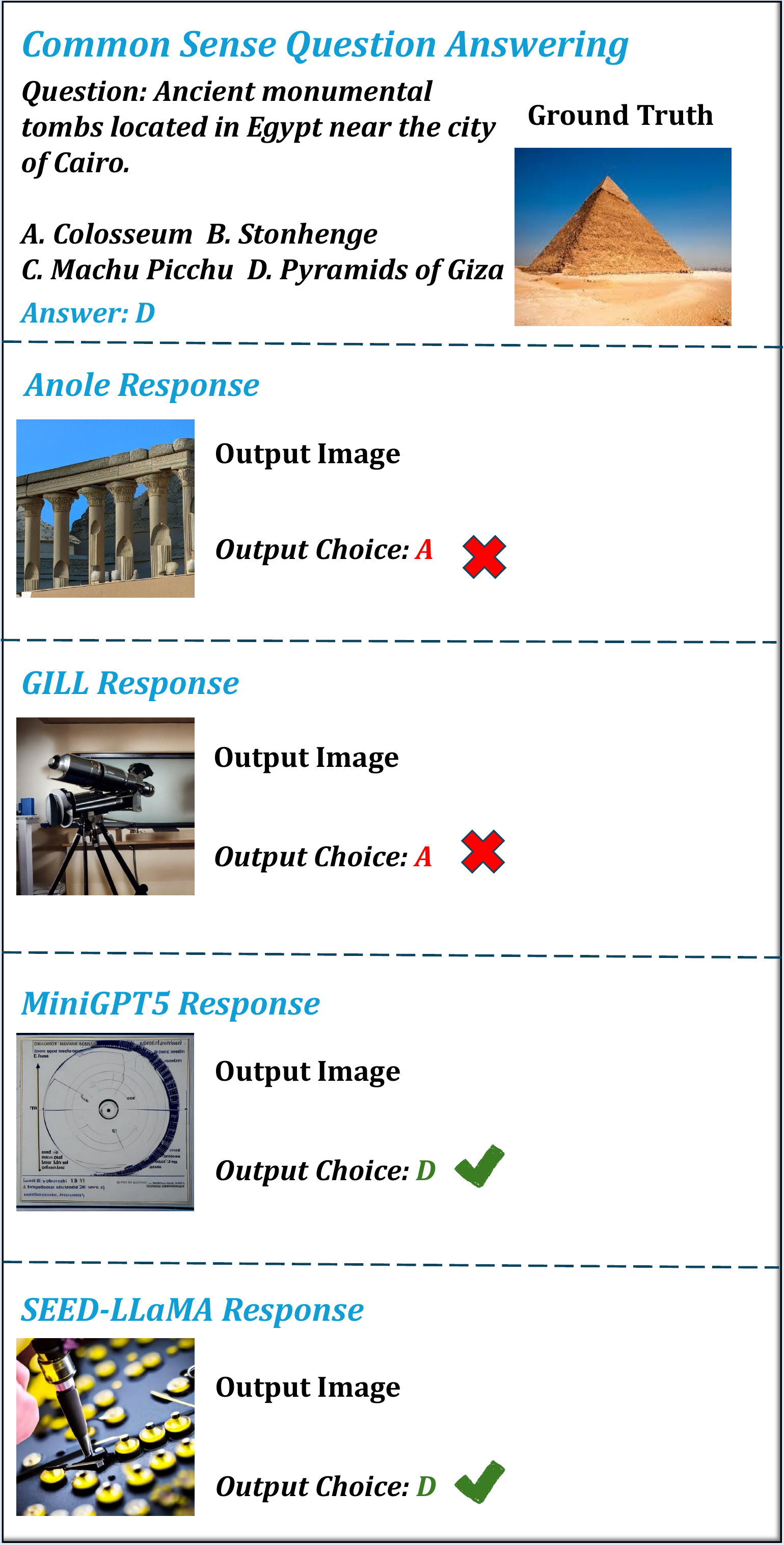}
\caption{The generated results from various models in the common sense question answering task.}
\label{figure:csq_results}
\end{figure*}

\begin{figure*}  
\centering
\includegraphics[width=0.7\linewidth]{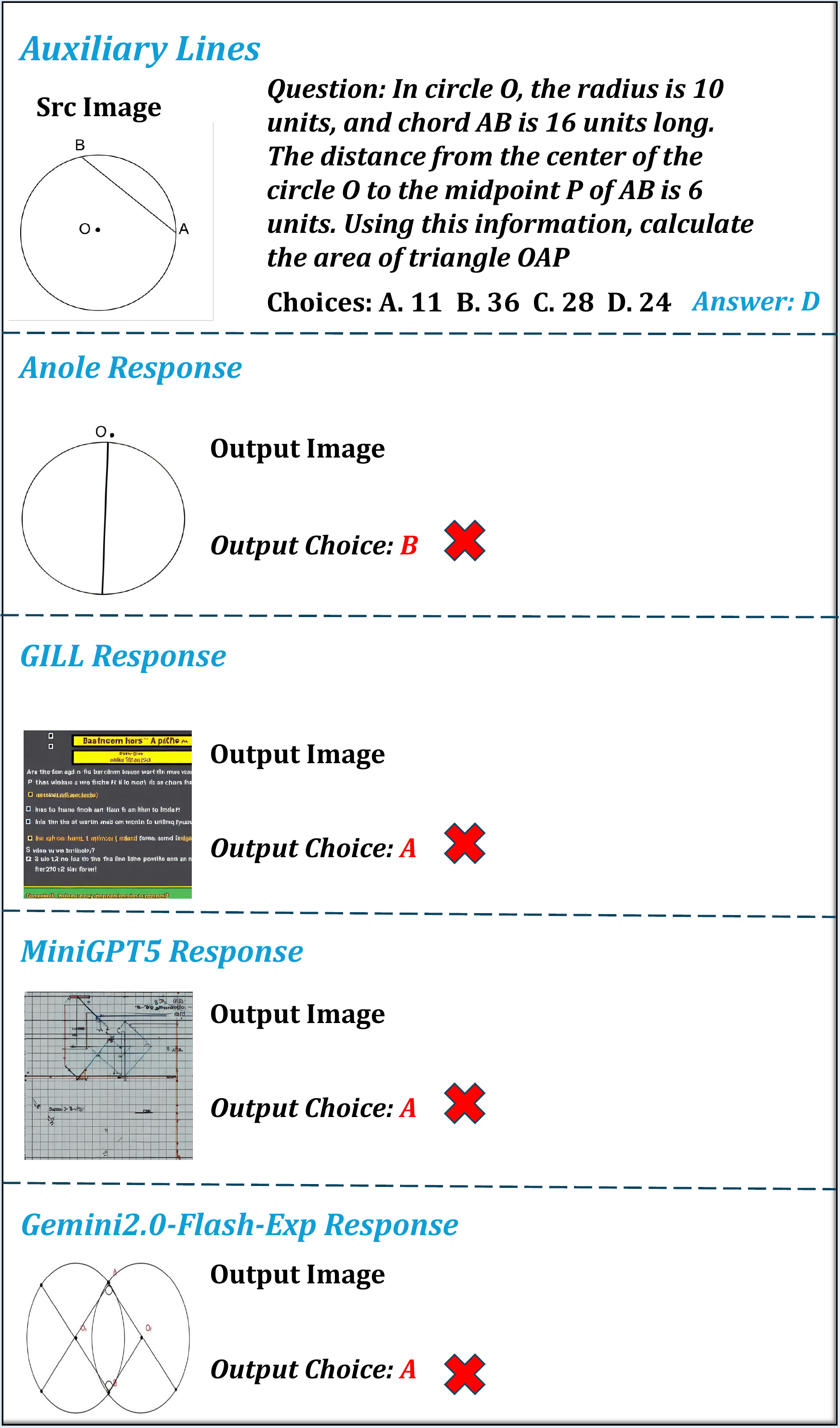}
\caption{The generated results from various models in the auxiliary lines task.}
\label{figure:al_results}
\end{figure*}

\begin{figure*}  
\centering
\includegraphics[width=0.7\linewidth]{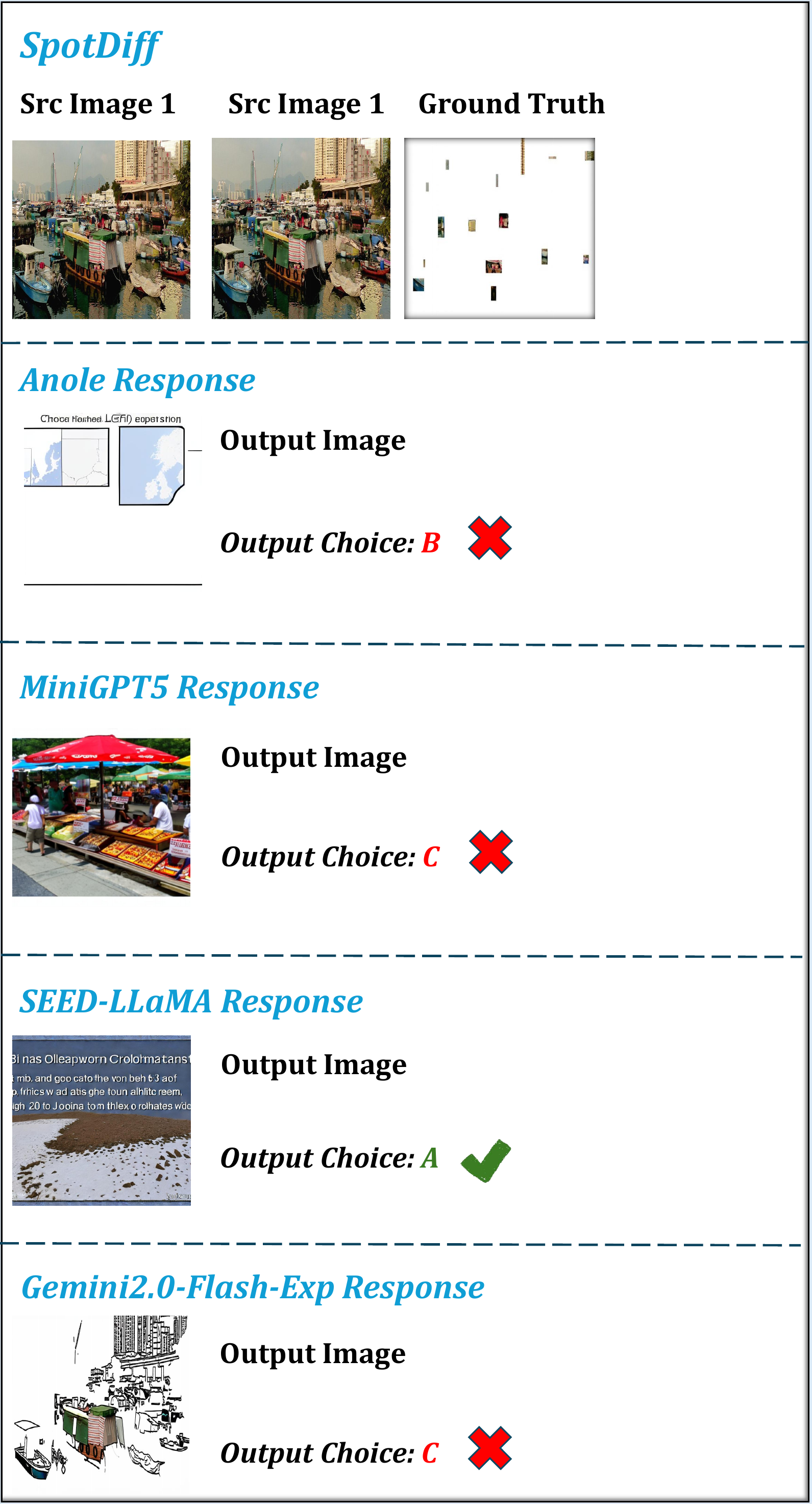}
\caption{The generated results from various models in the spotdiff task.}
\label{figure:sd_results}
\end{figure*}


\begin{table*}[htbp]
\centering
\renewcommand{\arraystretch}{1.5} 
\setlength{\tabcolsep}{4pt} 
\resizebox{1\textwidth}{!}{
\Large 
\begin{tabular}{l|cccc|cc|ccc|ccc|cccc|ccc|c}
\hline
\textbf{Method} & \multicolumn{4}{c|}{\textbf{CIVG}} & \multicolumn{2}{c|}{\textbf{FIR}} & \multicolumn{3}{c|}{\textbf{TIE}} & \multicolumn{3}{c|}{\textbf{TIG}} & \multicolumn{4}{c|}{\textbf{TVG}} & \multicolumn{3}{c|}{\textbf{VP}} & \textbf{Generation Score} \\
\hline
 \textbf{Metric} & \textbf{FVD Score} & \textbf{FID Score} & \textbf{CLIPSIM} & \textbf{Avg} & \textbf{1-LPIPS} & \textbf{Avg} & \textbf{CLIP-I} & \textbf{CLIP-T} & \textbf{Avg} & \textbf{CLIP-I} & \textbf{CLIP-T} & \textbf{Avg} & \textbf{FVD Score} & \textbf{FID Score} & \textbf{CLIPSIM} & \textbf{Avg} & \textbf{FVD Score} & \textbf{FID Score} & \textbf{Avg} & \textbf{Avg} \\
\hline
\rowcolor{gray!15} \multicolumn{21}{c}{\LARGE\textit{\textbf{Generative Models}}} \\
DALL-E-2 & - & - & - & - & - & - & - & - & - & 69.33 & 31.91 & 50.62 & - & - & - & - & - & - & - & 8.44 \\
DALL-E-3 & - & - & - & - & - & - & - & - & - & 70.11 & \underline{32.68} & 51.40 & - & - & - & - & - & - & - & 8.57 \\
OmniGen & - & - & - & - & 48.82 & 48.82 & 65.63 & \underline{22.00} & \underline{43.82} & 73.97 & 28.12 & 51.05 & - & - & - & - & - & - & - & 23.95 \\
CogVideoX & \underline{83.91} & \underline{87.02} & \underline{33.23} & \underline{68.05} & - & - & - & - & - & - & - & - & \underline{87.82} & \underline{84.28} & \underline{36.77} & \underline{69.62} & \underline{89.92} & \underline{85.30} & \underline{87.61} & 37.54 \\
\hline
\rowcolor{gray!15} \multicolumn{21}{c}{\LARGE\textit{\textbf{Unified Models}}} \\
DeepSeek-Flow & - & - & - & - & - & - & - & - & - & 52.38 & 13.38 & 32.88 & - & - & - & - & - & - & - & 5.48 \\
DeepSeek-Janus-Pro & - & - & - & - & - & - & - & - & - & 55.46 & 15.11 & 35.29 & - & - & - & - & - & - & - & 5.88 \\
Show-o & - & - & - & - & - & - & - & - & - & 62.10 & 24.97 & 43.54 & - & - & - & - & - & - & - & 7.26 \\
HermesFlow & - & - & - & - & - & - & - & - & - & 65.37 & 27.58 & 46.48 & - & - & - & - & - & - & - & 7.75 \\
Emu3 & - & - & - & - & - & - & - & - & - & 68.54 & 29.62 & 49.08 & - & - & - & - & - & - & - & 8.18 \\
VILA-U & - & - & - & - & - & - & - & - & - & 62.54 & 27.66 & 45.10 & 57.35 & 66.36 & 25.22 & 49.64 & - & - & - & 15.80 \\
MiniGPT-5 & - & - & - & - & 38.96 & 38.96 & 55.86 & 14.21 & 35.04 & 56.33 & 14.62 & 35.48 & - & - & - & - & - & - & - & 18.25 \\
Anole & - & - & - & - & 36.64 & 36.64 & 62.35 & 21.24 & 41.80 & 60.23 & 21.75 & 41.00 & - & - & - & - & - & - & - & 19.91 \\
GILL & - & - & - & - & 50.67 & 50.67 & 54.15 & 17.27 & 35.71 & 67.75 & 25.44 & 46.60 & - & - & - & - & - & - & - & 22.16 \\
SEED-LLaMA & - & - & - & - & 57.00 & 57.00 & 67.12 & 17.39 & 42.26 & 60.57 & 23.34 & 41.96 & - & - & - & - & - & - & - & 23.54 \\
Gemini-2.0-flash-exp & - & - & - & - & \underline{77.61} & \underline{77.61} & 67.77 & 19.30 & 43.54 & \underline{84.59} & 30.53 & \underline{57.56} & - & - & - & - & - & - & - & 29.79 \\
MIO-Instruct & 59.93 & 70.38 & 23.41 & 51.24 & 59.29 & 59.29 & \underline{68.12} & 19.20 & 43.66 & 72.69 & 23.77 & 48.23 & 60.03 & 69.22 & 26.40 & 51.88 & 64.08 & 68.66 & 66.37 & \underline{53.45} \\
\hline
\end{tabular}%
}
\caption{\textbf{Comparison of multimodal models on various generation tasks.} \textbf{CIVG:} Conditional Image-to-Video Generation; \textbf{FIR:} Fine-grained Image Reconstruction; \textbf{TIE:} Text-Guided Image Editing; \textbf{TIG:} Text-to-Image Generation; \textbf{TVG:} Text-to-Video Generation; \textbf{VP:} Video Prediction. $^*$ denotes MLLMs with the ability to generate interleaved images and texts, while `-' indicates that the model does not have the ability to achieve the corresponding task and \underline{underlined} content signifies the best performance within a single model across all methods on this task.}
\label{tab:main_generation}
\end{table*}

\begin{table*}[htbp]
\centering
\setlength{\tabcolsep}{4pt} 
\resizebox{1\textwidth}{!}{
\Large 
\begin{tabular}{l|ccccccccccccccc|c}
\hline
\textbf{Task} & \multicolumn{15}{c|}{\textbf{Dataset}} & \textbf{Total} \\
\cline{2-16}
 & \textbf{MME} & \textbf{MMBench} & \textbf{\makecell{MME-\\Realworld}} & \textbf{\makecell{SEED-\\Bench-2}} & \textbf{\makecell{Video-\\MME}} & \textbf{\makecell{Imagen\\Hub}} & \textbf{\makecell{Emu-\\Edit}} & \textbf{\makecell{TIP-\\I2V}} & \textbf{COCO} & \textbf{\makecell{Image\\Net}} & \textbf{\makecell{MSR-\\VTT}} & \textbf{\makecell{Pexel\\Videos}} & \textbf{\makecell{Geometry\\3K}} & \textbf{\makecell{Spot\\Diff}} & \textbf{\makecell{Open\\AI}} & \textbf{Samples} \\
\hline
\rowcolor{gray!15} \multicolumn{17}{c}{\LARGE\textit{\textbf{Understanding Task}}} \\
SIPU & 400 & 400 & 400 & 0 & 0 & 0 & 0 & 0 & 0 & 0 & 0 & 0 & 0 & 0 & 0 & 1,200 \\
MITIU & 0 & 0 & 0 & 400 & 0 & 0 & 0 & 0 & 0 & 0 & 0 & 0 & 0 & 0 & 0 & 400 \\
VPU & 0 & 0 & 0 & 0 & 364 & 0 & 0 & 0 & 0 & 0 & 0 & 0 & 0 & 0 & 0 & 364 \\
\hline
\rowcolor{gray!15} \multicolumn{17}{c}{\LARGE\textit{\textbf{Generative Task}}} \\
CIVG & 0 & 0 & 0 & 0 & 0 & 0 & 0 & 200 & 0 & 0 & 0 & 0 & 0 & 0 & 0 & 200 \\
FIR & 0 & 0 & 0 & 0 & 0 & 0 & 0 & 0 & 0 & 200 & 0 & 0 & 0 & 0 & 0 & 200 \\
TIG & 0 & 0 & 0 & 0 & 0 & 0 & 0 & 0 & 200 & 0 & 0 & 0 & 0 & 0 & 0 & 200 \\
TIE & 0 & 0 & 0 & 0 & 0 & 400 & 200 & 0 & 0 & 0 & 0 & 0 & 0 & 0 & 0 & 600 \\
TVG & 0 & 0 & 0 & 0 & 0 & 0 & 0 & 0 & 0 & 0 & 200 & 0 & 0 & 0 & 0 & 200 \\
VP & 0 & 0 & 0 & 0 & 0 & 0 & 0 & 0 & 0 & 0 & 0 & 194 & 0 & 0 & 0 & 194 \\
\hline
\rowcolor{gray!15} \multicolumn{17}{c}{\LARGE\textit{\textbf{Unify Task}}} \\
IEE & 0 & 0 & 0 & 0 & 0 & 0 & 200 & 0 & 0 & 0 & 0 & 0 & 0 & 0 & 0 & 200 \\
CSQ & 0 & 0 & 0 & 0 & 0 & 0 & 0 & 0 & 0 & 0 & 0 & 0 & 0 & 0 & 100 & 100 \\
AL & 0 & 0 & 0 & 0 & 0 & 0 & 0 & 0 & 0 & 0 & 0 & 0 & 52 & 0 & 0 & 52 \\
SD & 0 & 0 & 0 & 0 & 0 & 0 & 0 & 0 & 0 & 0 & 0 & 0 & 0 & 104 & 0 & 104 \\
VCoT & 0 & 0 & 0 & 0 & 0 & 0 & 0 & 0 & 0 & 0 & 0 & 0 & 0 & 0 & 90 & 90 \\
\hline
\textbf{Dataset Total} & \textbf{400} & \textbf{400} & \textbf{400} & \textbf{400} & \textbf{364} & \textbf{400} & \textbf{400} & \textbf{200} & \textbf{200} & \textbf{200} & \textbf{200} & \textbf{194} & \textbf{52} & \textbf{104} & \textbf{190} & \textbf{4104} \\
\textbf{Dataset \%} & \textbf{9.75\%} & \textbf{9.75\%} & \textbf{9.75\%} & \textbf{9.75\%} & \textbf{8.87\%} & \textbf{9.75\%} & \textbf{9.75\%} & \textbf{4.87\%} & \textbf{4.87\%} & \textbf{4.87\%} & \textbf{4.87\%} & \textbf{4.73\%} & \textbf{1.27\%} & \textbf{2.54\%} & \textbf{4.63\%} & \textbf{100\%} \\
\hline
\end{tabular}%
}
\caption{\textbf{Task-Dataset Sampling Statistics.} This table presents the distribution of samples across different multimodal AI tasks and their source datasets. Tasks are categorized into three main groups: Understanding Tasks (SIPU: Single Image Perception and Understanding, MITIU: Multi-Image \& Interleaved Text-Image Understanding, VPU: Video Perception and Understanding), Generative Tasks (CIVG: Conditional Image-to-Video Generation, FIR: Fine-grained Image Reconstruction, TIG: Text-to-Image Generation, TIE: Text-Guided Image Editing, TVG: Text-to-Video Generation, VP: Video Prediction), and Unify Tasks (IEE: Image Editing and Explanation, CSQ: Common Sense Question Answring, AL: Auxiliary Lines., SD: SpotDiff, VCoT: Visual CoT). The rightmost column shows the total number of samples used for each task across all datasets. A value of 0 indicates that no samples were drawn from that dataset for the corresponding task.}
\label{tab:task-dataset-sampling}
\end{table*}

\section{Extended Experimental Results}\label{sec:app_exp}

\subsection{Most U-MLLMs Exhibit Inferior Generation Capabilities}
While the methods in Table~\ref{tab:main_generation} show relatively small differences compared to the current state-of-the-art (SOTA) generation techniques, we found that using CLIP scores for evaluation introduces certain risks of manipulation. 

In Figure~\ref{fig:fir_results}, we present the results on the fine-grained image reconstruction task. For each model, we used a unified prompt: ``Reconstruct high-fidelity images from degraded inputs, preserving fine-grained details, textures, and structural integrity with perceptual realism.'' It is evident that GILL, SEED-LLaMA, and MIO-Instruct effectively capture the structural details of the input images and produce noticeably clearer outputs. In particular, SEED-LLaMA and MIO-Instruct demonstrate strong performance in restoring color fidelity, while Gemini2.0-flash-exp tends to preserve the integrity of the input images.  In contrast, MiniGPT-5 and Anole fail to effectively extract the necessary visual information: while MiniGPT-5 does generate an image, its output deviates significantly from the source, and Anole is unable to generate a coherent image at all.

Figure~\ref{fig:tie_results} displays the results for the text-guided image editing task, where the editing instruction was ``Change this image into a watercolor art.'' Similar to the reconstruction task, SEED-LLaMA and MIO-Instruct generate images that more closely resemble the source image; however, they fall short in accurately executing the specified editing instruction. Meanwhile, GILL, MiniGPT-5, and Anole show limited capability in capturing and manipulating the requisite visual details for the transformation. Notably, Gemini2.0-flash-exp not only preserves the content of the source image effectively but also accurately implements modifications according to the editing instructions.

Figure~\ref{fig:cvig_results} illustrates the performance gap between pure video generation models and U-MLLMs on the conditional image-to-video generation task. Using the text prompt ``The man is so tired. -camera zoom in,'' we observe that although MIO-Instruct produces video outputs with richer visual details compared to CogVideoX, it struggles to effectively generate a coherent video sequence that adheres to the given instruction based on the initial image.

In Figure~\ref{fig:tvg_results}, the generation results of CogVideoX and MIO-Instruct in the Text-to-Video Generation task are compared. The results clearly indicate that, in terms of both instruction adherence and video consistency, MIO-Instruct significantly underperforms compared to dedicated video generation models.

Overall, while some U-MLLMs exhibit promising capabilities in capturing visual details and producing high-fidelity reconstructions, challenges remain in faithfully executing complex editing instructions and generating consistent video sequences. These findings highlight critical areas for further improvement in enhancing the generation capabilities of U-MLLM systems.

\subsection{Challenges in Simultaneously Generating High-Quality Text and Images in U-MLLMs}
Figures~\ref{figure:iee_results}, ~\ref{figure:csq_results}, ~\ref{figure:al_results}, and ~\ref{figure:sd_results} present the results of U-MLLMs on the Unify tasks. Notably, MIO-Instruct fails to perform any text-image generation across all Unify tasks, GILL is unable to generate multimodal outputs in the SpotDiff task, and SEED-LLaMA does not support text-image generation in the Auxiliary Lines task. Overall, these results indicate that most U-MLLMs struggle to generate images that faithfully adhere to provided instructions or reference images, and their comprehension of the instructions is often flawed.

In the Image Editing and Explanation task, for instance, MiniGPT-5 produced images that bore no relation to the source images. Additionally, the textual outputs from GILL, MiniGPT-5, and SEED-LLaMA were insufficient for accurately describing the editing objects or the instructions. Similarly, in both the Commonsense Question Answering and SpotDiff tasks, although MiniGPT-5 and SEED-LLaMA correctly answered the textual multiple-choice questions, the images they generated were clearly unrelated to the corresponding options. This further emphasizes the difficulty U-MLLMs face in maintaining consistency between textual and visual outputs.

For the Auxiliary Lines task, while Anole managed to generate images that retained some of the visual details of the source images, it failed to correctly draw the required auxiliary lines as per the instructions. GILL and MiniGPT-5, on the other hand, generated content that was completely disconnected from the original images.

These findings suggest several critical limitations in current U-MLLM systems. First, there is a notable gap in their ability to integrate and utilize multimodal cues effectively, as evidenced by the misalignment between textual instructions and visual outputs. Second, while some models can capture certain visual details, they often lack the robust reasoning required to follow complex instructions, especially in tasks demanding precise visual modifications. Finally, the decoupling between text and image generation in these systems underscores the need for further research aimed at improving cross-modal coherence and instruction fidelity.

Overall, the experimental results highlight that, despite progress in individual modalities, existing U-MLLMs have considerable challenges in simultaneously generating high-quality, coherent text and images that align with complex, multimodal instructions.

%% file: main.bbl
\begin{thebibliography}{49}
\providecommand{\natexlab}[1]{#1}
\providecommand{\url}[1]{\texttt{#1}}
\expandafter\ifx\csname urlstyle\endcsname\relax
  \providecommand{\doi}[1]{doi: #1}\else
  \providecommand{\doi}{doi: \begingroup \urlstyle{rm}\Url}\fi

\bibitem[Anthropic(2024)]{anthropic2024claude}
Anthropic.
\newblock Introducing claude 3.5 sonnet, 2024.

\bibitem[Brooks et~al.(2023)Brooks, Holynski, and Efros]{brooks2022instructpix2pix}
Tim Brooks, Aleksander Holynski, and Alexei~A. Efros.
\newblock Instructpix2pix: Learning to follow image editing instructions.
\newblock In \emph{CVPR}, 2023.

\bibitem[Chen et~al.(2025)Chen, Wu, Liu, Pan, Liu, Xie, Yu, and Ruan]{chen2025janus}
Xiaokang Chen, Zhiyu Wu, Xingchao Liu, Zizheng Pan, Wen Liu, Zhenda Xie, Xingkai Yu, and Chong Ruan.
\newblock Janus-pro: Unified multimodal understanding and generation with data and model scaling.
\newblock \emph{arXiv}, 2025.

\bibitem[Chern et~al.(2024)Chern, Su, Ma, and Liu]{chern2024anole}
Ethan Chern, Jiadi Su, Yan Ma, and Pengfei Liu.
\newblock Anole: An open, autoregressive, native large multimodal models for interleaved image-text generation.
\newblock \emph{arXiv}, 2024.

\bibitem[DeepMind(2024)]{deepmind2024gemini}
Google DeepMind.
\newblock Gemini flash, 2024.

\bibitem[Deng et~al.(2009)Deng, Dong, Socher, Li, Li, and Fei-Fei]{5206848}
Jia Deng, Wei Dong, Richard Socher, Li-Jia Li, Kai Li, and Li Fei-Fei.
\newblock Imagenet: A large-scale hierarchical image database.
\newblock In \emph{CVPR}, 2009.

\bibitem[Fang et~al.(2024)Fang, Duan, Wang, Li, Tian, Zeng, Zhao, Dai, Li, and Liu]{fang2024puma}
Rongyao Fang, Chengqi Duan, Kun Wang, Hao Li, Hao Tian, Xingyu Zeng, Rui Zhao, Jifeng Dai, Hongsheng Li, and Xihui Liu.
\newblock Puma: Empowering unified mllm with multi-granular visual generation.
\newblock \emph{arXiv}, 2024.

\bibitem[Fang et~al.(2025)Fang, Mao, Duan, Zhao, Li, Lin, and Chen]{fang2025mmbench}
Xinyu Fang, Kangrui Mao, Haodong Duan, Xiangyu Zhao, Yining Li, Dahua Lin, and Kai Chen.
\newblock Mmbench-video: A long-form multi-shot benchmark for holistic video understanding.
\newblock \emph{NeurIPS}, 2025.

\bibitem[Fei et~al.(2024)Fei, Wu, Zhang, Chua, and Yan]{fei2024vitron}
Hao Fei, Shengqiong Wu, Hanwang Zhang, Tat-Seng Chua, and Shuicheng Yan.
\newblock Vitron: A unified pixel-level vision llm for understanding, generating, segmenting, editing.
\newblock In \emph{NeurIPS}, 2024.

\bibitem[Fu et~al.(2023)Fu, Chen, Shen, Qin, Zhang, Lin, Yang, Zheng, Li, Sun, et~al.]{fu2023mme}
Chaoyou Fu, Peixian Chen, Yunhang Shen, Yulei Qin, Mengdan Zhang, Xu Lin, Jinrui Yang, Xiawu Zheng, Ke Li, Xing Sun, et~al.
\newblock Mme: A comprehensive evaluation benchmark for multimodal large language models.
\newblock \emph{arXiv}, 2023.

\bibitem[Fu et~al.(2024{\natexlab{a}})Fu, Dai, Luo, Li, Ren, Zhang, Wang, Zhou, Shen, Zhang, et~al.]{fu2024video}
Chaoyou Fu, Yuhan Dai, Yondong Luo, Lei Li, Shuhuai Ren, Renrui Zhang, Zihan Wang, Chenyu Zhou, Yunhang Shen, Mengdan Zhang, et~al.
\newblock Video-mme: The first-ever comprehensive evaluation benchmark of multi-modal llms in video analysis.
\newblock \emph{arXiv}, 2024{\natexlab{a}}.

\bibitem[Fu et~al.(2024{\natexlab{b}})Fu, Zhang, Yin, Li, Fang, Zhao, Duan, Sun, Liu, Wang, et~al.]{fu2024mme}
Chaoyou Fu, Yi-Fan Zhang, Shukang Yin, Bo Li, Xinyu Fang, Sirui Zhao, Haodong Duan, Xing Sun, Ziwei Liu, Liang Wang, et~al.
\newblock Mme-survey: A comprehensive survey on evaluation of multimodal llms.
\newblock \emph{arXiv}, 2024{\natexlab{b}}.

\bibitem[Fu et~al.(2025)Fu, Lin, Wang, Zhang, Shen, Liu, Li, Long, Gao, Li, et~al.]{fu2025vita}
Chaoyou Fu, Haojia Lin, Xiong Wang, Yi-Fan Zhang, Yunhang Shen, Xiaoyu Liu, Yangze Li, Zuwei Long, Heting Gao, Ke Li, et~al.
\newblock Vita-1.5: Towards gpt-4o level real-time vision and speech interaction.
\newblock \emph{arXiv preprint arXiv:2501.01957}, 2025.

\bibitem[Ge et~al.(2024)Ge, Zhao, Zeng, Ge, Li, Wang, and Shan]{ge2023making}
Yuying Ge, Sijie Zhao, Ziyun Zeng, Yixiao Ge, Chen Li, Xintao Wang, and Ying Shan.
\newblock Making llama see and draw with seed tokenizer.
\newblock In \emph{ICLR}, 2024.

\bibitem[Ghosh et~al.(2023)Ghosh, Hajishirzi, and Schmidt]{ghosh2023geneval}
Dhruba Ghosh, Hannaneh Hajishirzi, and Ludwig Schmidt.
\newblock Geneval: An object-focused framework for evaluating text-to-image alignment.
\newblock \emph{NeurIPS}, 2023.

\bibitem[Huang et~al.(2024)Huang, He, Yu, Zhang, Si, Jiang, Zhang, Wu, Jin, Chanpaisit, et~al.]{huang2024vbench}
Ziqi Huang, Yinan He, Jiashuo Yu, Fan Zhang, Chenyang Si, Yuming Jiang, Yuanhan Zhang, Tianxing Wu, Qingyang Jin, Nattapol Chanpaisit, et~al.
\newblock Vbench: Comprehensive benchmark suite for video generative models.
\newblock In \emph{CVPR}, 2024.

\bibitem[Koh et~al.(2023)Koh, Fried, and Salakhutdinov]{koh2023generating}
Jing~Yu Koh, Daniel Fried, and Ruslan Salakhutdinov.
\newblock Generating images with multimodal language models.
\newblock \emph{NeurIPS}, 2023.

\bibitem[Ku et~al.(2024)Ku, Li, Zhang, Lu, Fu, Zhuang, and Chen]{ku2024imagenhub}
Max Ku, Tianle Li, Kai Zhang, Yujie Lu, Xingyu Fu, Wenwen Zhuang, and Wenhu Chen.
\newblock Imagenhub: Standardizing the evaluation of conditional image generation models.
\newblock In \emph{ICLR}, 2024.

\bibitem[Li et~al.(2023)Li, Ge, Ge, Wang, Wang, Zhang, and Shan]{li2023seed2}
Bohao Li, Yuying Ge, Yixiao Ge, Guangzhi Wang, Rui Wang, Ruimao Zhang, and Ying Shan.
\newblock Seed-bench-2: Benchmarking multimodal large language models.
\newblock \emph{arXiv}, 2023.

\bibitem[Li et~al.(2025)Li, Wu, Zhang, Xia, Mao, Dong, Vuli{\'c}, and Wei]{li2025imagine}
Chengzu Li, Wenshan Wu, Huanyu Zhang, Yan Xia, Shaoguang Mao, Li Dong, Ivan Vuli{\'c}, and Furu Wei.
\newblock Imagine while reasoning in space: Multimodal visualization-of-thought.
\newblock \emph{arXiv}, 2025.

\bibitem[Lin et~al.(2014{\natexlab{a}})Lin, Maire, Belongie, Hays, Perona, Ramanan, Doll{\'a}r, and Zitnick]{lin2014microsoft}
Tsung-Yi Lin, Michael Maire, Serge Belongie, James Hays, Pietro Perona, Deva Ramanan, Piotr Doll{\'a}r, and C~Lawrence Zitnick.
\newblock Microsoft coco: Common objects in context.
\newblock In \emph{ECCV}, 2014{\natexlab{a}}.

\bibitem[Lin et~al.(2014{\natexlab{b}})Lin, Maire, Belongie, Hays, Perona, Ramanan, Doll{\'a}r, and Zitnick]{Lin2014MicrosoftCC}
Tsung-Yi Lin, Michael Maire, Serge~J. Belongie, James Hays, Pietro Perona, Deva Ramanan, Piotr Doll{\'a}r, and C.~Lawrence Zitnick.
\newblock Microsoft coco: Common objects in context.
\newblock In \emph{ECCV}, 2014{\natexlab{b}}.

\bibitem[Liu et~al.(2024)Liu, Duan, Zhang, Li, Zhang, Zhao, Yuan, Wang, He, Liu, et~al.]{liu2023mmbench}
Yuan Liu, Haodong Duan, Yuanhan Zhang, Bo Li, Songyang Zhang, Wangbo Zhao, Yike Yuan, Jiaqi Wang, Conghui He, Ziwei Liu, et~al.
\newblock Mmbench: Is your multi-modal model an all-around player?
\newblock \emph{ECCV}, 2024.

\bibitem[Ma et~al.(2024)Ma, Liu, Chen, Liu, Wu, Wu, Pan, Xie, Zhang, Zhao, et~al.]{ma2024janusflow}
Yiyang Ma, Xingchao Liu, Xiaokang Chen, Wen Liu, Chengyue Wu, Zhiyu Wu, Zizheng Pan, Zhenda Xie, Haowei Zhang, Liang Zhao, et~al.
\newblock Janusflow: Harmonizing autoregression and rectified flow for unified multimodal understanding and generation.
\newblock \emph{NeurIPS}, 2024.

\bibitem[OpenAI(2024{\natexlab{a}})]{openai2024dalle}
OpenAI.
\newblock Dall·e 2, 2024{\natexlab{a}}.

\bibitem[OpenAI(2024{\natexlab{b}})]{openai2024dalle3}
OpenAI.
\newblock Dall·e 3, 2024{\natexlab{b}}.

\bibitem[OpenAI(2024{\natexlab{c}})]{openai2024gpt4o}
OpenAI.
\newblock Hello gpt-4o, 2024{\natexlab{c}}.

\bibitem[Sheynin et~al.(2024)Sheynin, Polyak, Singer, Kirstain, Zohar, Ashual, Parikh, and Taigman]{sheynin2024emu}
Shelly Sheynin, Adam Polyak, Uriel Singer, Yuval Kirstain, Amit Zohar, Oron Ashual, Devi Parikh, and Yaniv Taigman.
\newblock Emu edit: Precise image editing via recognition and generation tasks.
\newblock In \emph{CVPR}, 2024.

\bibitem[Team(2024)]{team2024chameleon}
Chameleon Team.
\newblock Chameleon: Mixed-modal early-fusion foundation models.
\newblock \emph{arXiv}, 2024.

\bibitem[Wang et~al.(2024{\natexlab{a}})Wang, Bai, Tan, Wang, Fan, Bai, Chen, Liu, Wang, Ge, et~al.]{wang2024qwen2}
Peng Wang, Shuai Bai, Sinan Tan, Shijie Wang, Zhihao Fan, Jinze Bai, Keqin Chen, Xuejing Liu, Jialin Wang, Wenbin Ge, et~al.
\newblock Qwen2-vl: Enhancing vision-language model's perception of the world at any resolution.
\newblock \emph{arXiv}, 2024{\natexlab{a}}.

\bibitem[Wang and Yang(2024)]{wang2024tip}
Wenhao Wang and Yi Yang.
\newblock Tip-i2v: A million-scale real text and image prompt dataset for image-to-video generation.
\newblock \emph{arXiv}, 2024.

\bibitem[Wang et~al.(2024{\natexlab{b}})Wang, Zhang, Luo, Sun, Cui, Wang, Zhang, Wang, Li, Yu, et~al.]{wang2024emu3}
Xinlong Wang, Xiaosong Zhang, Zhengxiong Luo, Quan Sun, Yufeng Cui, Jinsheng Wang, Fan Zhang, Yueze Wang, Zhen Li, Qiying Yu, et~al.
\newblock Emu3: Next-token prediction is all you need.
\newblock \emph{arXiv}, 2024{\natexlab{b}}.

\bibitem[Wang et~al.(2024{\natexlab{c}})Wang, Zhu, Xu, Zhou, Liu, Zhang, Wang, Shi, Li, Li, et~al.]{wang2024mio}
Zekun Wang, King Zhu, Chunpu Xu, Wangchunshu Zhou, Jiaheng Liu, Yibo Zhang, Jiashuo Wang, Ning Shi, Siyu Li, Yizhi Li, et~al.
\newblock Mio: A foundation model on multimodal tokens.
\newblock \emph{arXiv}, 2024{\natexlab{c}}.

\bibitem[Wu et~al.(2024)Wu, Zhang, Chen, Tang, Li, Fang, Zhu, Xie, Yin, Yi, et~al.]{wu2024vila}
Yecheng Wu, Zhuoyang Zhang, Junyu Chen, Haotian Tang, Dacheng Li, Yunhao Fang, Ligeng Zhu, Enze Xie, Hongxu Yin, Li Yi, et~al.
\newblock Vila-u: a unified foundation model integrating visual understanding and generation.
\newblock \emph{arXiv}, 2024.

\bibitem[Xiao et~al.(2024)Xiao, Wang, Zhou, Yuan, Xing, Yan, Wang, Huang, and Liu]{xiao2024omnigen}
Shitao Xiao, Yueze Wang, Junjie Zhou, Huaying Yuan, Xingrun Xing, Ruiran Yan, Shuting Wang, Tiejun Huang, and Zheng Liu.
\newblock Omnigen: Unified image generation.
\newblock \emph{arXiv}, 2024.

\bibitem[Xie et~al.(2025)Xie, Mao, Bai, Zhang, Wang, Lin, Gu, Chen, Yang, and Shou]{xie2024show}
Jinheng Xie, Weijia Mao, Zechen Bai, David~Junhao Zhang, Weihao Wang, Kevin~Qinghong Lin, Yuchao Gu, Zhijie Chen, Zhenheng Yang, and Mike~Zheng Shou.
\newblock Show-o: One single transformer to unify multimodal understanding and generation.
\newblock \emph{ICLR}, 2025.

\bibitem[Xu et~al.(2016)Xu, Mei, Yao, and Rui]{xu2016msr}
Jun Xu, Tao Mei, Ting Yao, and Yong Rui.
\newblock Msr-vtt: A large video description dataset for bridging video and language.
\newblock In \emph{CVPR}, 2016.

\bibitem[Yan et~al.(2024)Yan, Wang, Huo, Li, Li, Su, Gao, Zhang, Xu, Chu, et~al.]{yan2024errorradar}
Yibo Yan, Shen Wang, Jiahao Huo, Hang Li, Boyan Li, Jiamin Su, Xiong Gao, Yi-Fan Zhang, Tianlong Xu, Zhendong Chu, et~al.
\newblock Errorradar: Benchmarking complex mathematical reasoning of multimodal large language models via error detection.
\newblock \emph{arXiv}, 2024.

\bibitem[Yang et~al.(2025{\natexlab{a}})Yang, Zhang, Tian, Shang, Xu, Zhang, and Cui]{hermesflow2024}
Ling Yang, Xinchen Zhang, Ye Tian, Chenming Shang, Minghao Xu, Wentao Zhang, and Bin Cui.
\newblock Hermesflow: Seamlessly closing the gap in multimodal understanding and generation.
\newblock \emph{arXiv preprint arXiv:2502.12148}, 2025{\natexlab{a}}.

\bibitem[Yang et~al.(2025{\natexlab{b}})Yang, Teng, Zheng, Ding, Huang, Xu, Yang, Hong, Zhang, Feng, Yin, Gu, Zhang, Wang, Cheng, Liu, Xu, Dong, and Tang]{Yang2024CogVideoXTD}
Zhuoyi Yang, Jiayan Teng, Wendi Zheng, Ming Ding, Shiyu Huang, Jiazheng Xu, Yuanming Yang, Wenyi Hong, Xiaohan Zhang, Guanyu Feng, Da Yin, Xiaotao Gu, Yuxuan Zhang, Weihan Wang, Yean Cheng, Ting Liu, Bin Xu, Yuxiao Dong, and Jie Tang.
\newblock Cogvideox: Text-to-video diffusion models with an expert transformer.
\newblock \emph{ICLR}, 2025{\natexlab{b}}.

\bibitem[Yu et~al.(2022)Yu, Li, Koh, Zhang, Pang, Qin, Ku, Xu, Baldridge, and Wu]{yu2021vector}
Jiahui Yu, Xin Li, Jing~Yu Koh, Han Zhang, Ruoming Pang, James Qin, Alexander Ku, Yuanzhong Xu, Jason Baldridge, and Yonghui Wu.
\newblock Vector-quantized image modeling with improved vqgan.
\newblock \emph{ICLR}, 2022.

\bibitem[Yu et~al.(2025)Yu, Fu, Wu, Lu, Wang, Lu, Shen, Zhang, Song, Yan, et~al.]{yu2025aligning}
Tao Yu, Chaoyou Fu, Junkang Wu, Jinda Lu, Kun Wang, Xingyu Lu, Yunhang Shen, Guibin Zhang, Dingjie Song, Yibo Yan, et~al.
\newblock Aligning multimodal llm with human preference: A survey.
\newblock \emph{arXiv preprint arXiv:2503.14504}, 2025.

\bibitem[Zhang et~al.(2023)Zhang, Mo, Chen, Sun, and Su]{zhang2023magicbrush}
Kai Zhang, Lingbo Mo, Wenhu Chen, Huan Sun, and Yu Su.
\newblock Magicbrush: A manually annotated dataset for instruction-guided image editing.
\newblock \emph{NeurIPS}, 2023.

\bibitem[Zhang et~al.(2024)Zhang, Wen, Fu, Wang, Zhang, Wang, and Jin]{zhang2024beyond}
Yi-Fan Zhang, Qingsong Wen, Chaoyou Fu, Xue Wang, Zhang Zhang, Liang Wang, and Rong Jin.
\newblock Beyond llava-hd: Diving into high-resolution large multimodal models.
\newblock \emph{arXiv preprint arXiv:2406.08487}, 2024.

\bibitem[Zhang et~al.(2025{\natexlab{a}})Zhang, Yu, Tian, Fu, Li, Zeng, Xie, Shi, Zhang, Wu, et~al.]{zhang2025mm}
Yi-Fan Zhang, Tao Yu, Haochen Tian, Chaoyou Fu, Peiyan Li, Jianshu Zeng, Wulin Xie, Yang Shi, Huanyu Zhang, Junkang Wu, et~al.
\newblock Mm-rlhf: The next step forward in multimodal llm alignment.
\newblock \emph{arXiv}, 2025{\natexlab{a}}.

\bibitem[Zhang et~al.(2025{\natexlab{b}})Zhang, Zhang, Tian, Fu, Zhang, Wu, Li, Wang, Wen, Zhang, et~al.]{zhang2024mme}
Yi-Fan Zhang, Huanyu Zhang, Haochen Tian, Chaoyou Fu, Shuangqing Zhang, Junfei Wu, Feng Li, Kun Wang, Qingsong Wen, Zhang Zhang, et~al.
\newblock Mme-realworld: Could your multimodal llm challenge high-resolution real-world scenarios that are difficult for humans?
\newblock \emph{ICLR}, 2025{\natexlab{b}}.

\bibitem[Zhao et~al.(2024)Zhao, Tang, Wu, Lin, Wei, Liu, Tan, Zhang, Huang, and Xie]{Zhao2024HarmonizingVT}
Zhen Zhao, Jingqun Tang, Binghong Wu, Chunhui Lin, Shubo Wei, Hao Liu, Xin Tan, Zhizhong Zhang, Can Huang, and Yuan Xie.
\newblock Harmonizing visual text comprehension and generation.
\newblock \emph{ArXiv}, 2024.

\bibitem[Zheng et~al.(2023)Zheng, He, and Wang]{zheng2023minigpt5}
Kaizhi Zheng, Xuehai He, and Xin~Eric Wang.
\newblock Minigpt-5: Interleaved vision-and-language generation via generative vokens.
\newblock \emph{arXiv}, 2023.

\bibitem[Zhou et~al.(2025)Zhou, Yu, Babu, Tirumala, Yasunaga, Shamis, Kahn, Ma, Zettlemoyer, and Levy]{zhou2024transfusion}
Chunting Zhou, Lili Yu, Arun Babu, Kushal Tirumala, Michihiro Yasunaga, Leonid Shamis, Jacob Kahn, Xuezhe Ma, Luke Zettlemoyer, and Omer Levy.
\newblock Transfusion: Predict the next token and diffuse images with one multi-modal model.
\newblock \emph{ICLR}, 2025.

\end{thebibliography}
